\documentclass[twoside]{article}
\usepackage{ecj,palatino,epsfig,latexsym,natbib}

\makeatletter
\def\set@curr@file#1{%
  \begingroup
    \escapechar\m@ne
    \xdef\@curr@file{\expandafter\string\csname #1\endcsname}%
  \endgroup
}
\def\quote@name#1{"\quote@@name#1\@gobble""}
\def\quote@@name#1"{#1\quote@@name}
\def\unquote@name#1{\quote@@name#1\@gobble"}
\makeatother

\usepackage{amsmath}
\usepackage{amsfonts}
\usepackage{amssymb}

\usepackage{graphicx}
\usepackage{xcolor}

\usepackage{algorithm}
\usepackage{algpseudocode}

\usepackage{enumitem}
\setlist[itemize]{noitemsep, nolistsep}

\usepackage{lineno,hyperref}
\usepackage{xparse,mathtools}
\modulolinenumbers[5]

\begin{document}







\title{\bf Generalized Self-Adapting Particle Swarm Optimization algorithm with archive of samples}

		\author{\name{\bf M. Okulewicz} \hfill \addr{M.Okulewicz@mini.pw.edu.pl}\\ 
        \addr{Faculty of Mathematics and Information Science, Warsaw University of Technology, 
        Warsaw, 00-662, Poland}
\AND
		\name{\bf M. Zaborski} \hfill \addr{M.Zaborski@mini.pw.edu.pl}\\
		\addr{Faculty of Mathematics and Information Science, Warsaw University of Technology, 
		Warsaw, 00-662, Poland}
\AND
		\name{\bf J. Ma{\'n}dziuk} \hfill \addr{J.Mandziuk@mini.pw.edu.pl}\\
		\addr{Faculty of Mathematics and Information Science, Warsaw University of Technology, 
		Warsaw, 00-662, Poland}
}

\maketitle

\begin{abstract}
In this paper we enhance Generalized Self-Adapting Particle Swarm Optimization algorithm (GAPSO), initially introduced at the Parallel Problem Solving from Nature 2018 conference, and to investigate its properties. The research on GAPSO is underlined by the two following assumptions:
(1) it is possible to achieve good performance of an optimization algorithm
through utilization of all of the gathered samples,
(2)~the~best performance can be accomplished by means of a combination of specialized sampling behaviors
(Particle Swarm Optimization, Differential Evolution, and locally fitted square functions).

From a software engineering point of view, GAPSO considers a standard Particle Swarm Optimization algorithm
as an ideal starting point for creating a general-purpose global optimization framework.
Within this framework hybrid optimization algorithms are developed,
and various additional techniques (like algorithm restart management
or adaptation schemes) are tested.

The paper introduces a new version of the algorithm, abbreviated as M-GAPSO.
In comparison with the original GAPSO formulation it includes the following four features:
a global restart management scheme,
samples gathering within an R-Tree based index (archive/memory of samples),
adaptation of a sampling behavior based on a global particle performance, and a specific approach to local search.

The above-mentioned enhancements resulted in improved performance of M-GAPSO over GAPSO,
observed on both COCO BBOB testbed and in the black-box optimization competition BBComp.
Also, for lower dimensionality functions (up to 5D) results of M-GAPSO are
better or comparable to the state-of-the art version of CMA-ES (namely the
KL-BIPOP-CMA-ES algorithm presented at the GECCO 2017 conference).
\end{abstract}



\section{Introduction}

The quest for the highly efficient general purpose optimization algorithms,
which started with the works on evolutionary computations by de Jong \cite{DeJong1975}
and Holland \cite{Holland1992},
resulted in creation of a few excellent optimization methods like Differential Evolution (DE)~\cite{DE}
or Covariance Matrix Adaptation Evolution Strategy (CMA-ES)~\cite{Hansen2003}.
Those methods have been thoroughly studied and gradually improved over the years following their
initial presentation \cite{Poaik2012,Loshchilov2013,Brest2016,TakahiroYamaguchi2017}.

The search for a universal optimization algorithm is a challenging task because optimization performance strongly
depends on the type of an optimized function~\cite{Wolpert1997}.
Additionally, the works on theoretical convergence of random sampling based methods
(i.e. Genetic Algorithms (GA) \cite{Eiben1991}, Simulated Annealing (SA) \cite{Eiben1991} and Particle Swarm Optimization (PSO) \cite{Poli2009,VanDenBergh2010}) are not necessarily helpful in practical setting of parameters for particular problem.

Proposed (M-)GAPSO framework provides a platform for seamless cooperation of various existing optimization approaches.
It also attempts to separate auxiliary techniques (e.g. population initialization
after algorithm's stagnation) from the actual optimization engine (e.g. PSO's particles sampling strategy).

While the resulting system built on M-GAPSO platform may be quite complex,
each of its parts remain relatively unsophisticated and independent, so as to make analysis of its impact and improvement of its performance easier.
This goal is achieved by taking a multi-agent view on the hybrid optimization.
In this view, a single search operator is indifferent to how exactly the current solutions were computed, it only needs to receive their arguments and function value.
This way, pure sampling-based approaches (e.g. PSO or DE) can be easily mixed with local function approximations, e.g. quasi-newton methods.


\subsection{Contribution}

\noindent The main contribution of this paper is five-fold:

\begin{itemize}
	\item Designing framework in which sampling methods and model-based optimization approaches seamlessly cooperate in a common environment.
	\item Performing extensive experimental studies on COCO benchmark set \cite{nikolaus_hansen_2019_2594848}, aimed at analyzing the impact of various optimization techniques and seeking efficient hybridization of various optimization methods.
	\item Achieving better results than the state-of-the-art CMA-ES \cite{TakahiroYamaguchi2017} for lower-dimen-sional functions (up to 5D) from COCO set with $DIM*10^6$ optimization budget.
	\item Accomplishing the 8th and the 4th place in the BBComp 2019 black-box optimization competition\footnote{\url{https://bbcomp.ini.rub.de/\#results}} in single-objective and expensive single-objective tracks, respectively.
\item Significantly improving the performance over the initial version of the GAPSO algorithm~\cite{Ulinski2018}.
\end{itemize}

\noindent The rest of the paper is organized as follows.
Section \ref{sec:literature} provides a literature review, with special emphasis on hybrid optimization approaches.
Section \ref{sec:algorithm} presents the original GAPSO algorithm and discusses its components (with Sections \ref{sec:initialization-scheme} and \ref{sec:function-models}
describing population re-initialization and model-based
optimization, respectively).
Section~\ref{sec:performance} summarizes results on the COCO benchmark set and those achieved during BBComp competitions%
.
Section \ref{sec:conclusions} concludes the paper.

\section{Related literature}
\label{sec:literature}
M-GAPSO is characterized by the following three main features: (1) \emph{hybridization} of optimization techniques, (2) \emph{memory-based space sampling} by means of caching, and (3) the focus on particular \emph{components of the solution method} in the context of their relevance and contribution to the overall method's efficacy.


In more detail, the work advocates the relevance of hybrid optimization methods which above all can gain from synergetic combination of advantages of the compound methods. The idea of combining two or more optimization techniques so as to mitigate the local minima problem and/or establish an optimal balance between exploration and exploitation \cite{Lynn2015} in the solution search process has recently become highly popular in Computational Intelligence (CI) community. The most prominent forms of its realization include hyperheuristic approaches and heterogeneous methods. In hyperheuristic solutions, typically, a top-level algorithm is responsible for selection of the locally optimal method to be used in a particular (current) context~\cite{VANDERSTOCKT2018127,Castro2018,LIN2017124,SCHLUNZ201858}. Heterogeneous methods, on the other hand, directly combine either two or more heuristics (a typical approach is to enhance the main method by some kind of local optimization~\cite{Loshchilov2013,mandziuk2016}) or to combine several variants of the same heuristic  method~\cite{nepomuceno2012self,deOCA4983013,Harrison2017a}). In either case the resulting optimization approach leads to a qualitatively new behavior with respect to the component methods. Witihin the domain of PSO algorithms, the problem of balancing exploration and exploitation can be also solved by introducing specialized subswarms \cite{Du2008,Lynn2015}.

Our particular focus is on a combination of PSO and DE, further enhanced by utilization of locally fitted square functions. While combinations of PSO and DE have been considered in the literature before, their further improvement stemming from locally fitted low-degree polynomials is - to our knowledge - proposed for the first time. The hybrids of PSO and DE have already proven to have potential extending beyond each of the component methods alone. For instance, Liu, Cai and  Wang~\cite{Liu2010} presented a hybrid algorithm combining PSO and DE, motivated by the fact that PSO activity is prone to stagnation when particles are unable to improve their personal best positions. In such a case DE is applied to update particles best positions and make the swarm jump out of the stagnation phase~\cite{Yu2014,Zhi-HuiZhan2009a}. The main difference between~\cite{Liu2010} and our approach is that in~\cite{Liu2010} DE is employed for searching for new personal best positions of particles while in our method samples of both PSO and DE are concurrently utilized during the whole optimization process.

A related idea is presented in the Brain Storm Optimization algorithm (BSO) proposed in~\cite{SHI10.1007/978-3-642-21515-5_36,Shi2011}, which is a swarm intelligence technique  inspired by collective behavior of humans - specifically the brainstorming approach to problem solving. BSO addresses the local minima problem by means of applying both converging and diverging operations in the quest for a COP optimum~\cite{Cheng2019,SHI10.1007/978-3-642-21515-5_36}.

Another key aspect of our method (besides hybridization of PSO and DE) is utilization of a memory-based approach by means of caching function values in sampled points and promoting the best-found particles (solutions) to the next generation. This way the allotted number of fitness function evaluations (FFEs) can be utilized in a more effective way, leading to visible improvement of results. In this context, one of the seminal papers is~\cite{Taillard2001} in which the authors discuss the memory-based properties of several popular metaheuristic algorithms, point their similarities, and consequently propose a unified view of these methods. Since~\cite{Taillard2001} does not include PSO in the presented analysis, exploration of the memory-based enhancements of PSO, proposed in this paper, seems to well complement the findings of~\cite{Taillard2001}.

The third focus of the paper is on the relevance of implementation of particular components of the optimization method (whether hybrid or pure) for its final efficacy. This research thread is a continuation of our previous works devoted to detailed analysis of this aspect in the case of pure (non-hybridized) PSO~\cite{DVRP:2PSO,DVRP:2MPSO,OkulewiczMandziuk2017,OkulewiczMandziuk2019} and Memetic Algorithm~\cite{mandziuk2016}. Specifically, in~\cite{DVRP:2MPSO,OkulewiczMandziuk2017} we analyzed a Two-Phase Multi-Swarm Particle Swarm Optimizer (2MPSO) solving the Dynamic Vehicle Routing Problem (DVRP)~\cite{Mandziuk2019survey} with the aim of finding an optimal configuration of several optimization improvement techniques dedicated to solving dynamic optimization problems within the 2MPSO framework. One of the main conclusions was that strong results achieved by 2MPSO should be mainly attributed to the following three factors: generating initial solutions with a clustering heuristic, optimizing the requests-to-vehicle assignment with a metaheuristic approach,  and direct passing of solutions obtained in the previous stage (times step) of the problem solving procedure to the next stage.

Current research directly extends our three previous papers~\cite{Okulewicz2016,Ulinski2018,ZaborskiOkulewiczMandziuk2019}.
The first one discussed the possibility of finding an optimal team of optimization agents
based on Belbin's Team Roles Inventory.
The second one introduced the initial version of GAPSO algorithm
as a hybrid of several variants of the PSO and DE methods.
The third one presented initial results on model-based optimization behaviors
within GAPSO framework.

\section{Description of M-GAPSO}
\label{sec:algorithm}

This section describes the proposed Generalized Self-Adapting Particle Swarm Optimization (GAPSO)~%
\cite{Ulinski2018} framework enhanced with samples archive (M-GAPSO).
The framework is based on the PSO algorithm, but allows the usage of virtually any other optimization method whose ``particles'' act independently. Additionally, the algorithm's restart
manager, the archive (external memory) of samples and the algorithm's search space manager are
defined as auxiliary modules independent from the core optimization module.
Algorithm~\ref{alg:GAPSO} presents a high--level pseudocode of M-GAPSO.


The underlying features of the M-GAPSO design can be listed as follows:
\begin{itemize}
	\item it relies on well-researched optimization algorithms,
	\item effectively utilizes samples already gathered by means of randomized search procedures (i.e. PSO, DE),
	\item guides the search process of the algorithm on the basis of already found local optima,
	\item keeps independence of the constituting modules to the highest possible extent.
\end{itemize}

\begin{algorithm}[t]
	\begin{algorithmic}[1]
	\footnotesize
  \State $F$ is optimized $\mathbb{R}^n \rightarrow \mathbb{R}$ function, $Bounds$ is an $\left(\mathbb{R}^2\right)^n$ vector
  \State $Swarm$ is a set of PSO particles, $Behavior$ is particle's velocity update rule
  \State $Initializer$ is particle's initial location sampler
  \State $SamplesArchive$ is an RTree based samples' index
  \State $BehaviorAdapter$ collects optimum value improvement data
  \State $RestartManager$ observes swarm state and performance
  \State $Bounds \gets f.getBounds()$ \Comment {Initially the whole area is considered}
  \State $PerformanceMonitor.behaviourProbabilities \gets initialProbabilities$
  \State $LocalOptima \gets \emptyset$ \Comment {Set of optima estimations}
  \While {Stopping criterion not met}
	  \For {$Particle \in Swarm$}
		  \State $Particle.x \gets Initializer.nextSample(Bounds)$
		  \State $Sample \gets F.evaluate(Particle.x)$
		  \State $BehaviorAdapter.registerValue(Sample)$
	  \EndFor
	  \For {$Particle \in Swarm$}
		  \State $Particle.v \gets \dfrac{(Particle_i.x - Particle_j.x)}{2.0}$
	  \EndFor
	  \While {$RestartManager.shouldOptimizationContinue(Swarm)$}
		  \For {$Particle \in Swarm$}
			  \State $Behaviour \gets BehaviorAdapter.sampleBehaviourPool()$ \Comment {Mixing behaviors \label{line:alg:GAPSO:mixing}}
			  \State $Particle.v \gets Behavior.computeVelocity(Particle,Swarm,Archive)$
			  \State $Particle.x \gets Particle.x + Particle.v$
			  \If {$SamplesArchive.stored(Particle.x)$}
				  \State $Sample \gets SamplesArchive.retrieve(Particle.x)$
			  \Else
				  \State $Sample \gets F.evaluate(Particle.x)$
				  \State $SamplesArchive.store(Sample)$
			  \EndIf
			  \State $BehaviorAdapter.registerImprovement(Sample,Behaviour)$
		  \EndFor
		  \State $BehaviorAdapter.recomputeBehaviourProbabilities()$
	  \EndWhile
	  \State $LocalOptima \gets LocalOptima \cup Swarm.bestSample$
	  \State $Bounds \gets Initializer(LocalOptima)$ \Comment {Guiding search process}
  \EndWhile
  \caption{M-GAPSO high--level pseudocode%
  \label{alg:GAPSO}}
  \end{algorithmic}
  \end{algorithm}
  
\begin{figure}[!ht]
	\includegraphics[width=\textwidth]{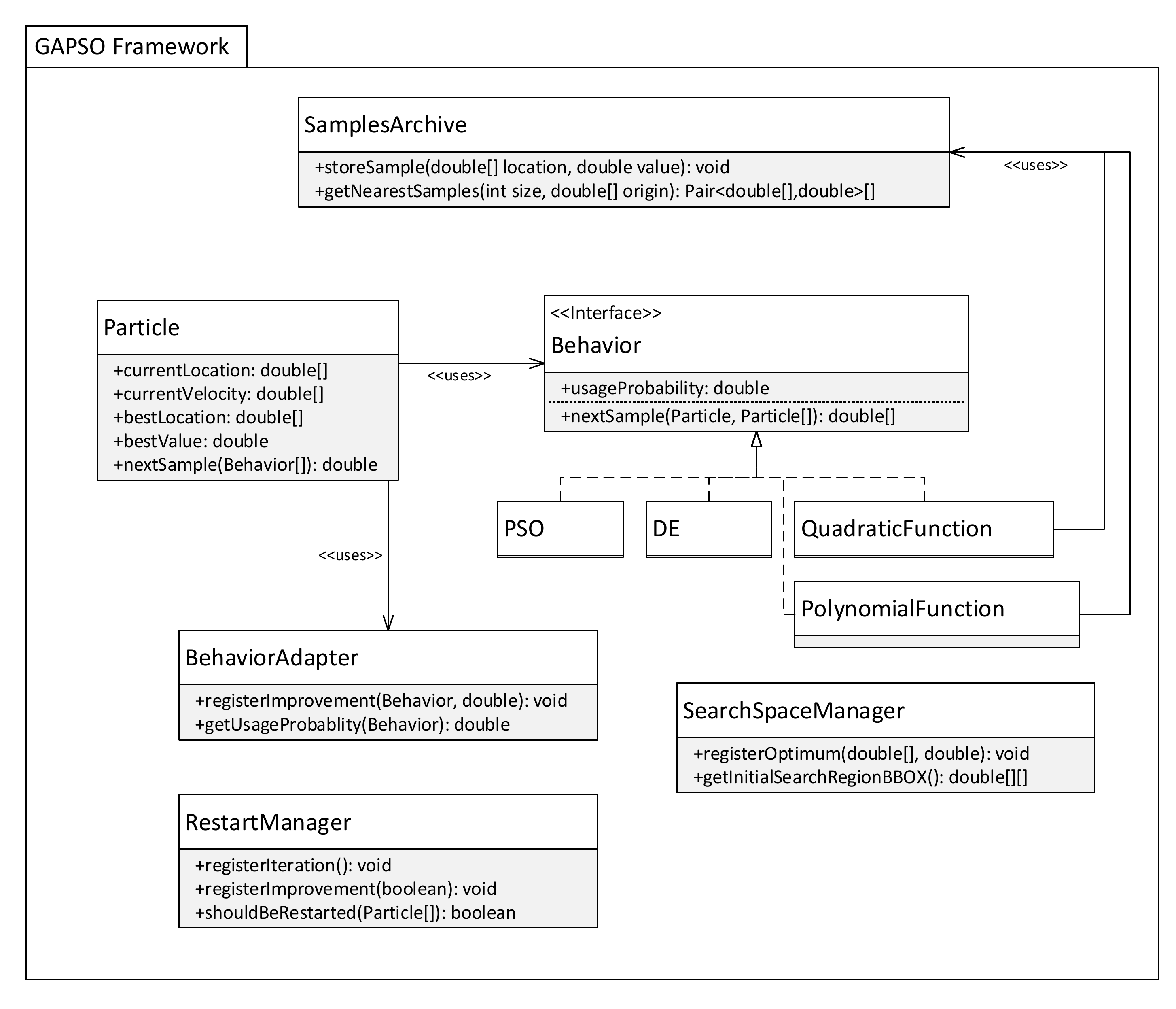}
	\caption{The UML class diagram of M-GAPSO.
	\label{fig:gapso-classes}}
\end{figure}

The main components of M-GAPSO are presented in the UML class diagram (Fig.~\ref{fig:gapso-classes}) and discussed in detail in the following subsections.

\subsection{General swarm--based optimization framework}

A starting point for the GAPSO design was a multi-agent view of the PSO algorithm~\cite{Okulewicz2016}. In PSO, the particles can be seen as independent optimization agents, each exposing its historically best location only and maintaining its own current location and velocity.
This view led to the most important design choice:
a separation of particles' locations from their velocity update formula.
Subsequently, regarding velocity update just as a method of specifying the next location
to be sampled by the algorithm, allowed utilization of any
population-based self-governing optimization algorithm in GAPSO.
Apart from various swarm-based approaches, it also meant the possibility of including other than PSO-like sampling equations (e.g. DE) in GAPSO framework~\cite{Ulinski2018}.

On top of the separation of locations from a sampling behavior, the performance of each of the considered types of behavior is registered and used to update the respective probability of selecting a given behavior
as a sampling mechanism.

M-GAPSO extends the above-described version of GAPSO with three more components,
that can be implemented and configured independently from the core optimization algorithm:
\begin{itemize}
\item samples archive - serving as a cache and a source of additional information about optimized function,
\item population state observer - serving as an algorithm restart manager,
\item high-level manager of the problem space exploration.
\end{itemize}
Additionally, the optimization behavior pool has been extended by adding local search approaches based on
fitting quadratic and polynomial functions.
Due to that change, Variable Neighborhood Search \cite{mladenovic1997variable}, utilized as the local search procedure at the end of the optimization process in the original GAPSO, has been removed.

The purpose of the samples archive is to enable efficient utilization
of function values gathered during the solution search process.

The restart manager together with the search space manager guide the optimization algorithm towards promising areas of the problem space,
when further exploitation of the currently searched area seems to be pointless.
With such a design it is possible to extend the GAPSO framework
and test the impact of optimization improvements somewhat independently of the main optimization engine.

On a final note, GAPSO is maintained as an open source application
on the MIT license and is available at:\\
{\footnotesize
\url{https://bitbucket.org/pl-edu-pw-mini-optimization/basic-pso-de-hybrid/}}.

\subsection{Samples archive}

In order to store and efficiently retrieve samples, M-GAPSO utilizes
a multi-dimensional R-Tree index.
After a series of preliminary experiments, a maximum capacity of samples index was capped at $20\;000$, as in some cases it grew too large to efficiently handle the samples.
After reaching the maximum capacity, the index is restarted from scratch.

\begin{figure}[t]
	\includegraphics[width=0.48\textwidth]{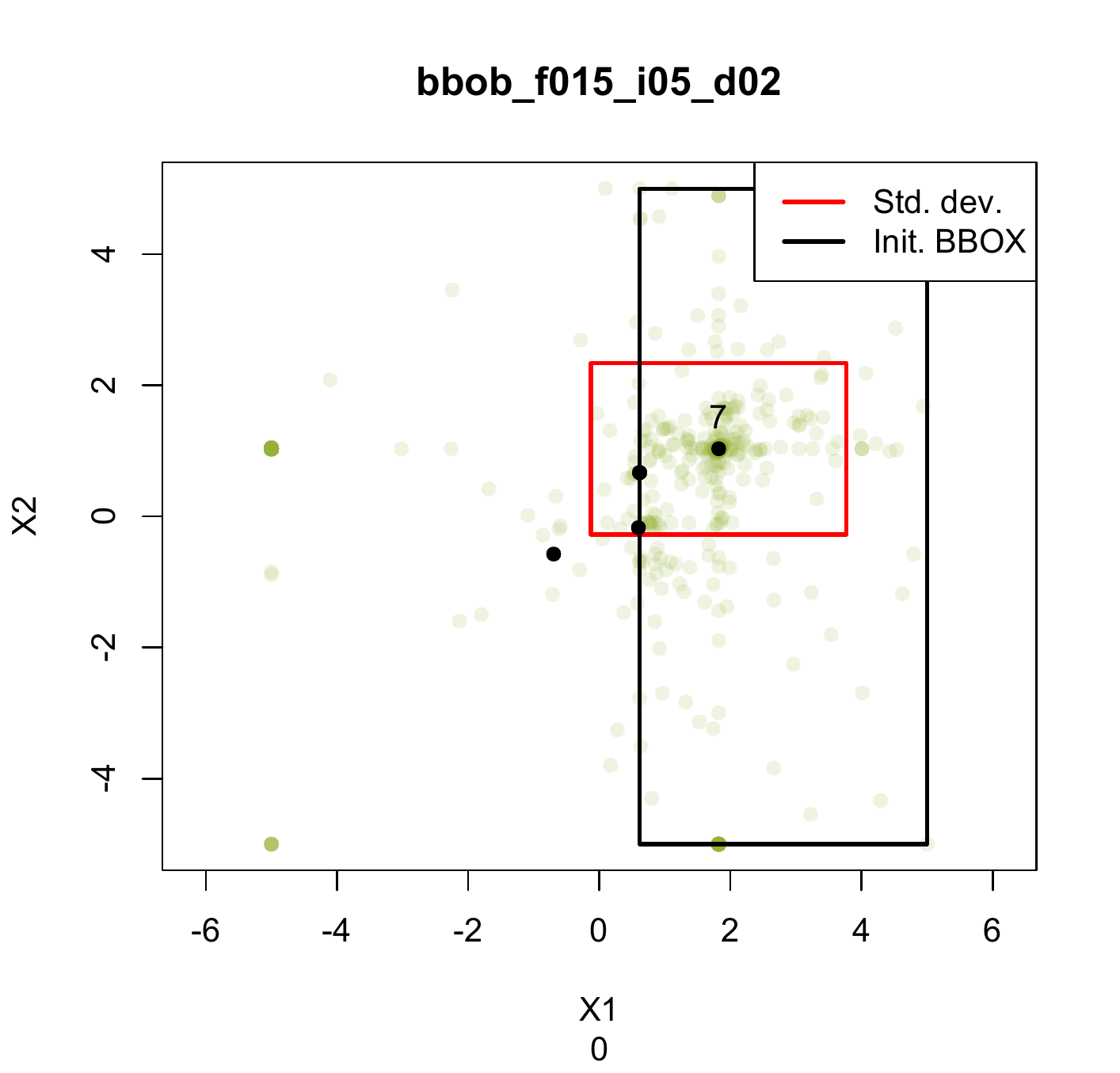}\hfill
	\includegraphics[width=0.48\textwidth]{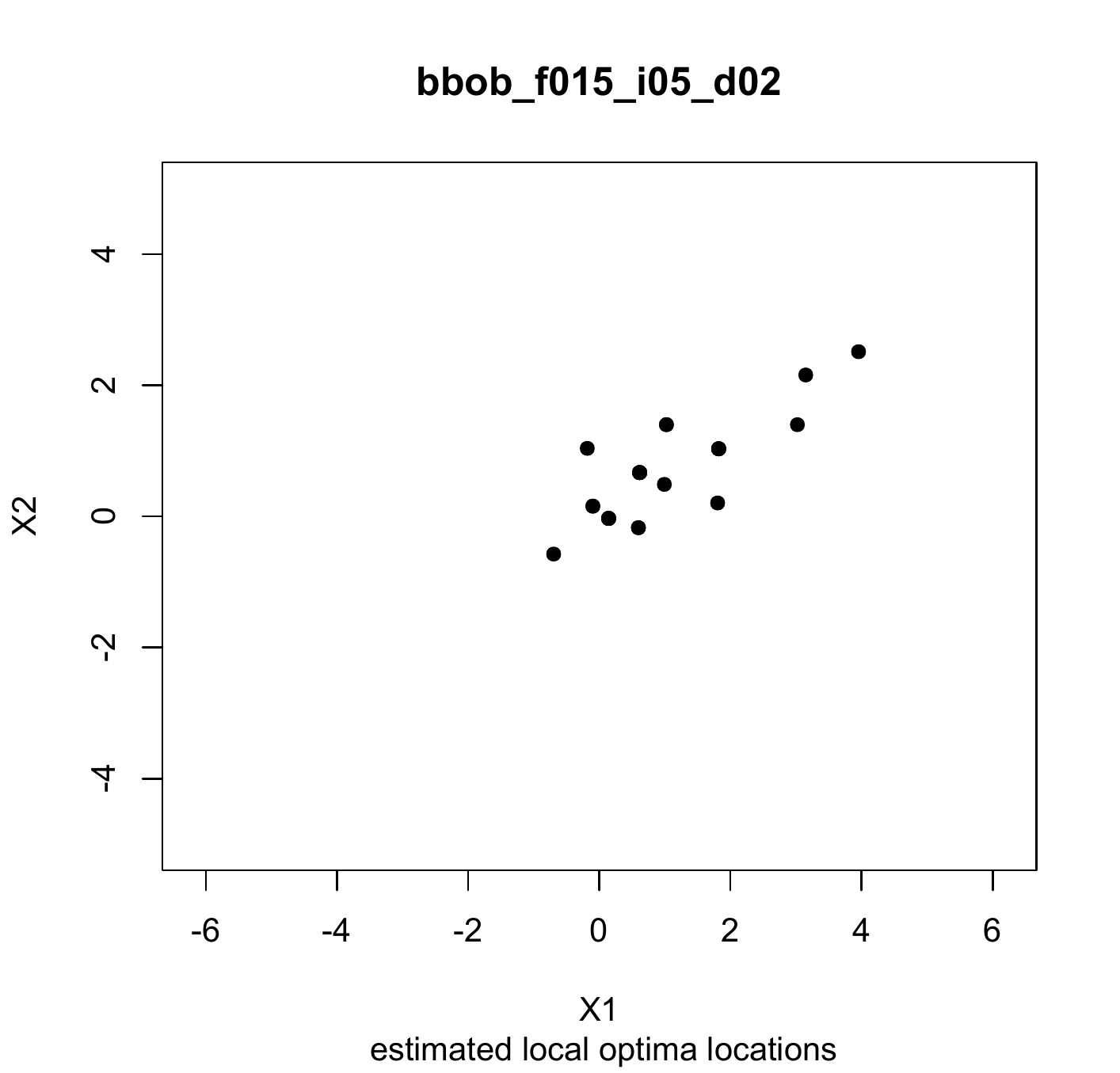}\hfill
	\caption{M-GAPSO outer loop sampling scheme.
	Black dots mark locations of local optima estimations,
	a black rectangle marks the area for initial swarm location in a given run,
	and a red rectangle marks the resulting distribution of samples in that run.
	\label{fig:algorithm-run}}
\end{figure}

\subsection{Restart management}
\label{sec:resetart-management}

A current version of GAPSO uses an enhanced version of JADE \cite{Poaik2012} restart manager.
In a basic JADE approach a spread of population locations was considered and combined with
the frequency of global optimum estimation improvements.
In M-GAPSO the $RestartManager$ registers iteration count intervals between global optimum updates,
considers a spread of personal best locations of particles
and additionally a spread of personal best locations values.
The last feature was added in order to better handle step functions,
where the population spread can be quite large,
even though the population reached a sort of a frozen state,
with each particle having exactly the same personal best value.

\subsection{Initialization scheme}
\label{sec:initialization-scheme}

\begin{figure}[t]
	\includegraphics[width=0.48\textwidth]{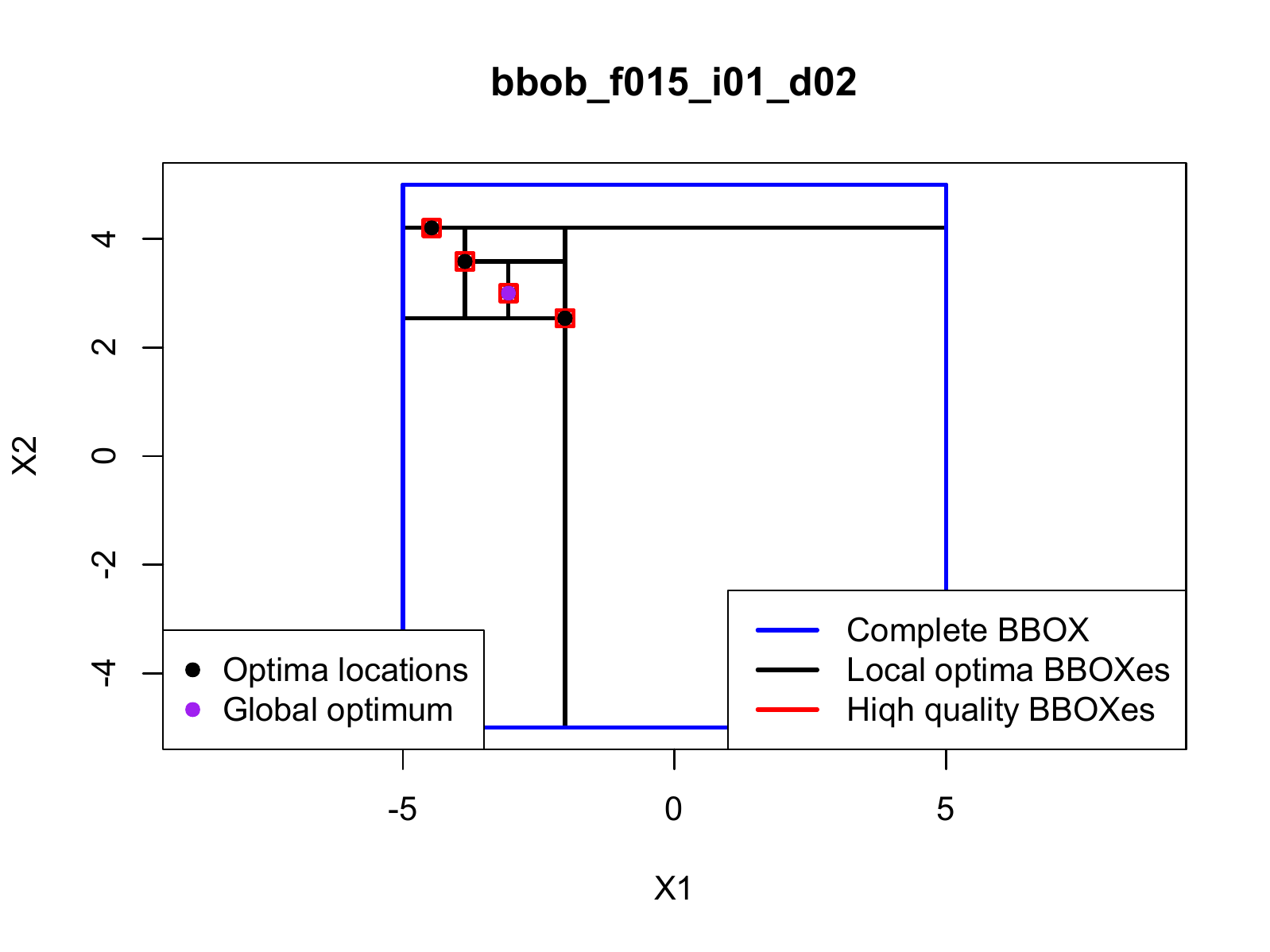}\hfill
	\includegraphics[width=0.48\textwidth]{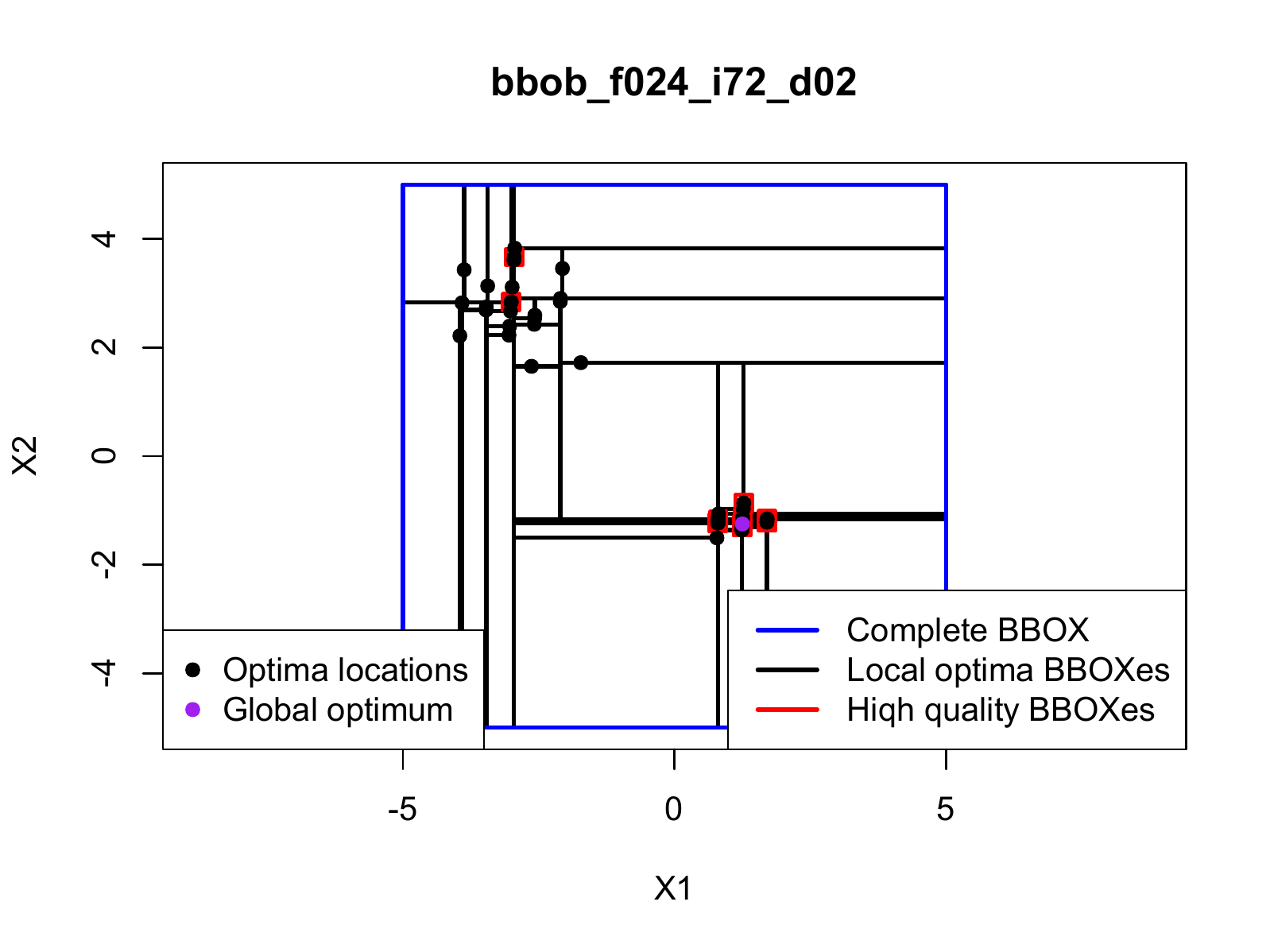}
	\caption{Possible M-GAPSO initialization bounding boxes
	generated during the optimization process for the functions with (\textit{f15})
	and without (\textit{f24}) a general structure.
	\label{fig:algorithm-bboxes}}
\end{figure}

An initialization scheme relies on selecting a smaller bounding box
as the initial search area on the basis of previously estimated function optima.
Figure~\ref{fig:algorithm-run} presents a sample step of M-GAPSO re-initialization
and multiple local optima estimations as the final result of such a procedure.

The initialization procedure includes the following possibilities:
\begin{itemize}
	\item starting from the original (full) bounding box
	defined for the optimized function,
	\item starting from a random bounding box defined by boundaries derived from
	locations of the local optima estimations,
	\item starting from a small bounding box centered around the high-quality
	optimum estimation.
\end{itemize}
One of those actions is selected at random with probabilities
set by the user of M-GAPSO.
Figure~\ref{fig:algorithm-bboxes} visualizes possible bounding boxes
for initializing the M-GAPSO population, created on the basis of optima estimations during the previous optimization process.

\subsection{Optimization behaviors}

One of the main conclusions from our initial work on GAPSO \cite{Ulinski2018} was that the highest synergy among optimization behaviors could be observed when DE and PSO are utilized, instead of a pool of various PSO variants (i.e. Standard PSO, Charged PSO, Fully-Informed PSO).
Therefore, the behavior pool considered in M-GAPSO consists of SPSO-2007 \cite{PSO:Introduction,clerc2012standard}, DE/best/1/bin \cite{DE} (previously used in~\cite{Ulinski2018})
and the local modeling technique (proposed in this paper) that employs quadratic and polynomial functions.

\subsubsection{PSO and DE behaviors}
The utilization of the SPSO-2007 within M-GAPSO framework is straightforward,
as the framework is based on that algorithm. Addition of DE behavior required a slight
adjustment, as original DE does not include a notions of individual's velocity
and current location.
Therefore, when DE behavior is applied, the particle's velocity is computed as follows:
\begin{equation}
	particle.v_{t+1} \gets x_{DE.sample} - particle.x_{t}
\end{equation}
The same approach is utilized in implementation of any other behavior not supporting the idea of velocity.
One of the consequences of the above assumption is the loss of previously computed velocity, resulting in its reset.
{However, the new velocity still makes sense from the point of view of SPSO-2007 behavior.
Please observe that, if a particle's move is successful (i.e. $particle.x_{best}$ is updated),
this particle (when treated with the PSO behavior) will continue to move roughly in the direction that already improved its value.
This direction might be perturbed only by the attraction vector of the best neighbor location.}

\subsubsection{Quadratic and polynomial model-based behaviors}
\label{sec:function-models}

\begin{figure}
	\begin{center}%
	\includegraphics[width=0.48\textwidth]{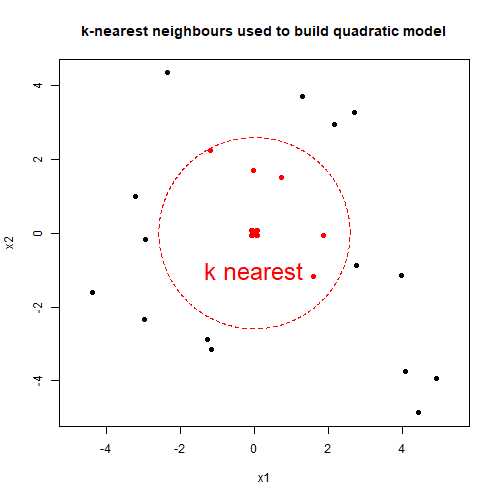}\hfill
	\includegraphics[width=0.48\textwidth]{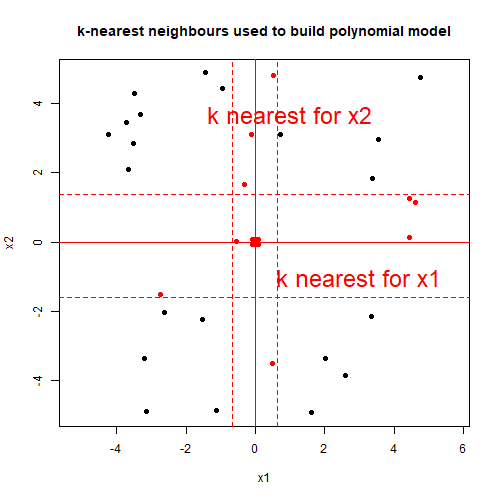}%
	\end{center}%
	\caption{Comparison of data sets of samples used for fitting quadratic and polynomial models.
	\label{fig:knn-data-sets}}
\end{figure}

In order to support fast convergence to local optima, the pool of behaviors was supplemented by model-based approaches~\cite{ZaborskiOkulewiczMandziuk2019}, which utilize samples already gathered in randomized behaviors (such as PSO and DE).
For all model behaviors a sample archive uses an R-Tree data structure which proven efficient in handling these samples.

\textbf{Quadratic model} is fitted on a data set made of $k$ samples nearest (in the Euclidean metric) to $particle.x_{best}$ location of the particle for which the quadratic behavior has been selected (see Fig.~\ref{fig:knn-data-sets} for an example).
Quadratic function based approach fits the following model:
\begin{equation}
\hat{f}_{quadratic.local}(x) = \sum\limits_{d=1}^{dim}\left(a_dx_d^2+b_dx_d\right) + c
\end{equation}
where
$x=[x_1, x_2, \ldots, x_{dim}]$ is a vector of coordinates.
	
Ordinary Least Squares method is applied to obtain linear regression coefficients:
$[a_1, a_2, \ldots, a_{dim}]$ and $[b_1, b_2, \ldots, b_{dim}]$.
Finding minimum of $\hat{f}_{quadratic.local}(x)$ is equivalent to finding $dim$ independent minima of $dim$ independent parabolas $a_dx_d^2+b_dx_d$.
Coordinates of the bounded parabola peak are computed in the following way:
\begin{equation}
    x^{bounded.parabola.peak}_d =
\begin{cases}
    -\dfrac{b_d}{2a_d},& \text{if } a_d > 0\,\wedge\,-\dfrac{b_d}{2a_d}\in [\gamma^{lower}_d, \gamma^{upper}_d]\\
    \gamma^{lower}_d,              & \hat{f}(\gamma^{lower}_d) \leq  \hat{f}(\gamma^{upper}_i)\\
    \gamma^{upper}_d,              & \hat{f}(\gamma^{lower}_d) >  \hat{f}(\gamma^{upper}_d)
\end{cases}
\end{equation}
where $\gamma^{lower}_d$ and $\gamma^{upper}_d$ are lower and upper bounds of the function domain, respectively.

\textbf{Polynomial model} enhances the quadratic model in the following way:
\begin{equation}
\hat{f}_{polynomial.local}(x_d) = \sum\limits_{i=1}^{p}a_{i,d}x_d^i + c
\label{eq:poly-val}
\end{equation}
where $ x_d$ is coordinate in a given dimension.

The polynomial model is fitted separately in each dimension. For each dimension $d$, the fitting data set is composed of $k$ samples nearest to a line with coordinates fixed to the
current location except for dimension $d$. The differences between methods of gathering samples in both models (quadratic and polynomial) are depicted in Fig.~\ref{fig:knn-data-sets}.

A coordinate of the minimum of the bounded polynomial is computed using grid search.
For each dimension $d$, a space between $\zeta^{lower}_d$ and $\zeta^{upper}_d$ is divided into $1000$ regular points, where
$\zeta^{lower}_d$ and $\zeta^{upper}_d$ are respectively minimum and maximum coordinates in dimension $d$ derived from the fitting data set.
A function value (\ref{eq:poly-val}) is calculated in each of these $1000$ points and the smallest one determines the optimal coordinate $\hat{x_d}$.
A vector of all $\hat{x_d}$ (for all dimensions) defines the next sampling point.

Both models are utilized as optimization behaviors in a similar way to DE sampling:
\begin{equation}
	particle.v_{t+1} \gets x_{bounded.function.peak} - particle.x_{t}
\end{equation}

\subsection{Adaptation scheme}

\begin{figure}[t]
	\includegraphics[width=\textwidth]{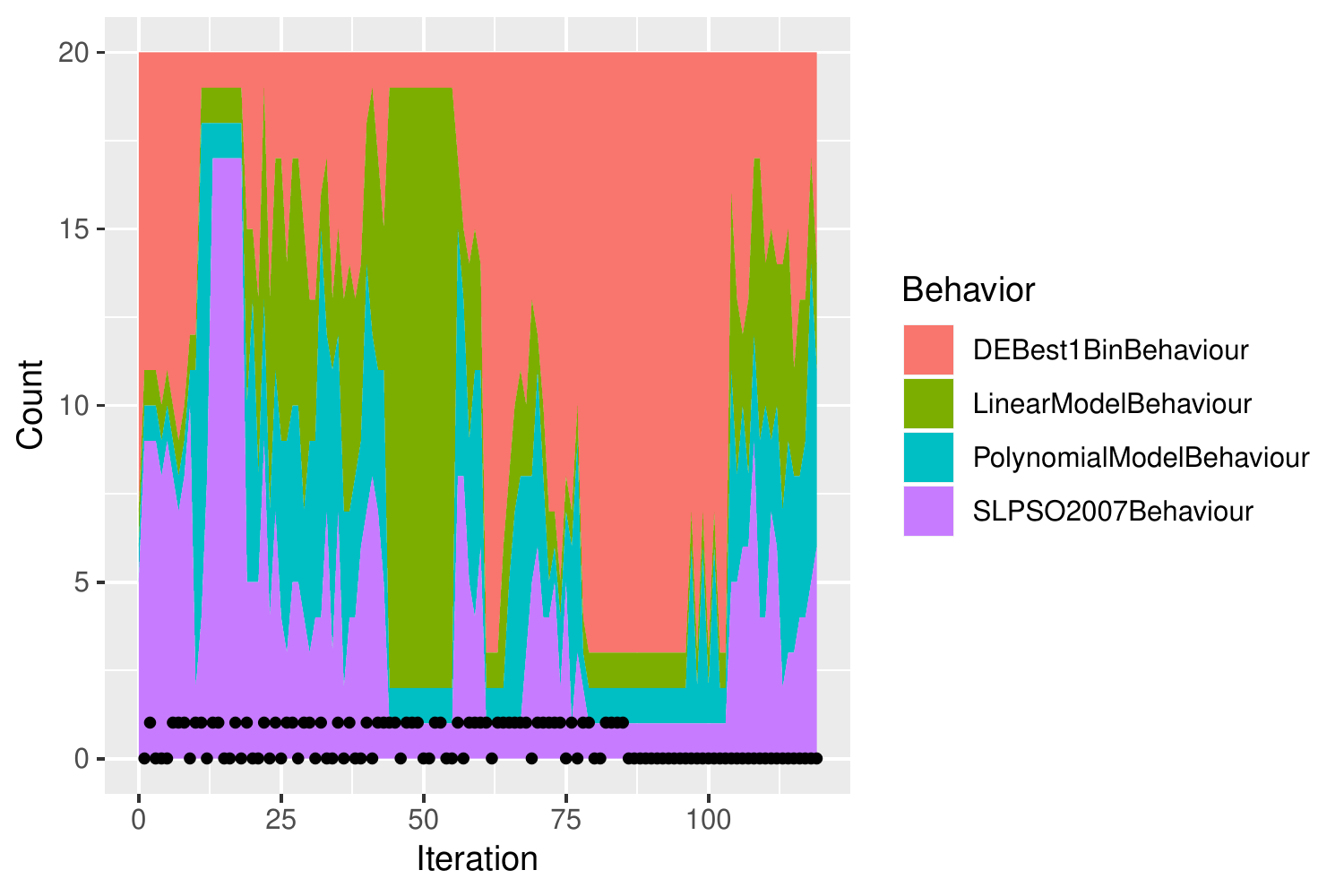}
	\caption{Fractions of using various behaviors during a single algorithm's run (i.e. till a restart).
	Black dots denote whether in a given iteration a global optimum value was improved.
	\label{fig:local-adaptation}}
\end{figure}

Adaptation scheme in M-GAPSO, for each optimization behavior takes into account its contributions to the improvement of the \textbf{global} best value.
These contributions are aggregated by a moving average scheme (separately for each optimization behavior) and normalized in order to account for the number of times a given behavior was applied.
This procedure is different than GAPSO adaptation scheme, which considered the average improvement of the \textbf{local} optimum estimation
(i.e. against the former value of $f(particle.x_{best})$ for each of the particles).

Figure~\ref{fig:local-adaptation} presents the effects
of applying adaptation scheme in a single run of the algorithm.
In the first few iterations the fractions of using various behaviors oscillate around predefined values set by the algorithm user.
Afterwards the currently best performing behavior starts to dominate over the others in terms of the frequency of its application.
In the final phase, all behaviors have pairwise equal probabilities of their application due to prolonged lack of improvement.
If such a situation lasts for a certain (user defined) number of iterations, the algorithm is reset (cf. Section~\ref{sec:resetart-management}).

\section{Experimental evaluation of M-GAPSO}
\label{sec:performance}

M-GAPSO was tested on the COCO BBOB benchmark set and also took part in the BBComp competition.
\begin{table}[!ht]
	\caption{Settings of the M-GAPSO method
	\label{tab:M-GAPSO-settings}}
	\begin{center}
	\begin{tabular}{lr}
		\multicolumn{2}{c}{\textbf{Core M-GAPSO parameters}} \\
		Population size & $10D$ \\
		Initial PSO behavior weight & 1000 \\
		Initial DE behavior weight & 1000 \\
		Initial quadratic model--based behavior weight & 1 \\
		Initial polynomial model--based behavior weight & 1 \\
		Adaptation module & OFF \\
		\multicolumn{2}{c}{\textbf{Optimization algorithms parameters}} \\
		PSO cognitive factor $c_1$ & 1.4 \\
		PSO social factor $c_2$ & 1.4 \\
		PSO velocity inertia factor $\omega$ & 0.64 \\
		DE cross-over probability & 0.9 \\
		DE mutation scaling factor $F$ & 0.0 - 1.4 \\
		Samples archive size & $2*10^5$ \\
		Quadratic model nearest samples count & $5D$ \\
		Polynomial model degree & 4 \\
		Polynomial model nearest samples count & $4D+1$ \\
	\end{tabular}
\end{center}
\end{table}

\subsection{COCO BBOB benchmark results}

The main assessment of M-GAPSO performance was made on $24$ noiseless continuous functions from COCO BBOB data set.
The algorithm has been run once on $15$ instances for each of the functions, which is a standard procedure
for this benchmark set.
Using this particular benchmark allowed for a straightforward comparison with results obtained by various other optimization algorithms, as the benchmark
comes with a database of results sent for evaluation in BBOB workshops.

The goal of the experiments was to assess the efficacy of various methods
implemented within M-GAPSO framework,
to compare M-GAPSO with its previous version (GAPSO),
and to observe its performance versus state-of-the-art optimization algorithms.
Values of M-GAPSO parameters (presented in Table~\ref{tab:M-GAPSO-settings}) were set partly based on the literature and partly based on results of preliminary tests.

\subsubsection{Improvements over GAPSO and comparison with one of the CMA-ES versions (state-of-the-art)}
\label{sec:external-comparison}

\begin{figure}[t]
	\begin{center}
		\includegraphics[width=0.465\textwidth]%
		{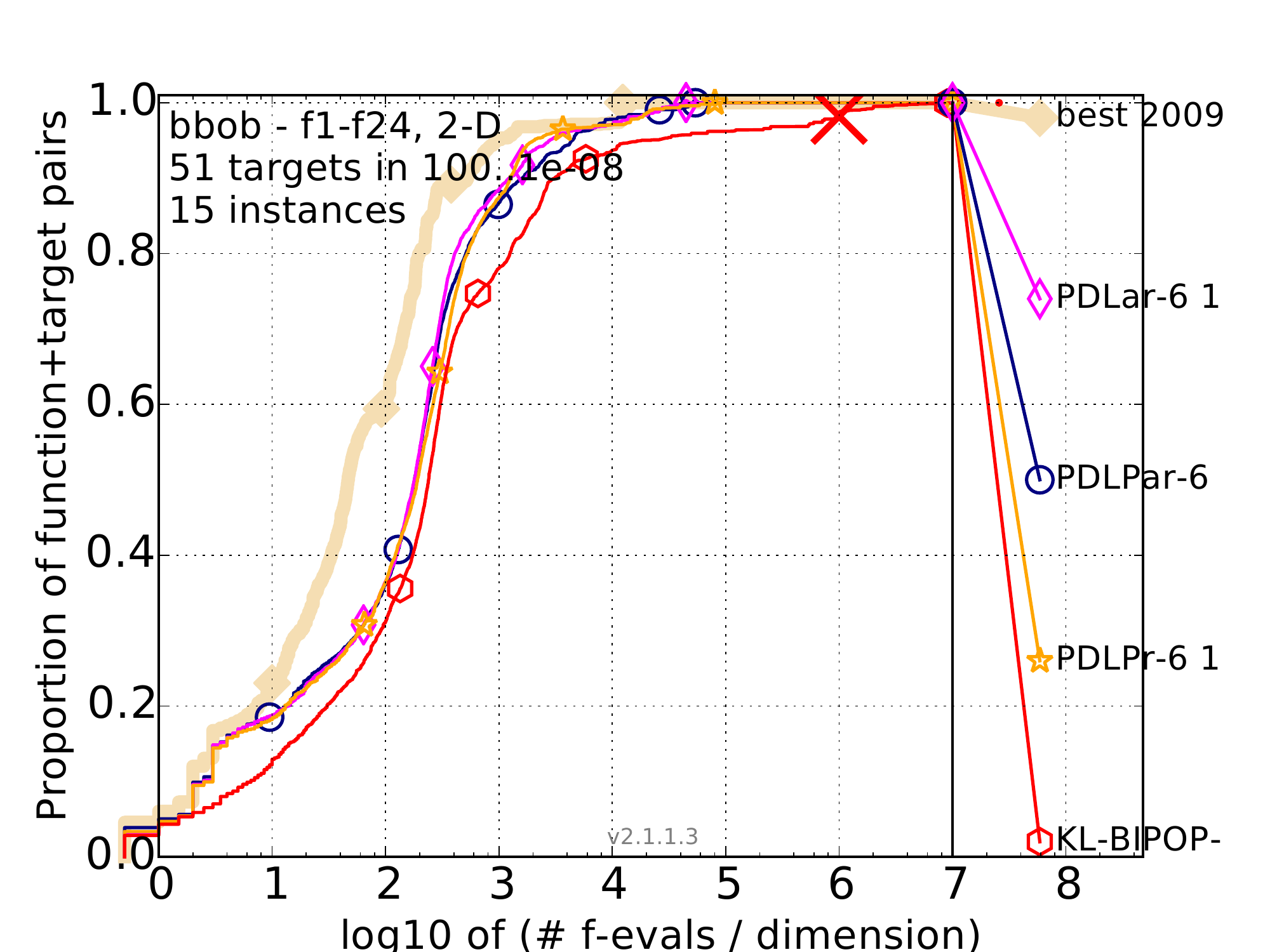}\hfill
		\includegraphics[width=0.465\textwidth]%
		{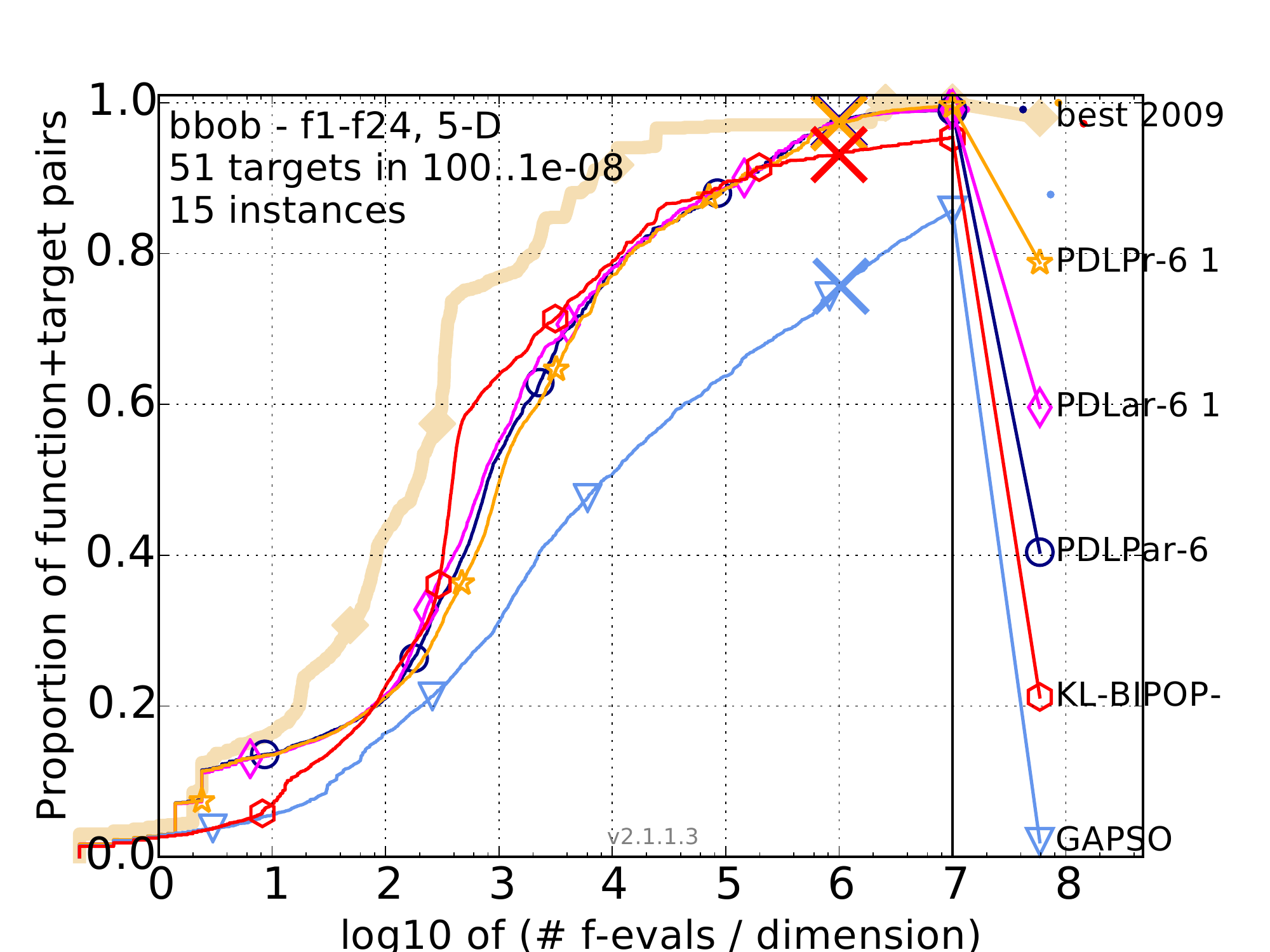}
		\includegraphics[width=0.465\textwidth]%
		{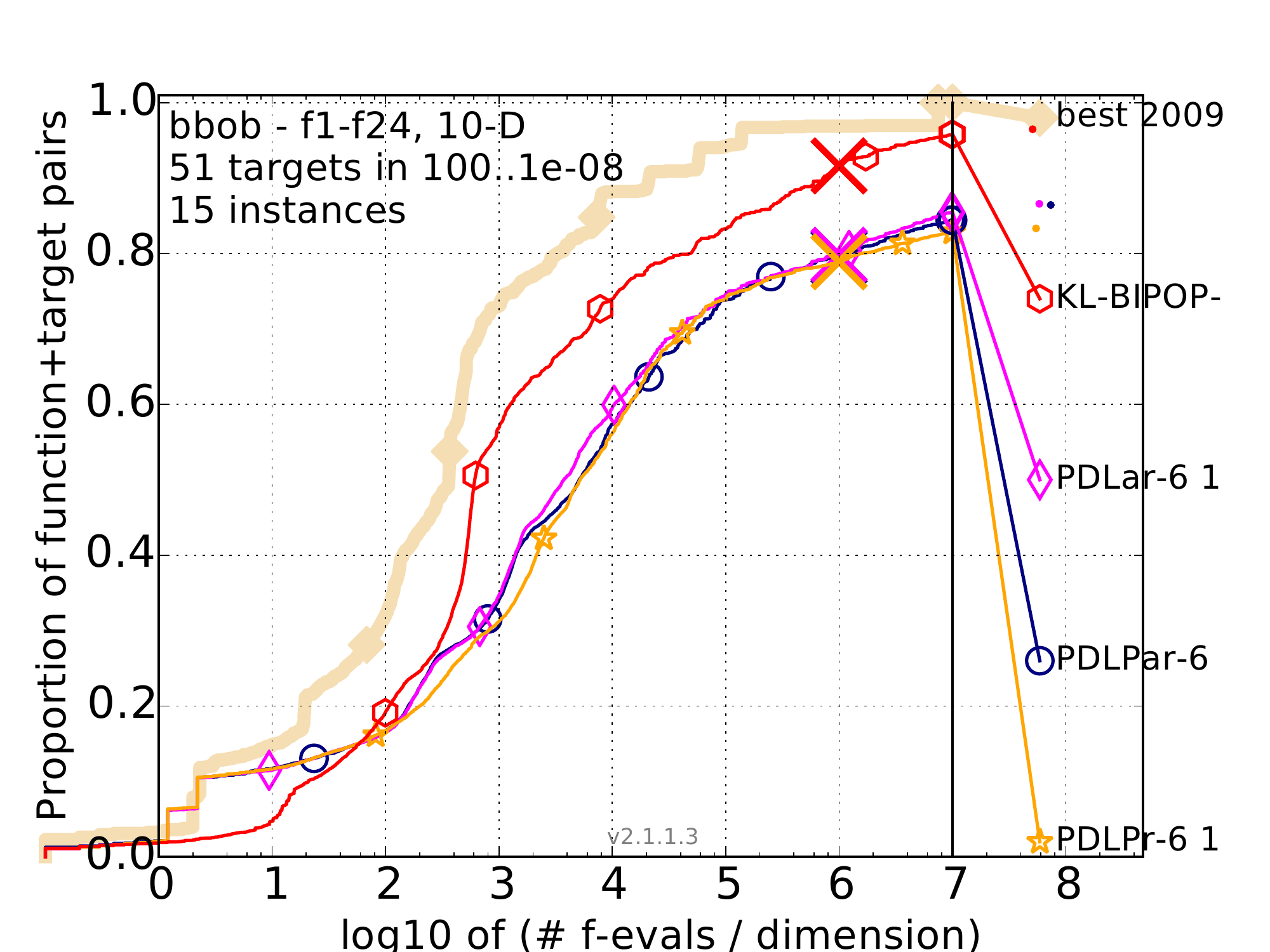}\hfill
		\includegraphics[width=0.465\textwidth]%
		{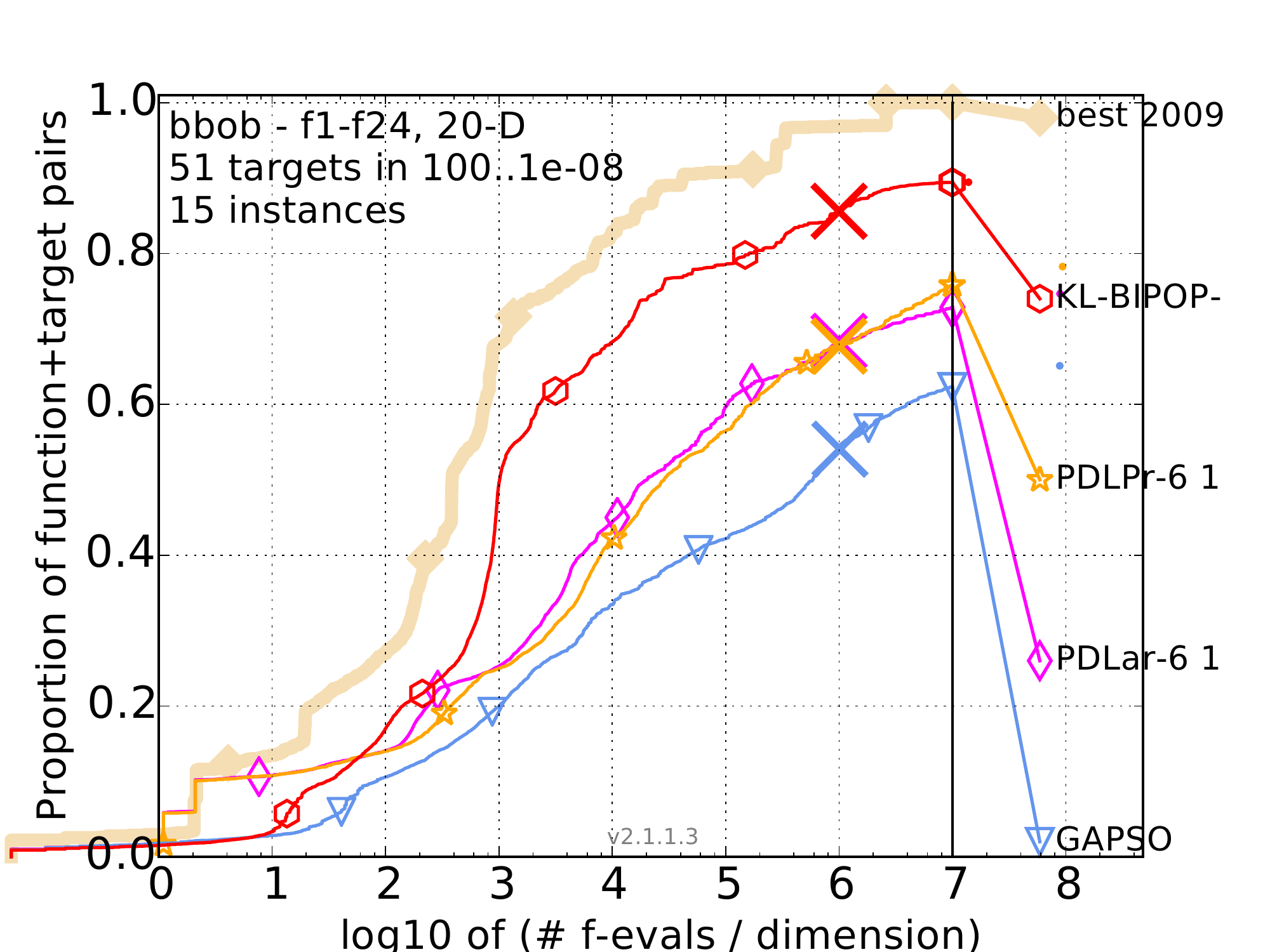}
	\end{center}
		\caption{Results of M-GAPSO versus GAPSO
		and one of the state-of-the-art variants of CMA-ES.
		\label{fig:algorithm-gapso-improvements}}
	\end{figure}

Figure~\ref{fig:algorithm-gapso-improvements} presents results of the experiments with $DIM \times 10^6$ optimization
budget, on 2D, 5D, 10D and 20D benchmark functions.
Three M-GAPSO configurations are compared with GAPSO \cite{Ulinski2018}
and one of the state-of-the-art versions of CMA-ES~\cite{TakahiroYamaguchi2017}.
All three GAPSO configurations utilize behaviors mixing  (cf. line \ref{line:alg:GAPSO:mixing} in pseudocode of Algorithm \ref{alg:GAPSO}), model-based behaviors and improved initialization scheme. \textit{PDLPr} is the fastest approach,
as it does not apply adaptation in any form. \textit{PDLar}
uses a square function model-based behavior only
and includes behaviors' adaptation. \textit{PDLPar}
includes all features discussed in this paper,
but (as presented in Figure~\ref{fig:computations-time}) is quite slow compared to the other two approaches
and was unable to produce results for 20D functions within reasonable amount of time.

\subsubsection{Performance analysis of M-GAPSO components}
The baseline experiment
selects DE/best/1/bin as the sole behavior in M-GAPSO.
This basic configuration is compared with three other approaches, each of them including additional features.
The first enhancement consists in adding local neighborhood SPSO-2007 to behavior pool.
The second one adds model based optimization behaviors (cf. Section~\ref{sec:function-models}).
The final enhancement relies on adding behaviors adaptation and restart management that refers to previously found optima.

\begin{figure}[H]
	\includegraphics[width=\textwidth]{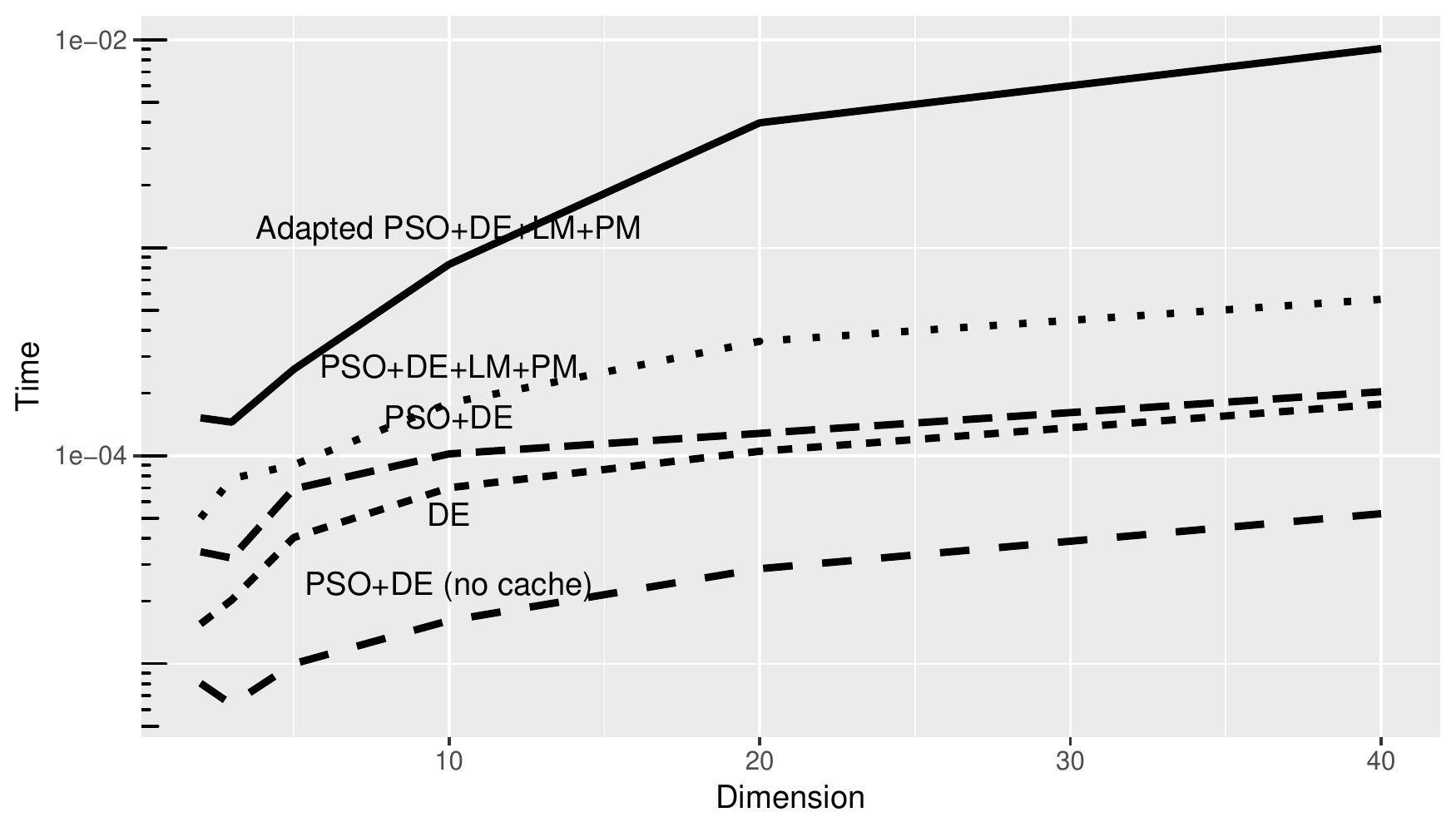}
	\caption{The average function value evaluation time of various M-GAPSO configurations
	with respect to function dimension.
	\label{fig:computations-time}}
	\vspace{-1em}
\end{figure}

\begin{figure}[H]
	\includegraphics[width=0.44\textwidth]{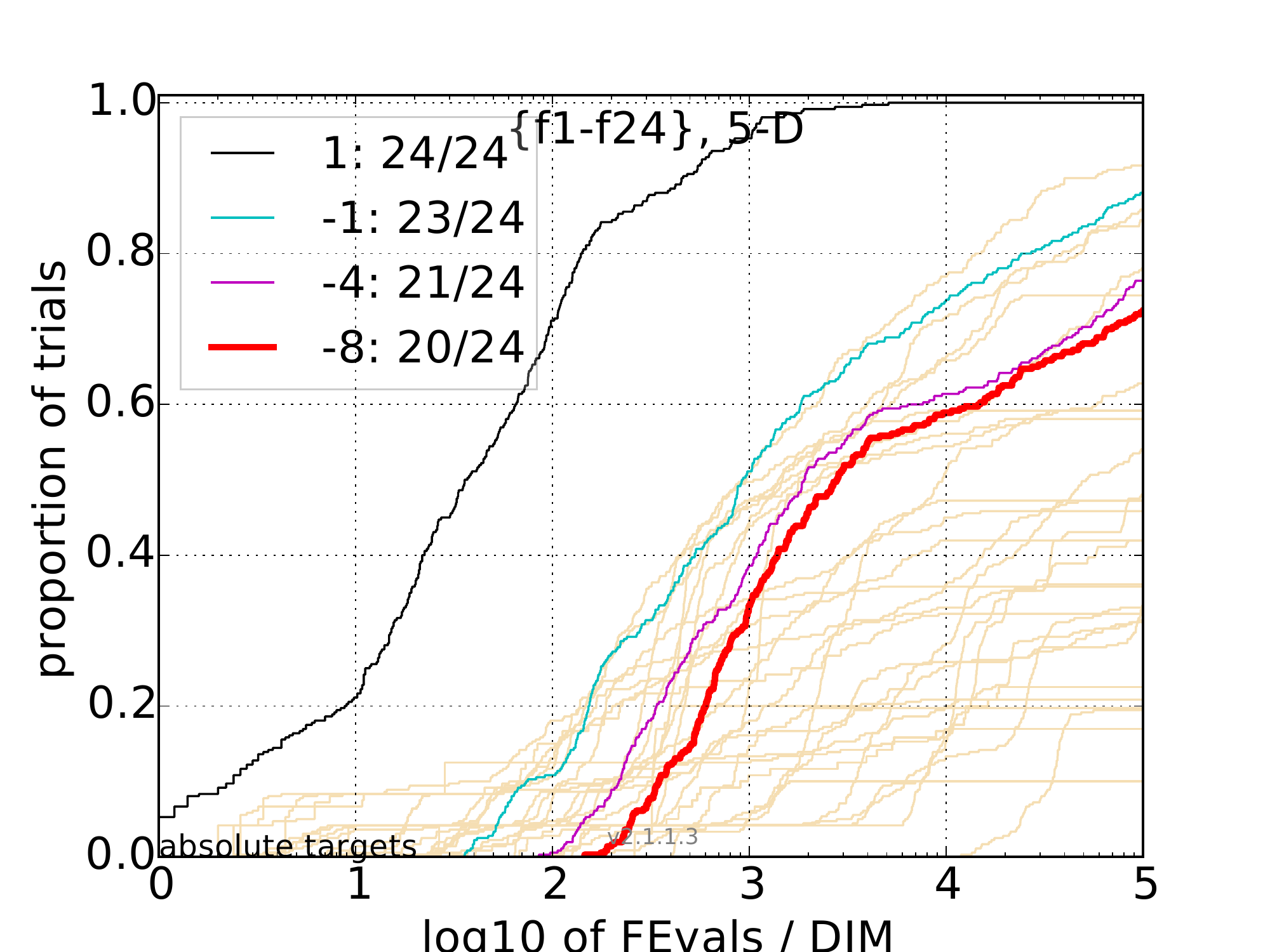}\hfill
	\includegraphics[width=0.44\textwidth]{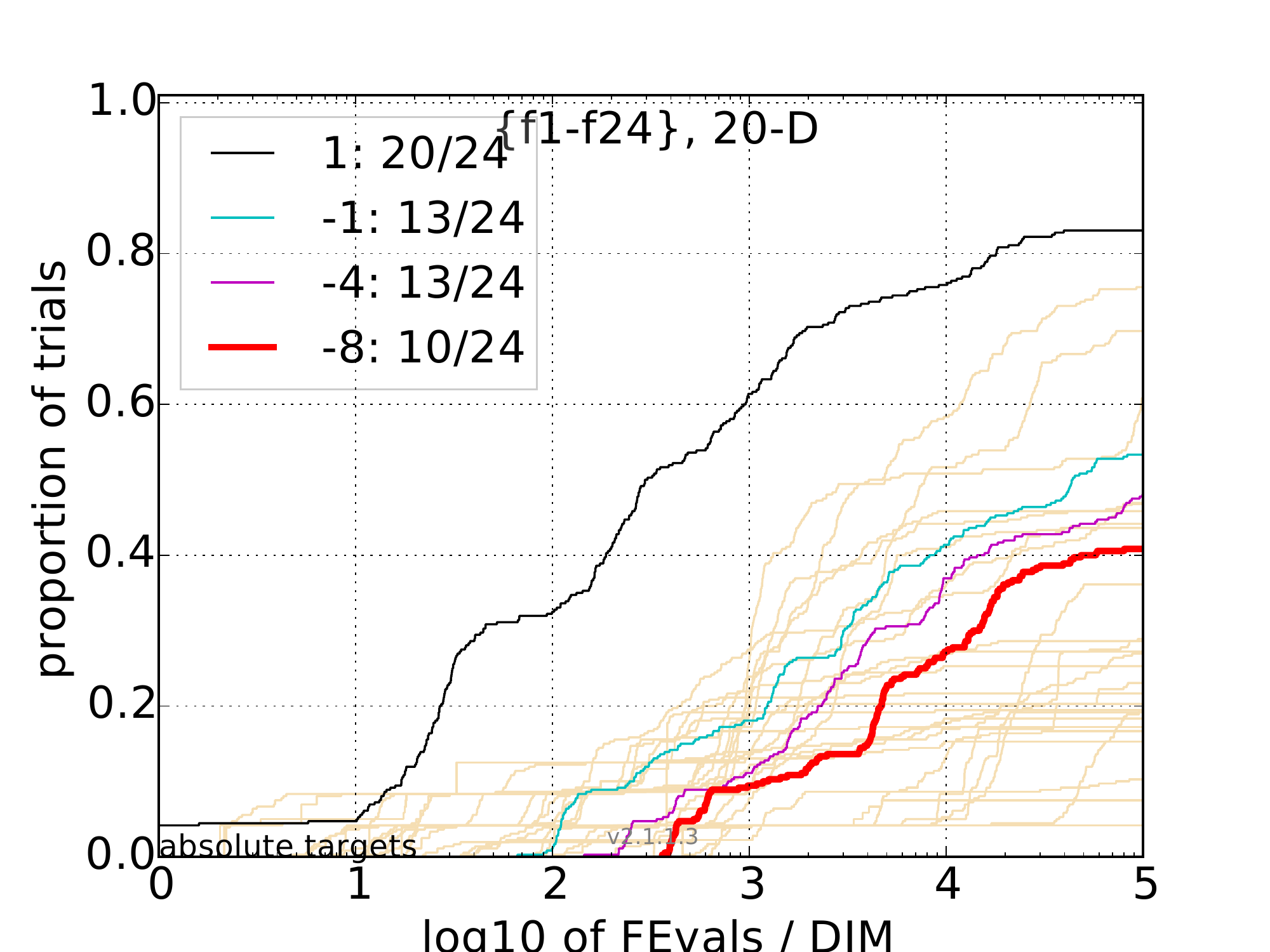}\\
	\vspace{-1.7em}
	\begin{center}%
		DE\\
	\end{center}%
	\vspace{-0.5em}
	\includegraphics[width=0.44\textwidth]{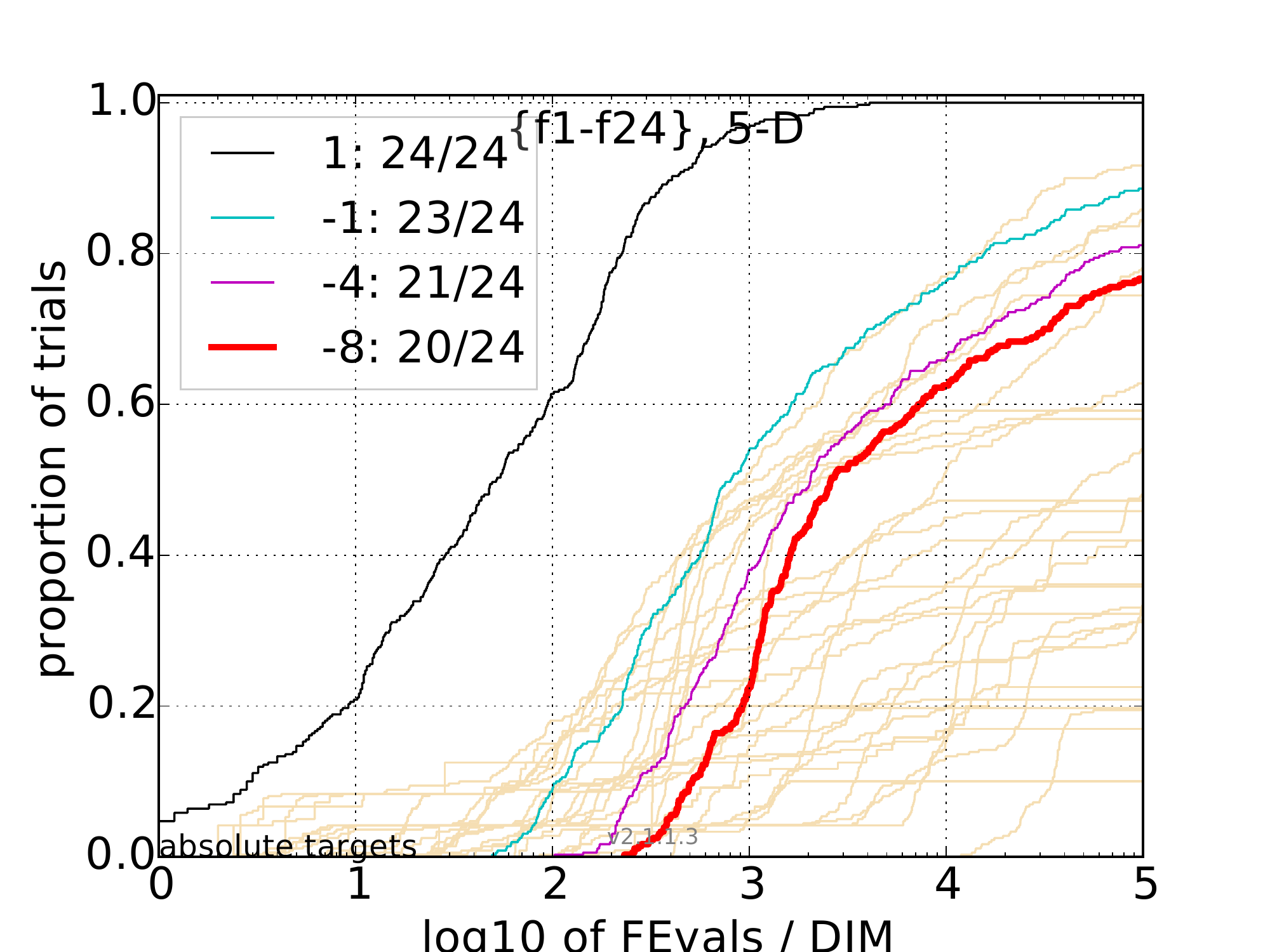}\hfill
	\includegraphics[width=0.44\textwidth]{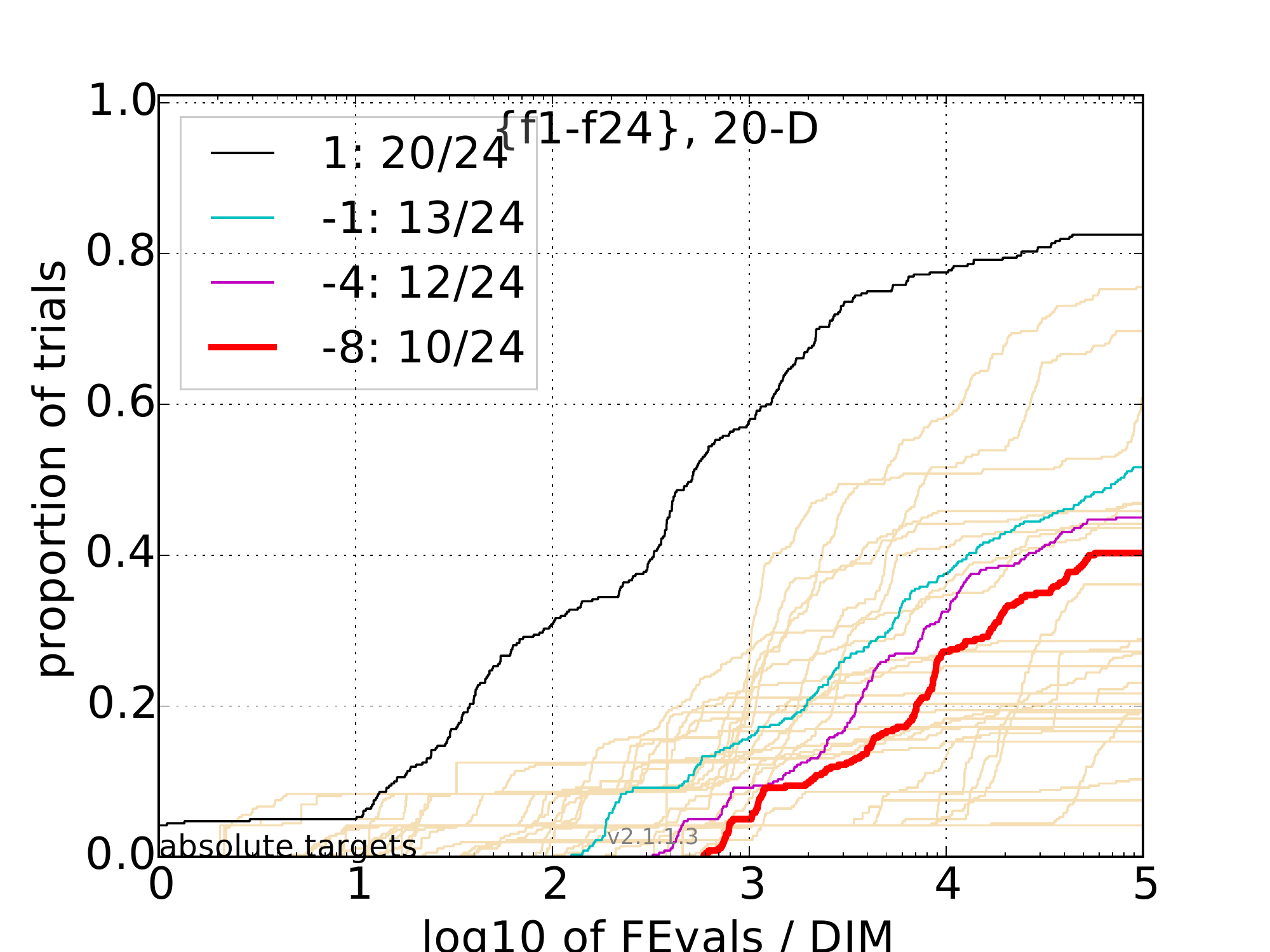}\\
	\vspace{-1.7em}
	\begin{center}%
		PSO+DE\\
	\end{center}%
	\vspace{-0.5em}
	\includegraphics[width=0.44\textwidth]{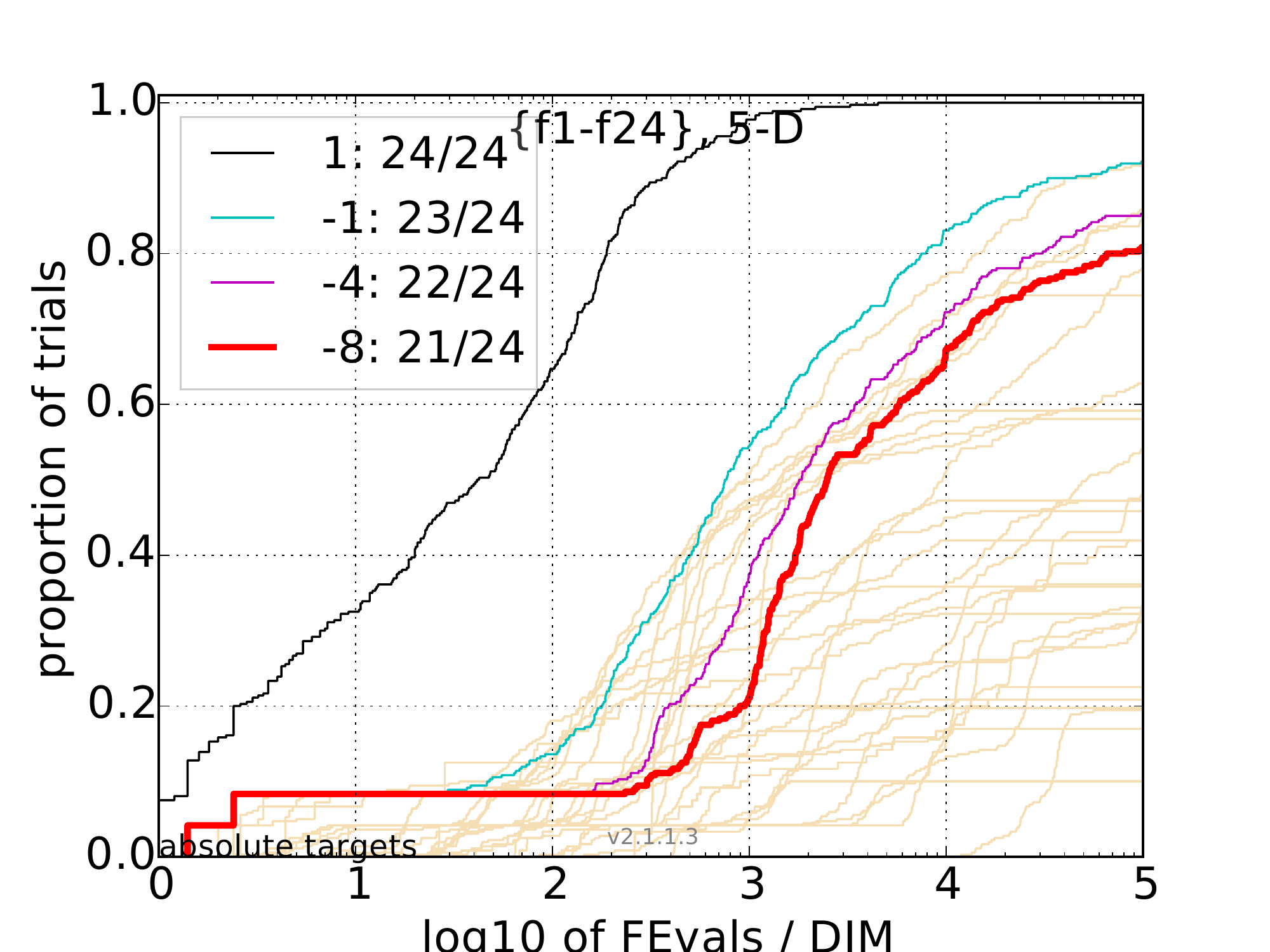}\hfill
	\includegraphics[width=0.44\textwidth]{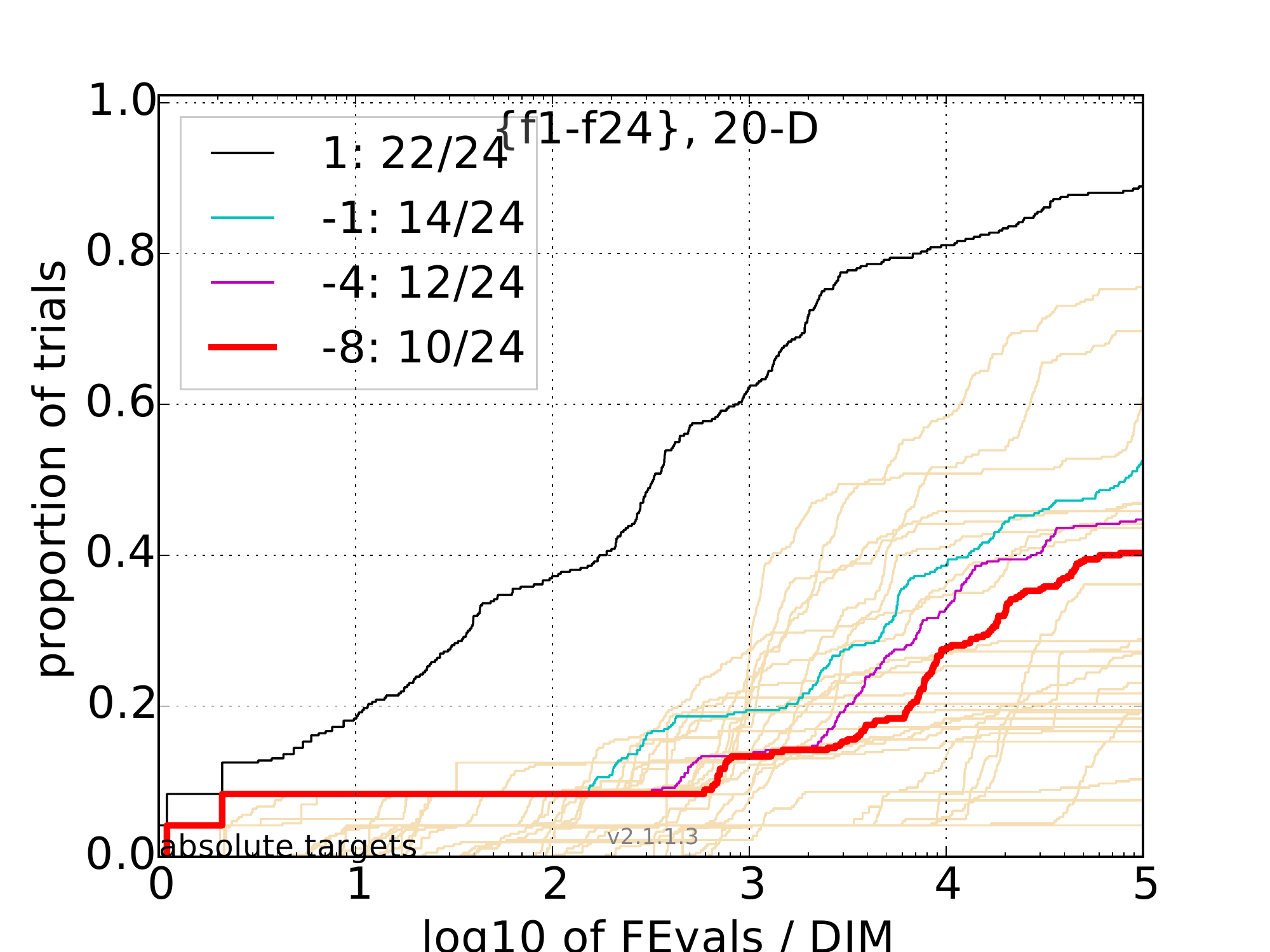}\\
	\vspace{-1.7em}
	\begin{center}%
		PSO+DE+Square function+Polynomial function\\
	\end{center}%
	\vspace{-0.5em}
	\includegraphics[width=0.44\textwidth]{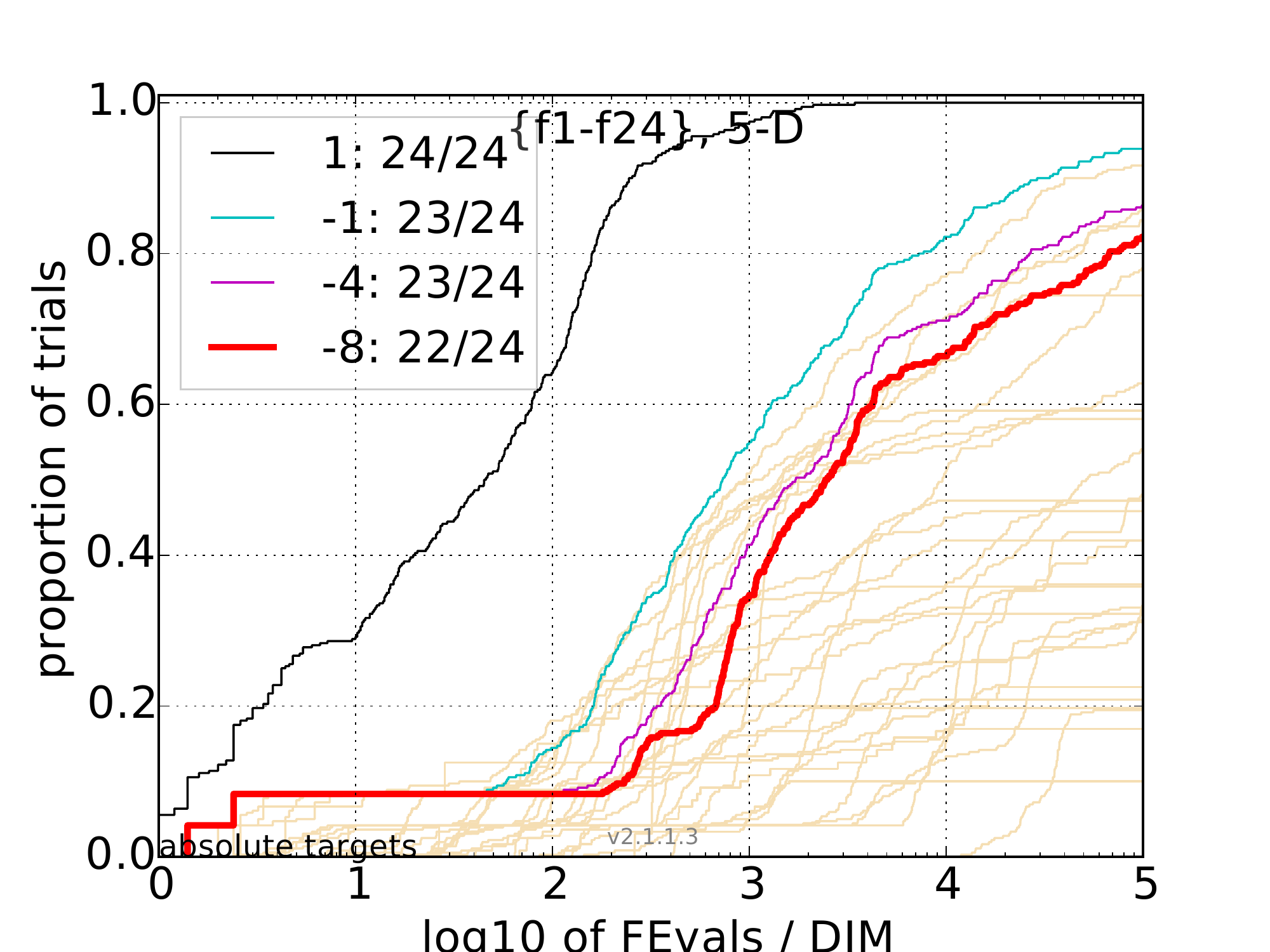}\hfill
	\includegraphics[width=0.44\textwidth]{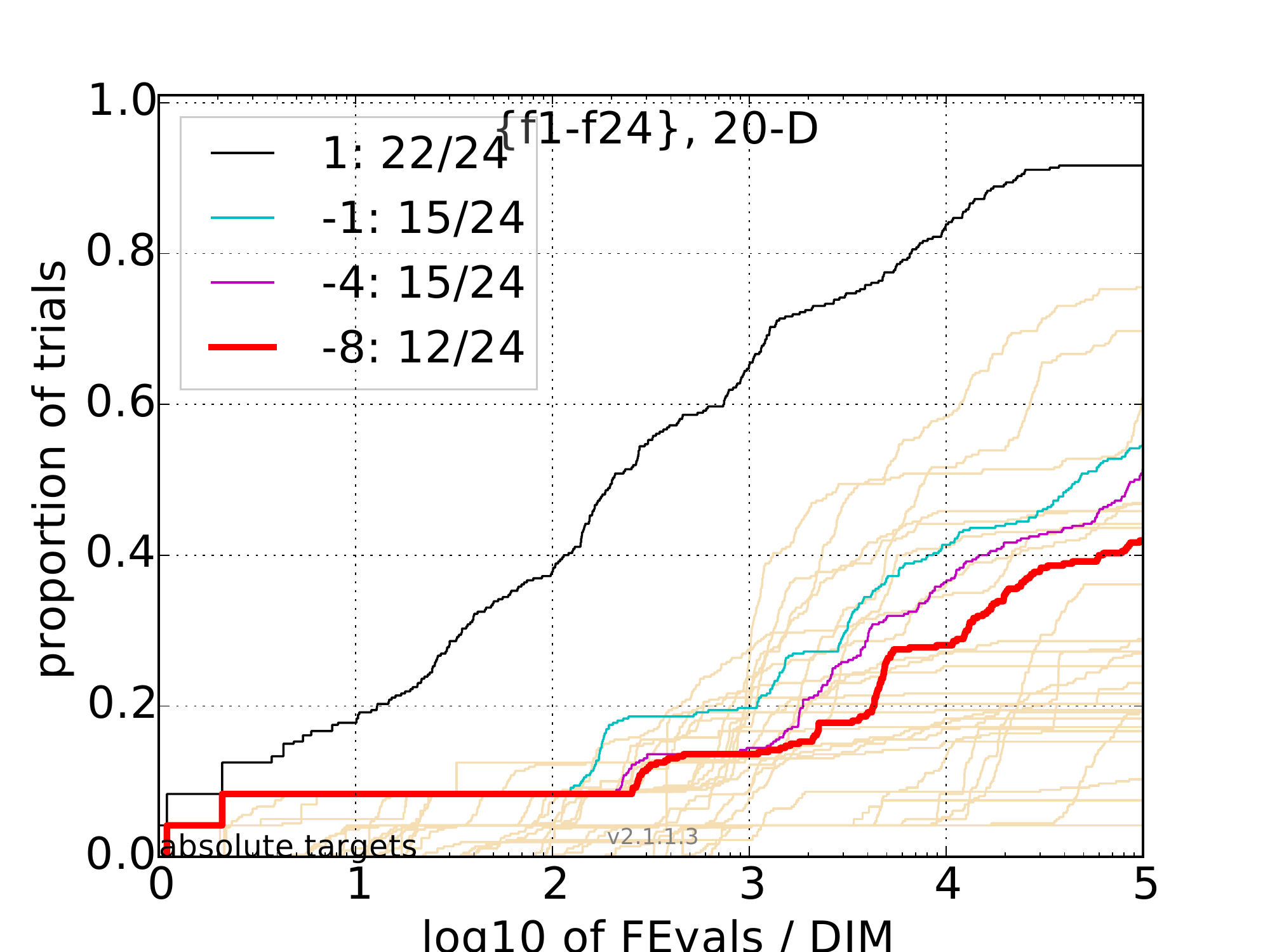}\\
	\vspace{-1.7em}
	\begin{center}%
		Adapted DE+PSO+Square function+Polynomial function\\
	\end{center}%
	\vspace{-0.5em}
	\caption{Results of various M-GAPSO configurations on 5D (left column) and 20D (right column) functions from COCO BBOB.
	\label{fig:algorithm-results-improvements}}
\end{figure}

Performance plots of the $4$ above-mentioned M-GAPSO configurations are presented in Figure~\ref{fig:algorithm-results-improvements}.
Enhancing behavior pool with local neighborhood SPSO-2007 algorithm,
and subsequently with square and polynomial function based local optimizers,
brings observable improvement in the case of 5D benchmark functions.
Adaptation of behaviors and restart management improves the results also on 20D functions.
A summary of function types with at least one successful run and the overall percentage
of successful algorithm runs are presented in Tables~\ref{tab:succ-count} and~\ref{tab:succ-perc},
respectively. Additionally, Figure~\ref{fig:computations-time} presents the average
computation times for various M-GAPSO configurations.


With DE considered to be a baseline algorithm, the following conclusions could be stated: (1) Switching off samples's caching results in around 3 times faster computations. (2) Mixing DE with PSO does not affect the speed significantly, while including model-based behaviors leads to $4$ times slower computation than that in the baseline experiment. (3) The most significant disadvantage, in terms of the average computation time of a single function evaluation, is caused by simultaneous inclusion of model based approaches and the adaptation mechanism. The reason for that slowdown are algorithm iterations in which no improvement was observed for a certain amount of time. Therefore, the adaptation mechanism guided around half of the function evaluations to be made by one of the model based approaches (these methods related to samples memory are inherently slower than sampling based PSO and DE
behaviors).

In order to further investigate which of the particular methods
had the highest impact on M-GAPSO results,
several additional experiments were performed.
Their goal was to address the following questions:
\begin{itemize}
	\item Should performance difference between plain DE
	and DE hybridized with PSO, be attributed to the adaptation of behaviors mechanism
	or is it simply enough to mix these behaviors among particles?
	\item How performance of the algorithm is affected by an addition of model based optimization behaviors?
	\item How selection of the swarm initialization strategy (i.e. selection of a bounding box
	for population initialization) influences the algorithm's performance?
	\item Would a combination of all of the above-mentioned aspects (adaptation, modeling-based behaviors, restart strategies) demonstrate the advantage over application of each of them alone?
	\item How does M-GAPSO fare against its previous version (GAPSO) and against state-of-the-art variant of CMA-ES?
\end{itemize}

\begin{table}[ht!]
	\caption{Numbers of distinct function types obtained by each algorithm for the respective optimization precision target ($\Delta$ Value) on 5D and 20D benchmark functions, respectively.
	\label{tab:succ-count}}%
	\small%
	\begin{center}%
		\begin{tabular}{rr|rrrr}
			 &  & &	&  & Adapted \\
			Dimension & $\Delta$Value &	DE &	PSO+DE & PSO+DE+ & PSO+DE+ \\
			 &  &	 &	 & LM+PM & LM+PM  \\\hline
	5D &	$10^1$ & \textbf{24} & \textbf{24} & \textbf{24} & \textbf{24} \\
	5D &	$10^{-1}$ & \textbf{23} & \textbf{23} & \textbf{23} & \textbf{23} \\
	5D &	$10^{-4}$ & 21 & 21 & 22 & \textbf{23} \\
	5D &	$10^{-8}$ & 20 & 20 & 21 & \textbf{22} \\\hline
	20D &	$10^{1}$ & 20 & 20 & \textbf{22} & \textbf{22} \\
	20D &	$10^{-1}$ & 13 & 13 & 14 & \textbf{15} \\
	20D &	$10^{-4}$ & 13 & 12 & 12 & \textbf{15} \\
	20D &	$10^{-8}$ & 10 & 10 & 10 & \textbf{12} \\
		\end{tabular}%
	\end{center}%
	\end{table}

	\begin{table}[ht!]
		\caption{Percentage rates of successfully completed runs by each algorithm for the respective optimization precision target ($\Delta$ Value) on 5D and 20D benchmark functions, respectively.
		\label{tab:succ-perc}}%
		\small%
		\begin{center}%
			\begin{tabular}{rr|rrrr}
				&  & &	&  & Adapted \\
				Dimension & $\Delta$Value &	DE &	PSO+DE & PSO+DE+ & PSO+DE+ \\
				 &  &	 &	 & LM+PM & LM+PM  \\\hline
		5D &	$10^1$ & \textbf{{1.00}} & \textbf{{1.00}} & \textbf{{1.00}} & \textbf{{1.00}} \\
		5D &	$10^{-1}$ & {0.88} & {0.88} & {0.92} & \textbf{{0.94}} \\
		5D &	$10^{-4}$ & {0.76} & {0.81} & {0.85} & \textbf{0.86} \\
		5D &	$10^{-8}$ & {0.72} & {0.76} & {0.80} & \textbf{0.82} \\\hline
		20D &	$10^{1}$ & 0.83 & 0.83 & {0.88} & \textbf{0.91} \\
		20D &	$10^{-1}$ & {0.53} & 0.52 & 0.52 & \textbf{0.55} \\
		20D &	$10^{-4}$ & {0.48} & 0.45 & 0.44 & \textbf{0.51} \\
		20D &	$10^{-8}$ & {0.41} & 0.40 & 0.40 & \textbf{0.42} \\
			\end{tabular}%
		\end{center}%
		\end{table}

\begin{figure}[H]
	\begin{tabular}{cc}
		\includegraphics[width=0.465\textwidth]{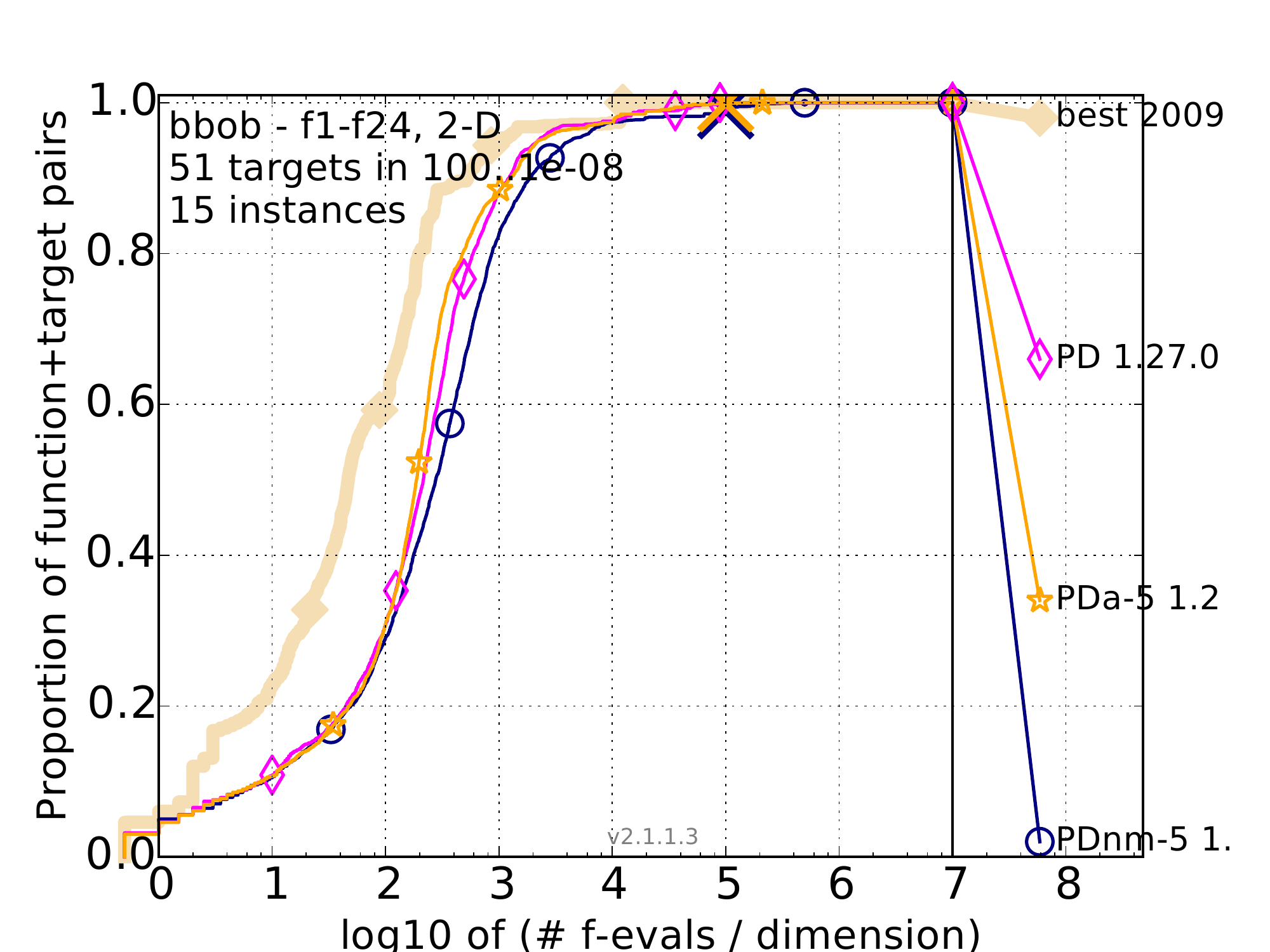} &
		\includegraphics[width=0.465\textwidth]{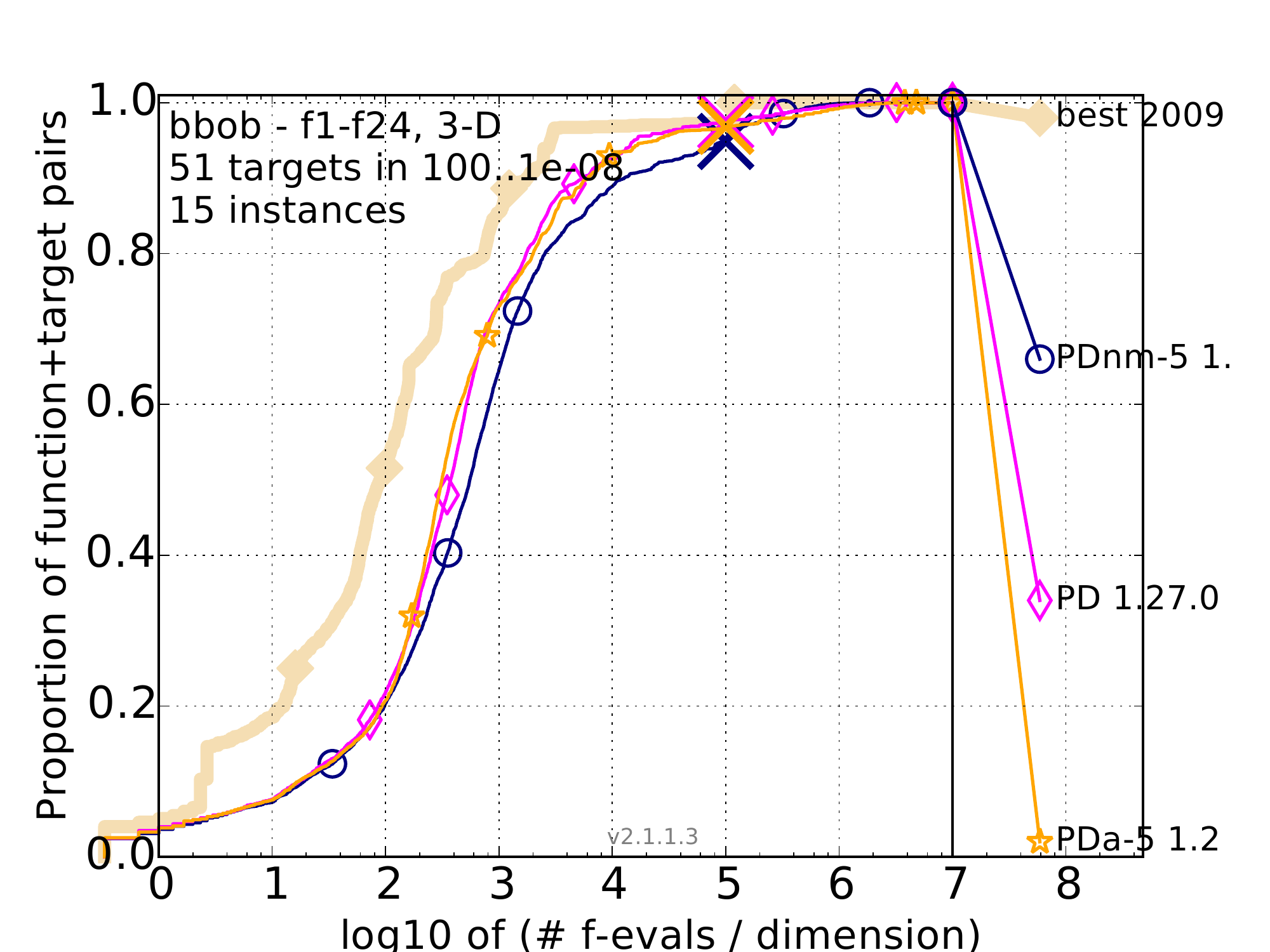}\\
		2D & 3D \\
		\includegraphics[width=0.465\textwidth]{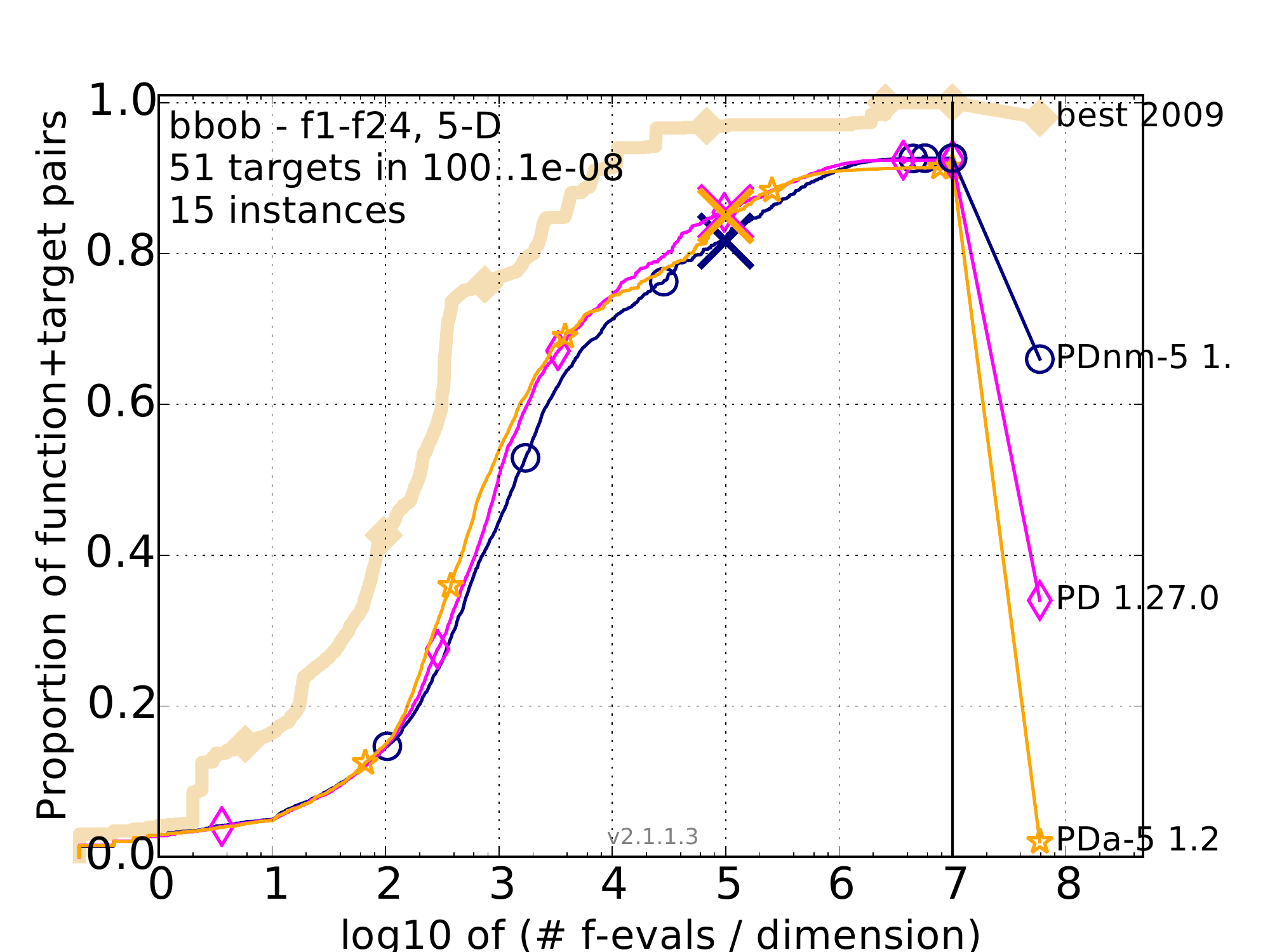} &
		\includegraphics[width=0.465\textwidth]{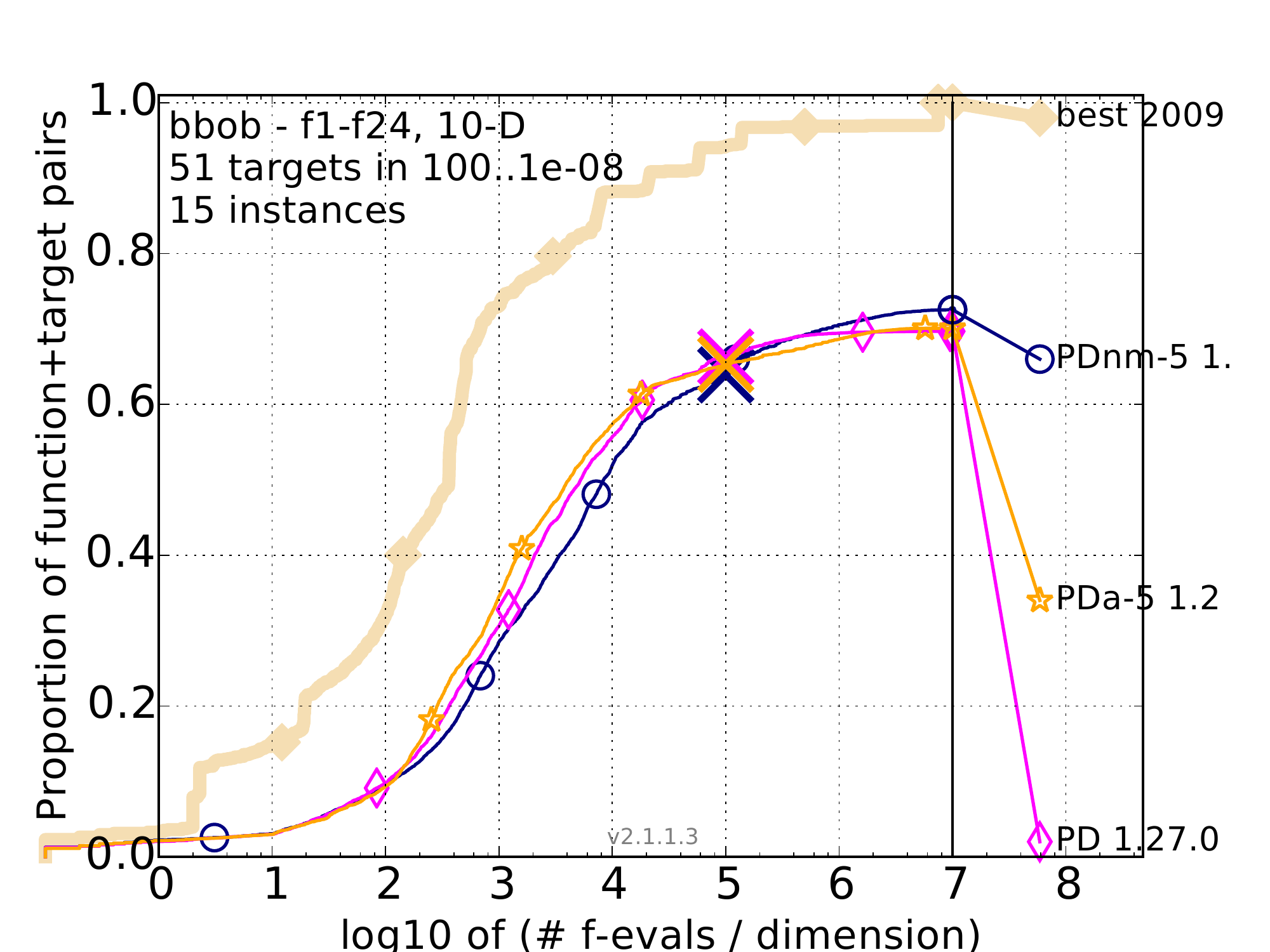}\\
		5D & 10D \\
		\includegraphics[width=0.465\textwidth]{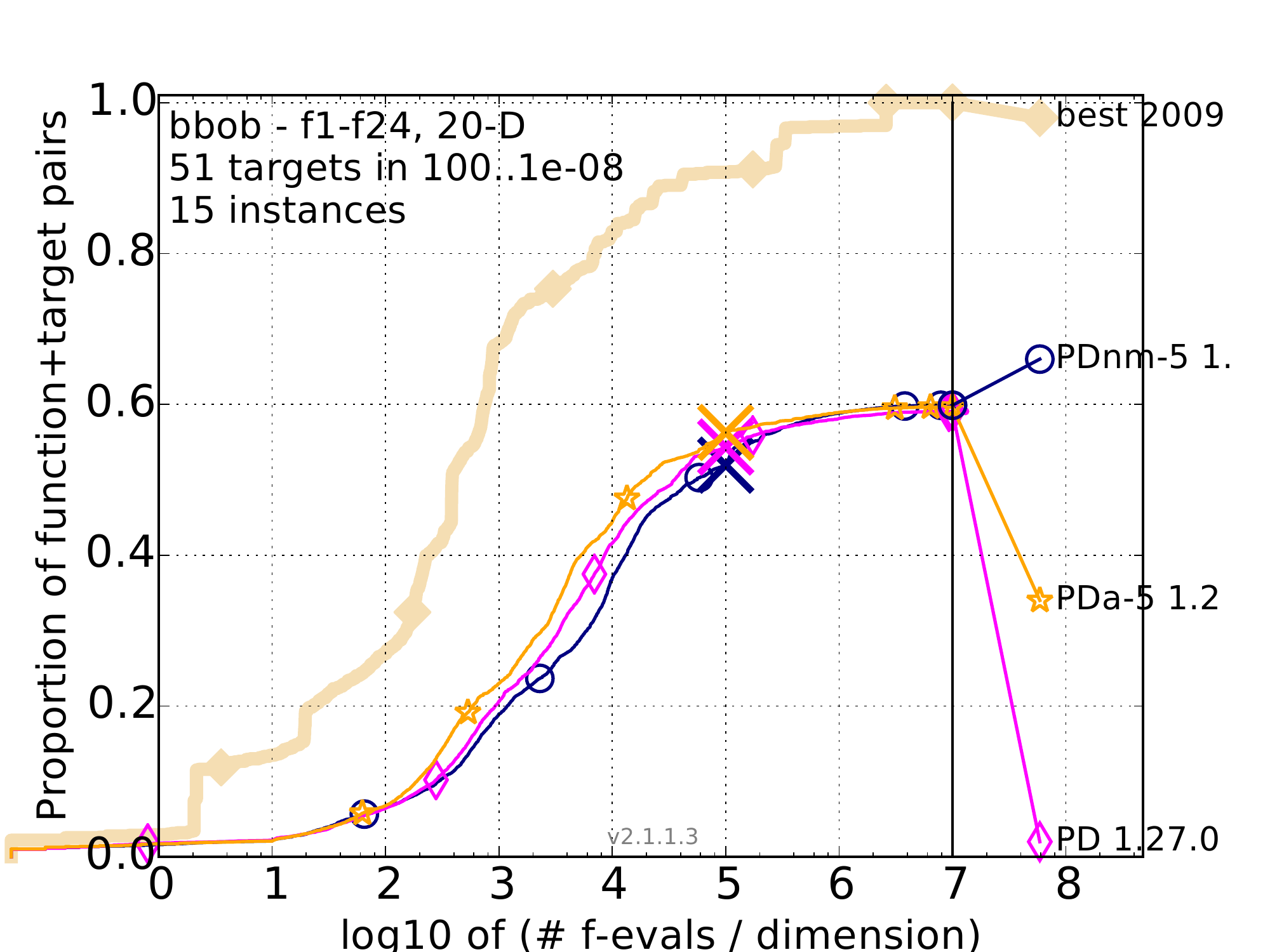} &
		\includegraphics[width=0.465\textwidth]{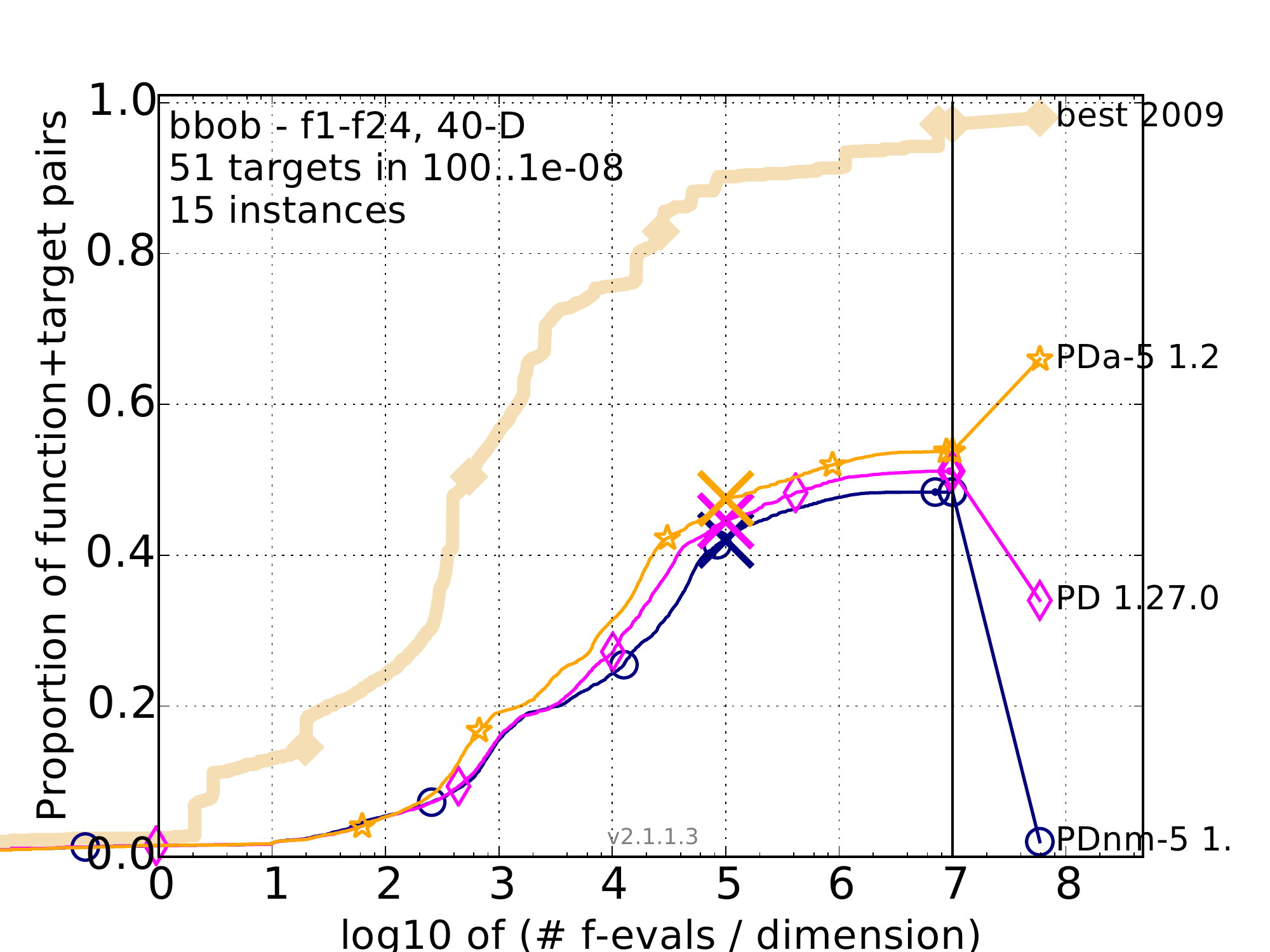}\\
		20D & 40D \\
	\end{tabular}
	\caption{Results of M-GAPSO on 2D, 3D, 5D, 10D, 20D and 40D COCO BBOB functions:
	PSO and DE behaviors applied within a static particles' behavior pool setup (PDnm),
	a setup with behaviors mixed among particles before each iteration (PD),
	and a setup with adapted behaviors pool (PDa).
	\label{fig:algorithm-results-all-D-mixing-effects}}
\end{figure}

In order to assess the impact of behaviors mixing and that of adaptation mechanism, M-GAPSO
(utilizing DE and PSO) was run in three configurations.
Figure~\ref{fig:algorithm-results-all-D-mixing-effects} presents the results.
In the first configuration (denoted \textit{PDnm}) the behaviors of particles
were set at the beginning of the optimization process.
In the second one (denoted \textit{PD}) the probabilities of behaviors
were manually set up at the beginning, but assignment of behaviors to particles took place before each iteration.
In the third one (denoted \textit{PDa}) the behaviors probabilities were adapted
during the entire optimization process, and assignment of behaviors to particles was made before each iteration.
It can be observed that addition of behaviors mixing mechanism (cf. line \ref{line:alg:GAPSO:mixing} of the pseudocode of Algorithm \ref{alg:GAPSO}) before each iteration has a clearly and positive impact on the performance of the PSO-DE hybrid.
Meanwhile, the adaptation procedure influences the algorithm's performance only for some function dimensions.
However, in order to outrank a pure DE algorithm, both these techniques were necessary.

Since the behaviors mixing procedure proven to be undoubtedly useful, this technique will be used in all remaining experiments.
While the adaptation mechanism is also clearly important, its application increases computational cost in the case of model-based behaviors
(cf. Figure~\ref{fig:computations-time}). Hence, adaptation will be applied only in a selected subset of experiments.

\begin{figure}[H]
	\begin{tabular}{cc}
		\includegraphics[width=0.465\textwidth]{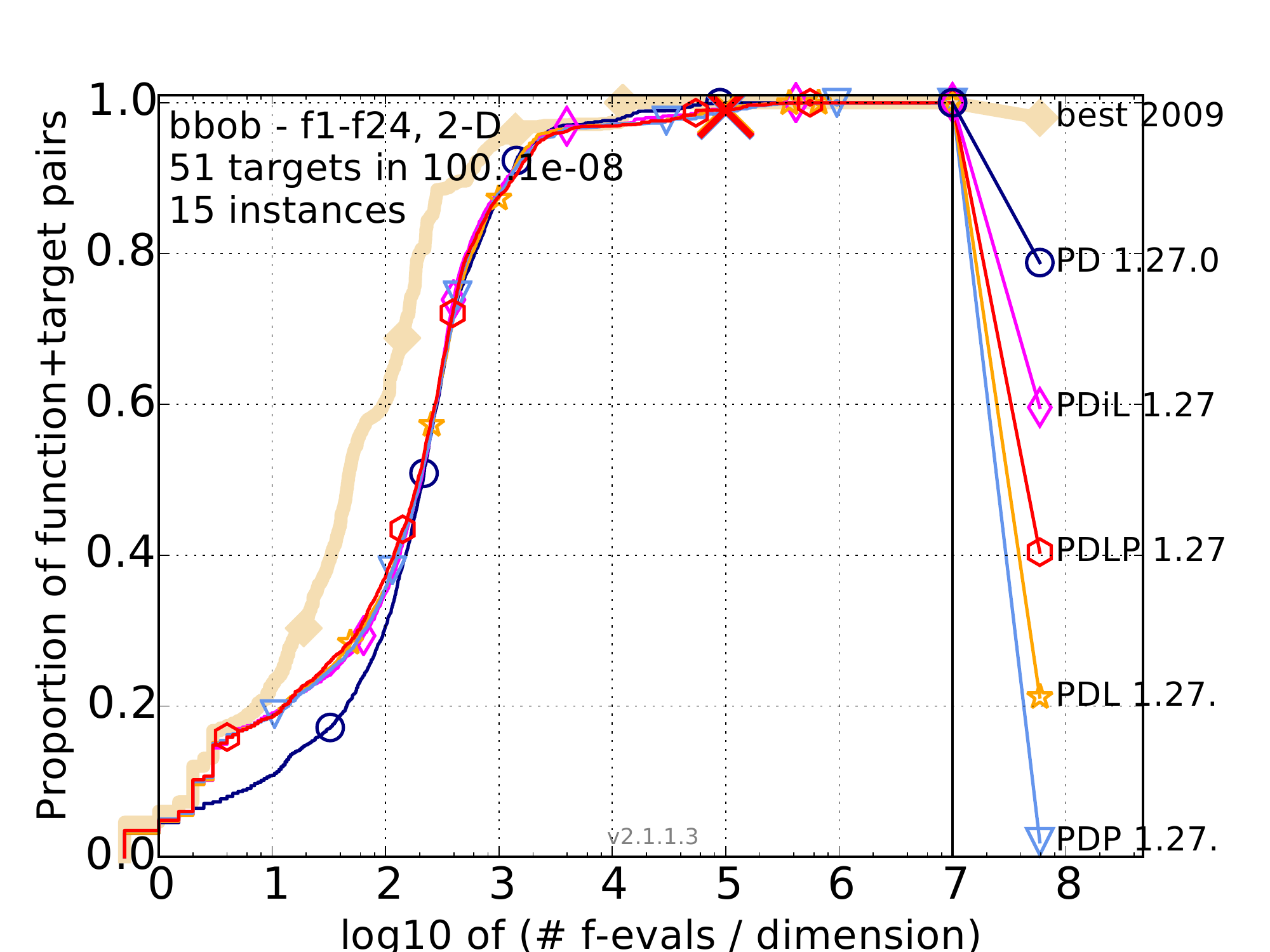} &
		\includegraphics[width=0.465\textwidth]{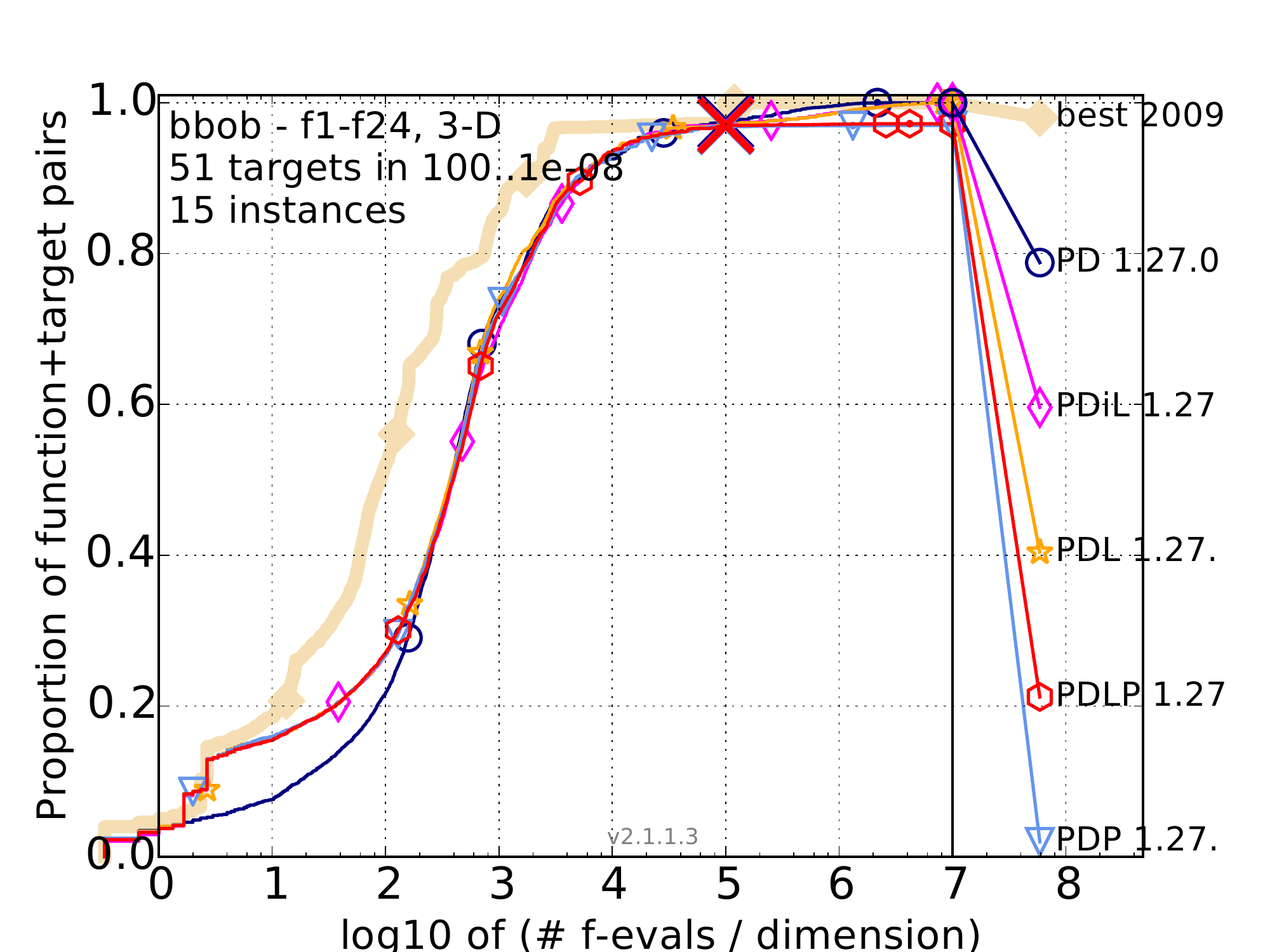}\\
		2D & 3D \\
		\includegraphics[width=0.465\textwidth]{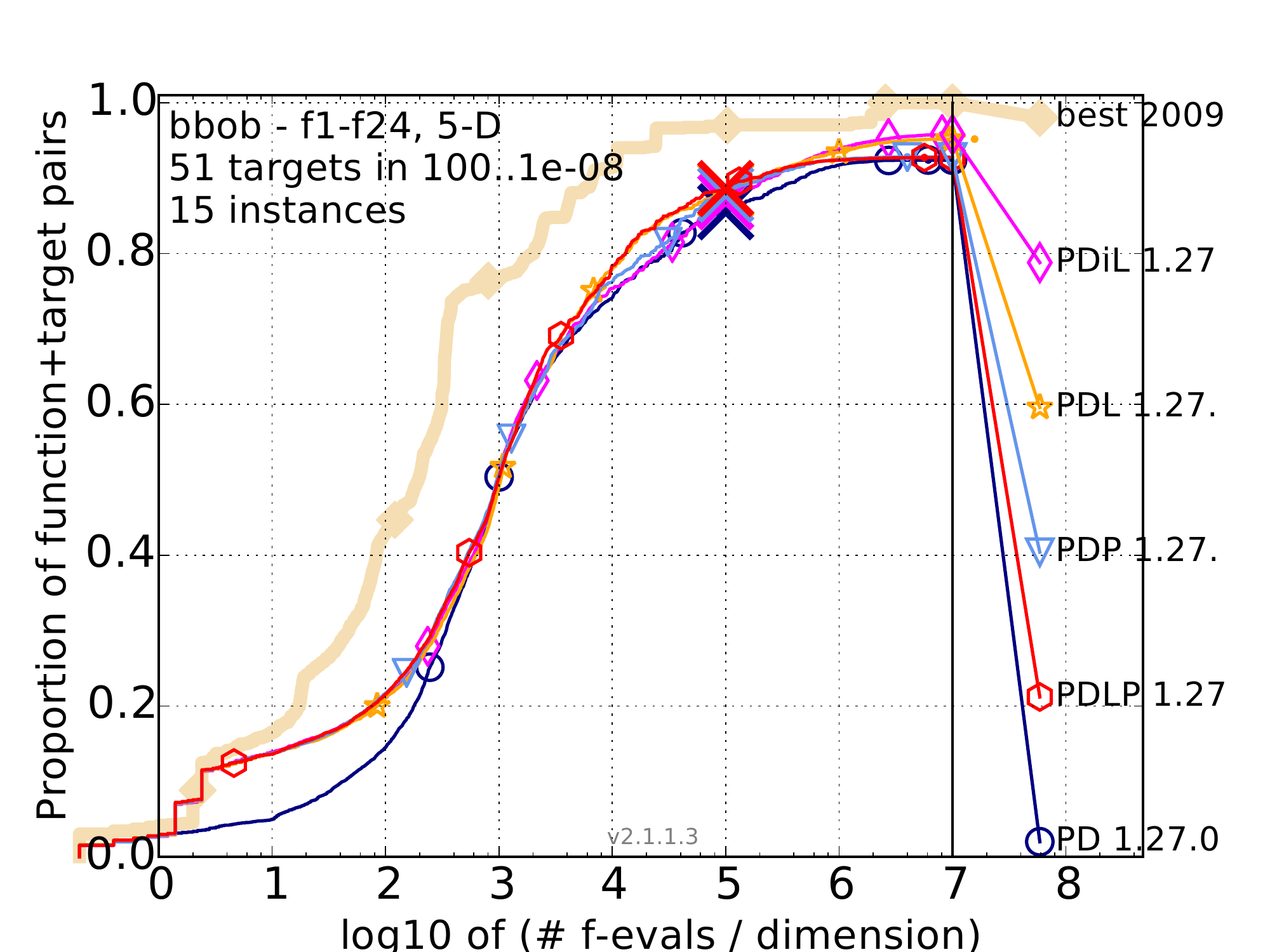} &
		\includegraphics[width=0.465\textwidth]{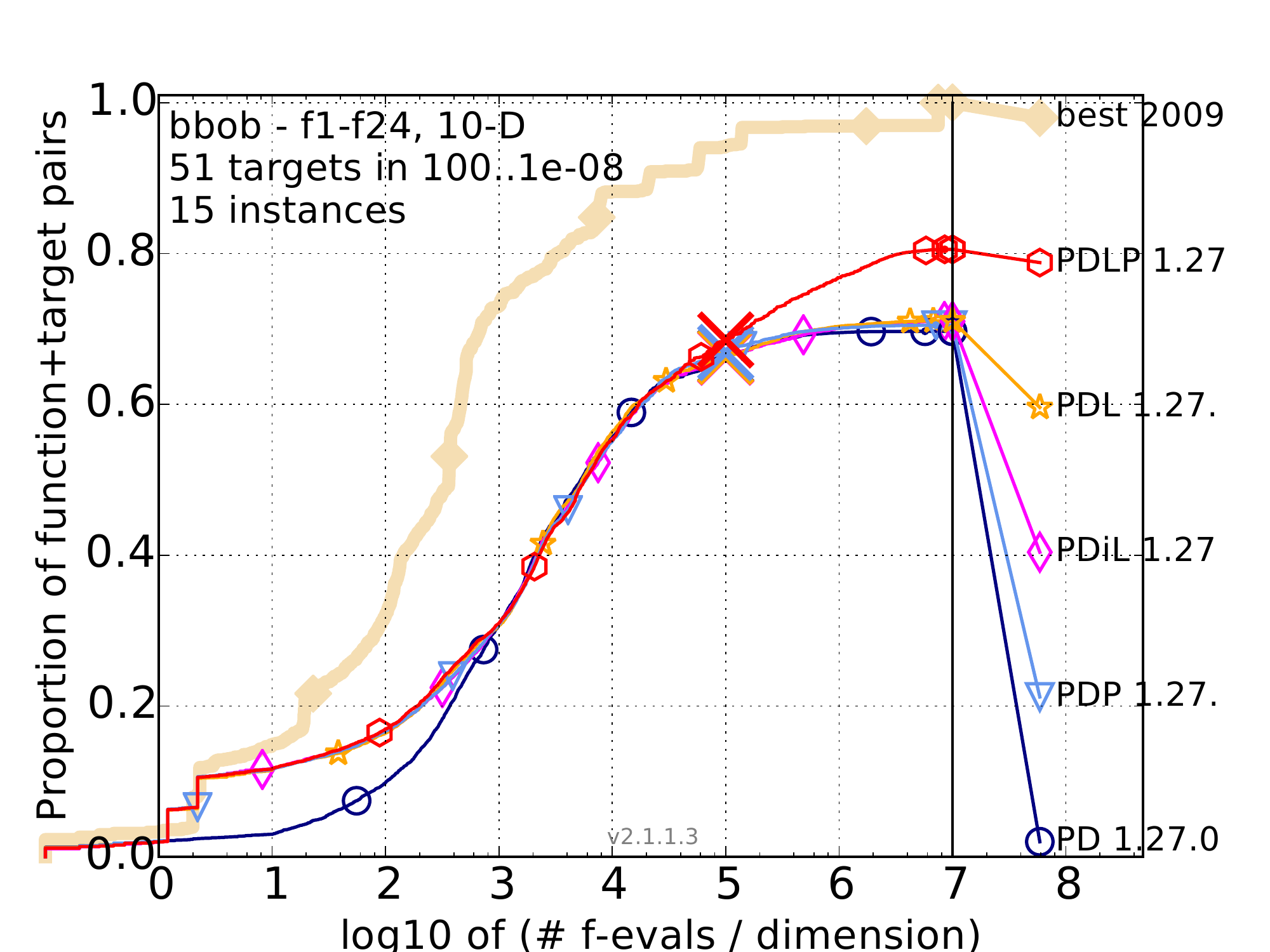}\\
		5D & 10D \\
		\includegraphics[width=0.465\textwidth]{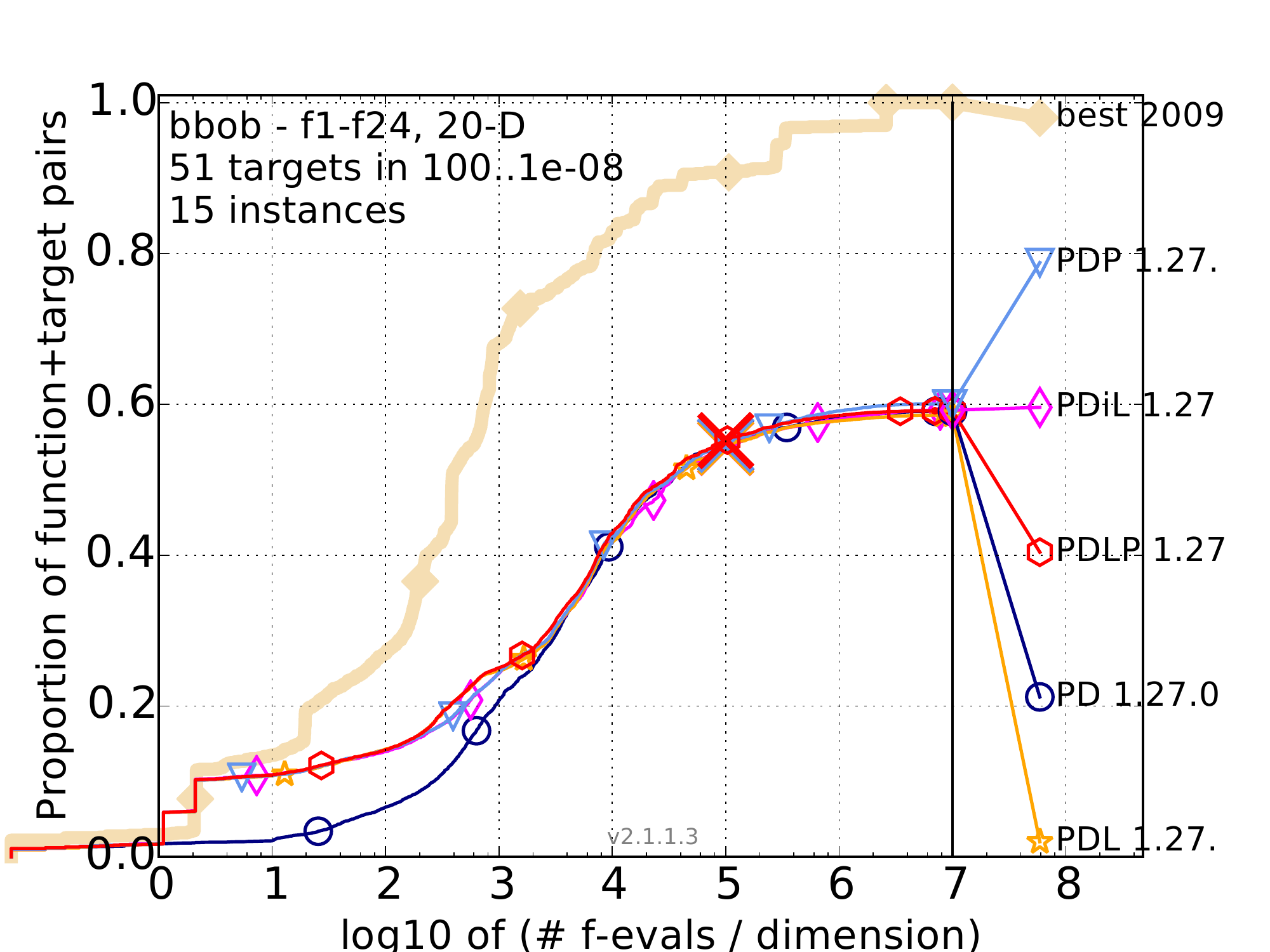} &
		\includegraphics[width=0.465\textwidth]{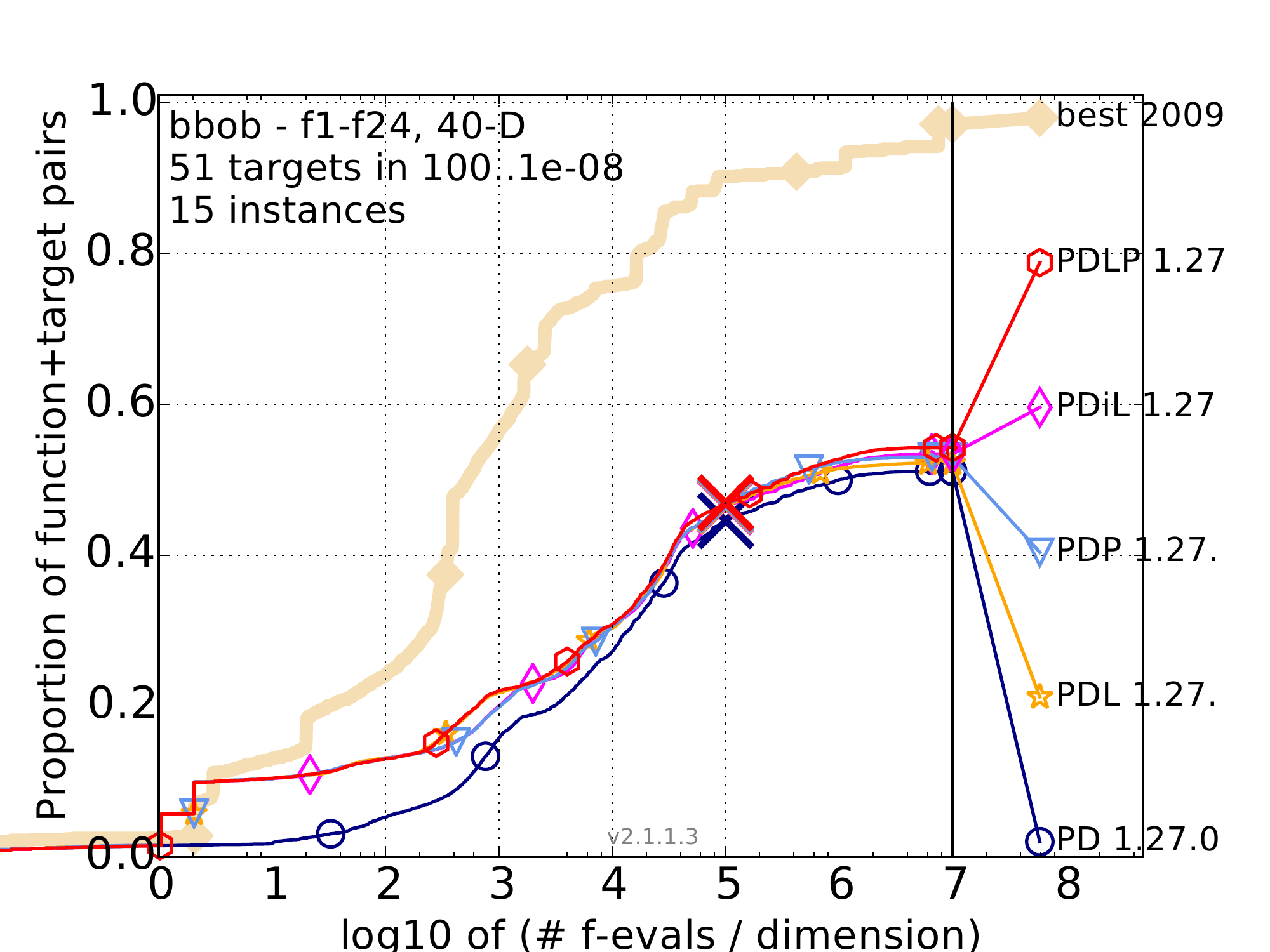}\\
		20D & 40D \\
	\end{tabular}
	\caption{Results of M-GAPSO on 2D, 3D, 5D, 10D, 20D and 40D COCO BBOB functions for
	various combinations of PSO, DE and linear and polynomial models from the behavior pool.
	\label{fig:algorithm-results-all-D-models-vs-no-models}}
\end{figure}

\begin{figure}[H]
	\begin{tabular}{cc}
		\includegraphics[width=0.465\textwidth]{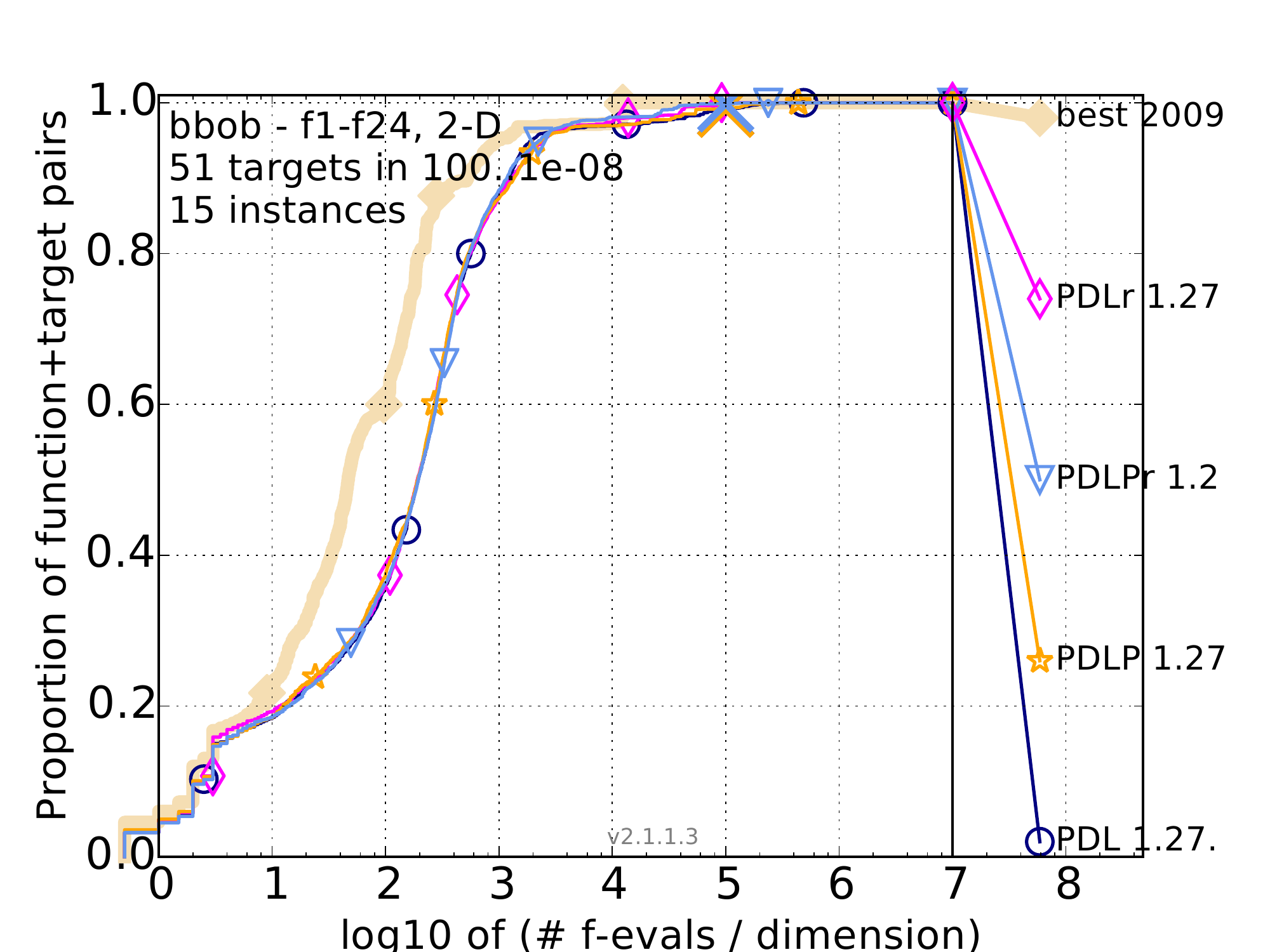} &
		\includegraphics[width=0.465\textwidth]{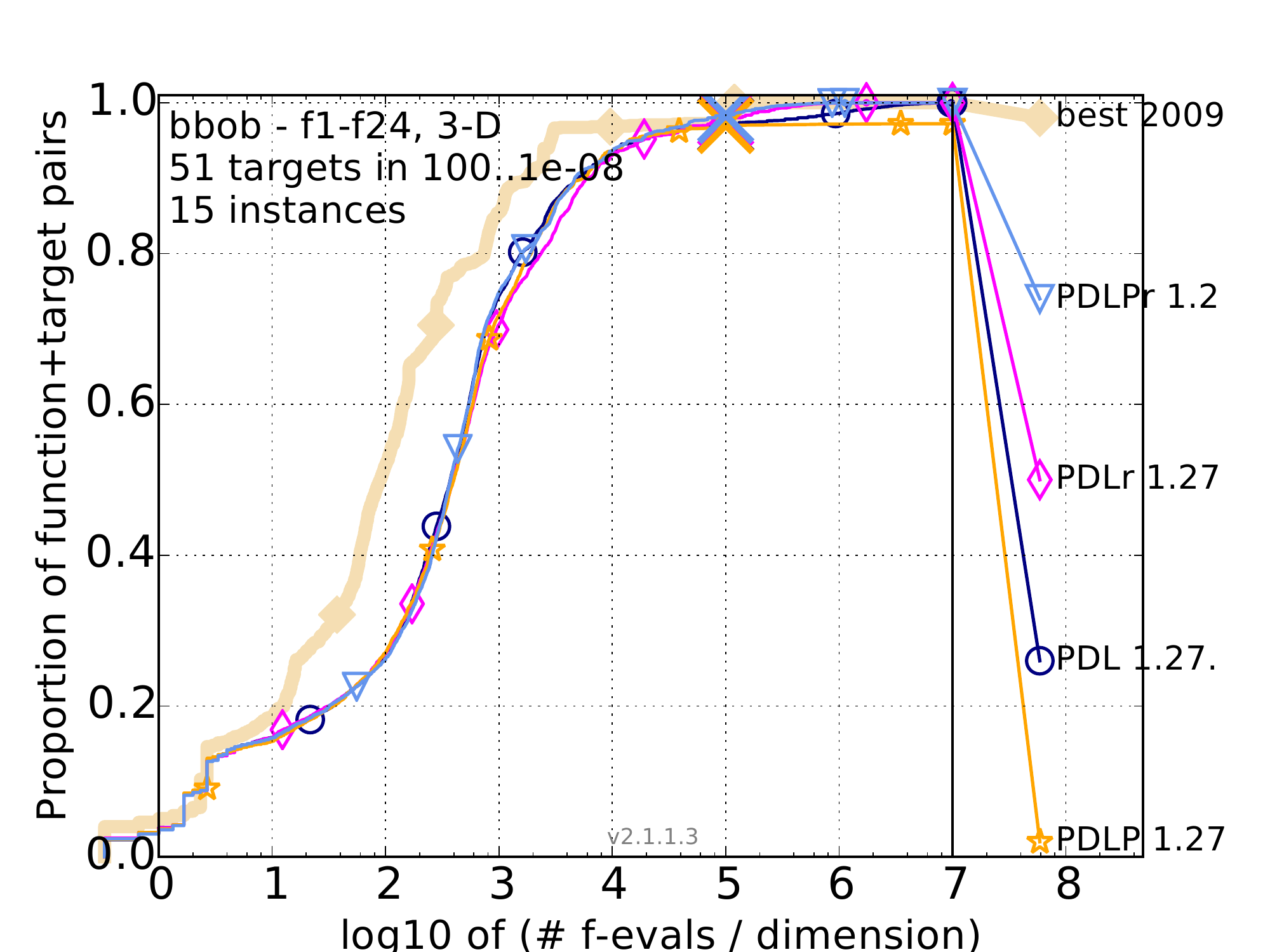}\\
		2D & 3D \\
		\includegraphics[width=0.465\textwidth]{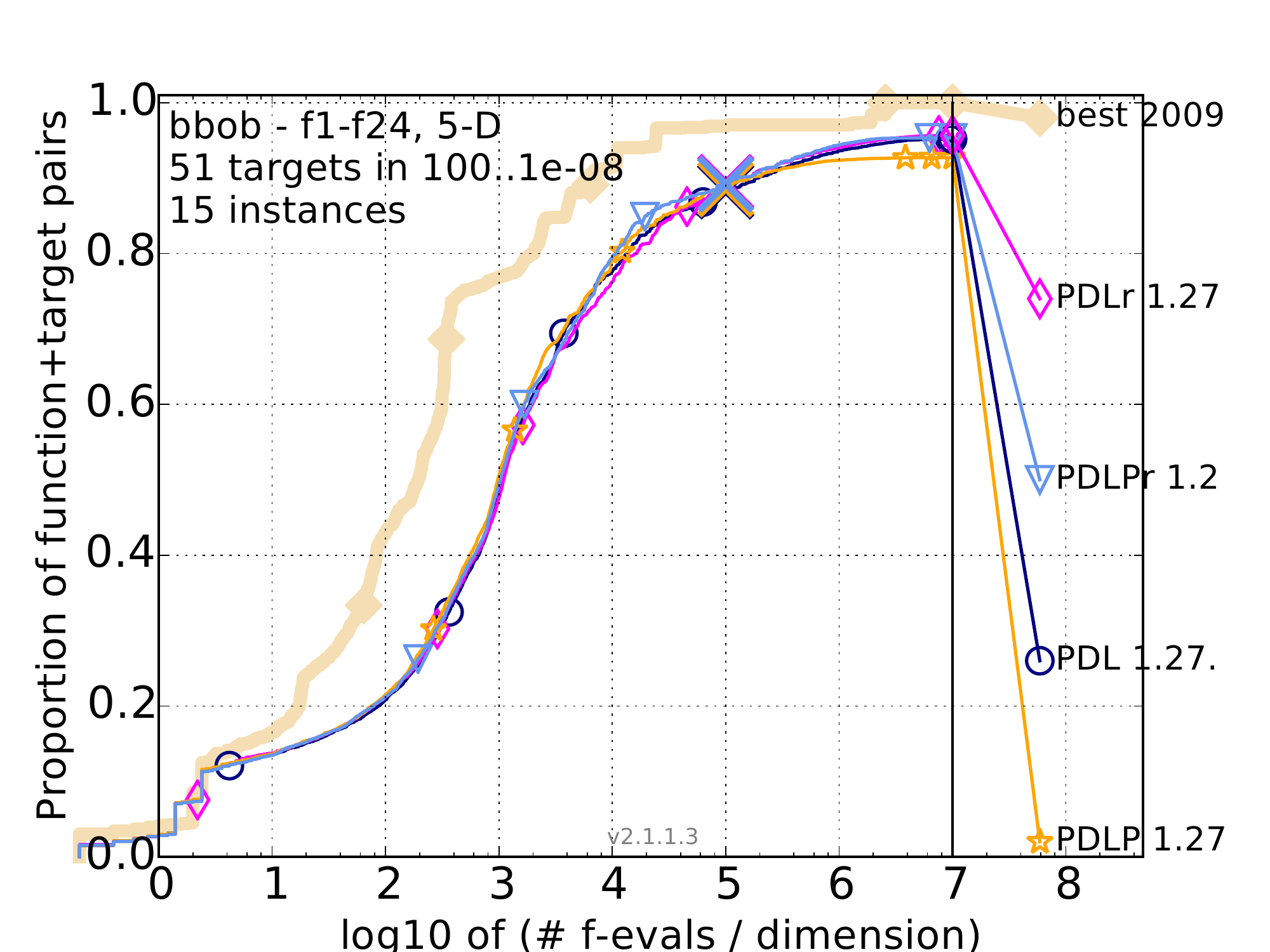} &
		\includegraphics[width=0.465\textwidth]{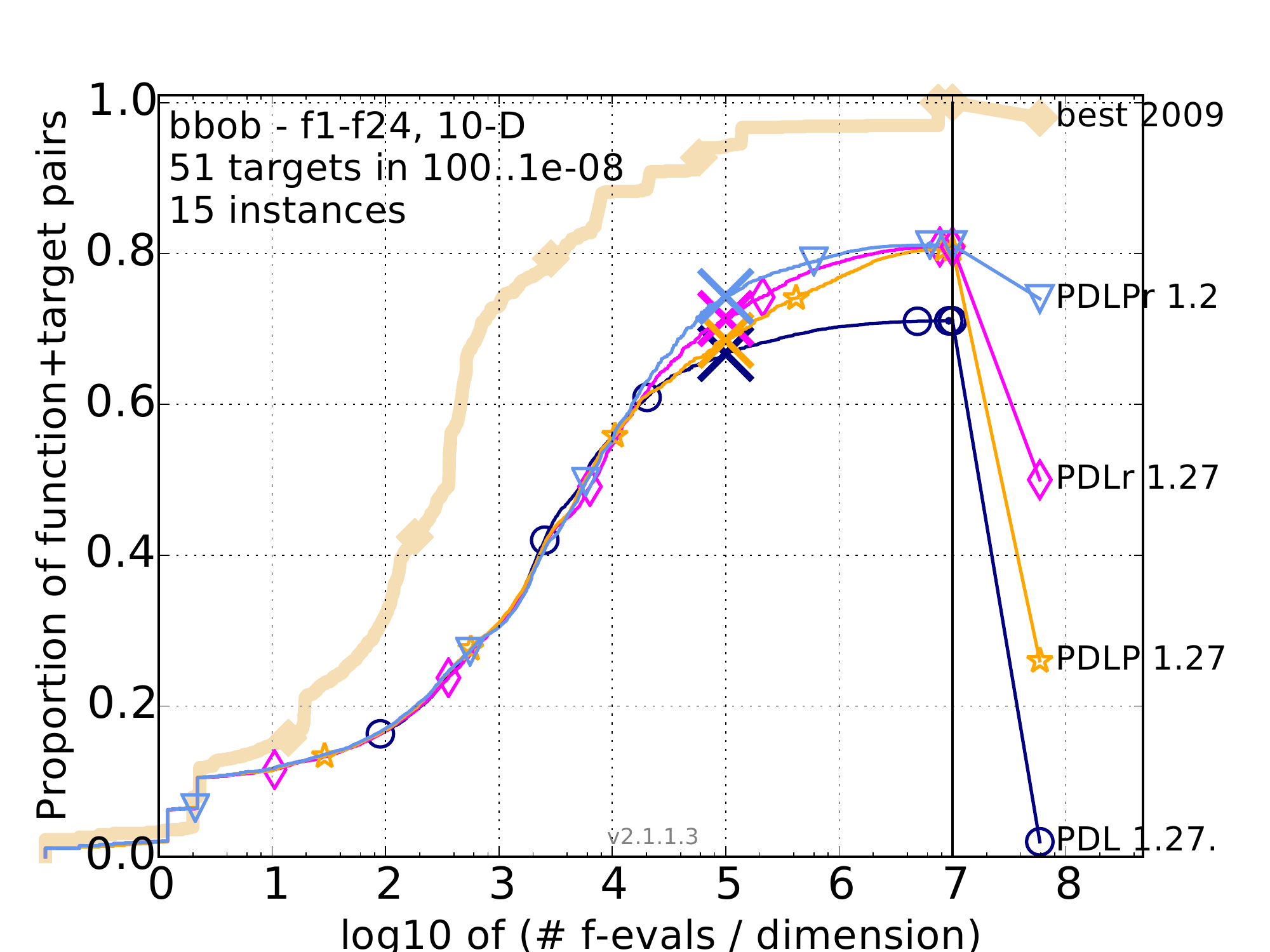}\\
		5D & 10D \\
		\includegraphics[width=0.465\textwidth]{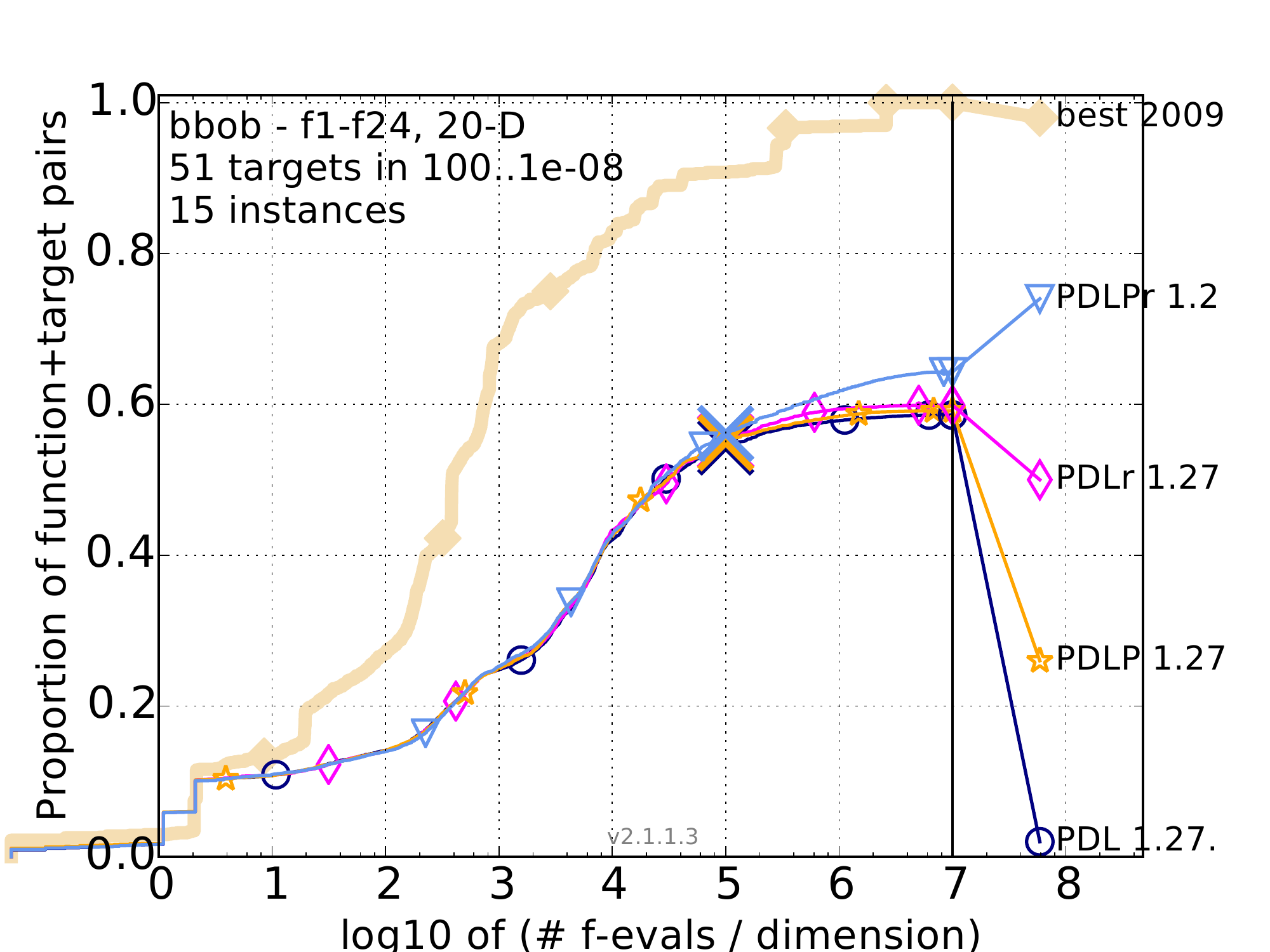} &
		\includegraphics[width=0.465\textwidth]{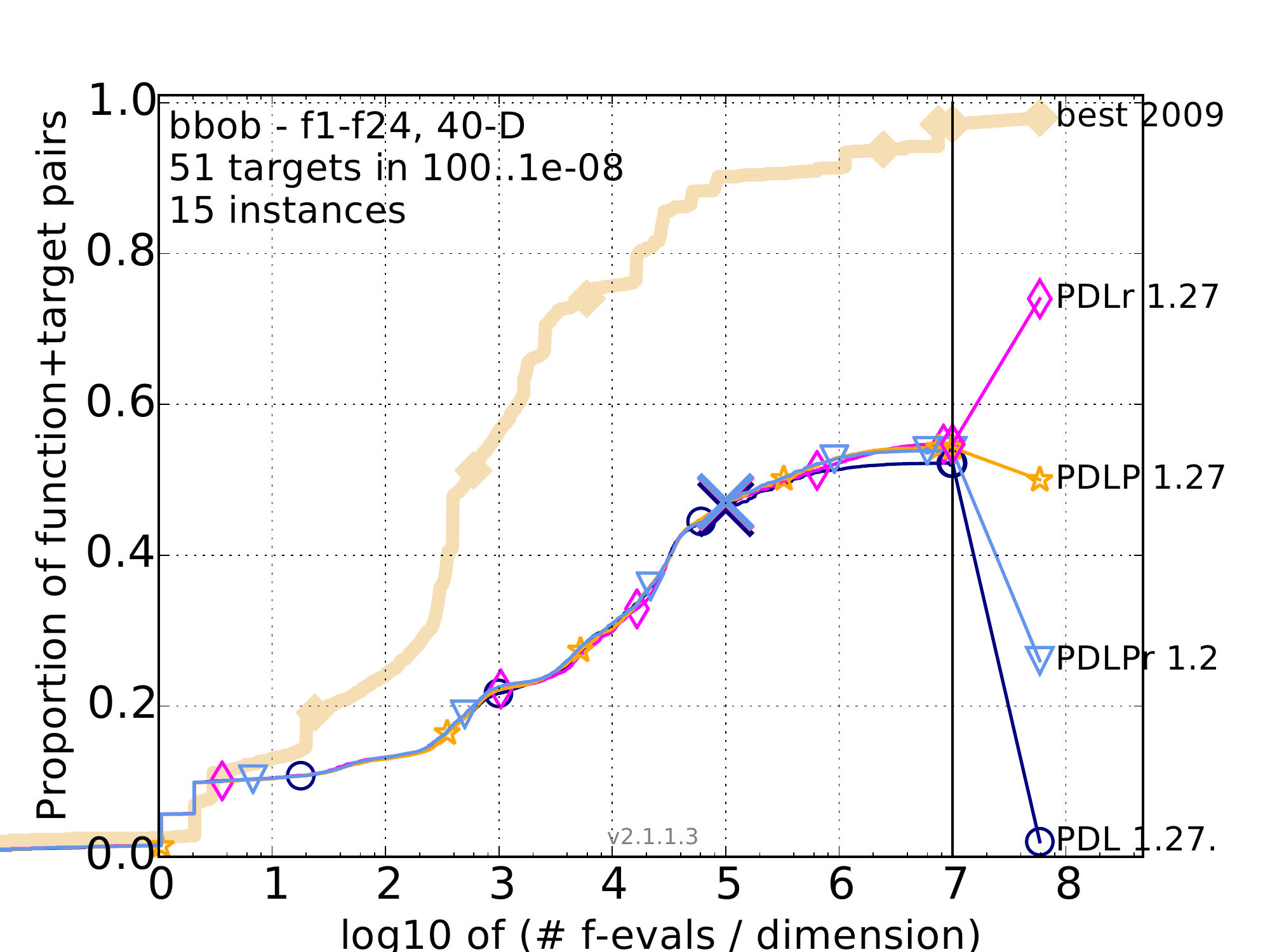}\\
		20D & 40D \\
	\end{tabular}
	\caption{The effects of M-GAPSO restart mechanism which refers to the previously found optima. Results on 2D, 3D, 5D, 10D, 20D and 40D COCO BBOB functions. 
	\label{fig:algorithm-results-all-D-restarts}}
\end{figure}




The focus of the first experiment is on assessment of the quadratic and polynomial
model behaviors. Figure~\ref{fig:algorithm-results-all-D-models-vs-no-models}
depicts cumulative plots of function evaluation counts required to reach an optimization
target for the $5$ following configurations: baseline PSO and DE behaviors (denoted \textit{PD}); baseline PSO and DE behaviors with additional square model
applied during population initialization phase (\textit{PDiL});
PSO, DE and square model behaviors (\textit{PDL});
PSO, DE and polynomial model behaviors (\textit{PDP});
and PSO, DE, square model and polynomial model behaviors (\textit{PDLP}).
In all of the above configurations behaviors were mixed.
In the resulting plots one can generally observe that having anyone of model-based behaviors is beneficial.
Additionally, configurations with more variable model-based behaviors performed
slightly better in terms of found optima.
As a side note, it should be mentioned that the square model
was significantly faster than the polynomial one, due to smaller
number of utilized samples and simpler formula for optimum identification.

The last mechanism to be tested is the
initialization scheme, described in Section~\ref{sec:initialization-scheme}, with the aim of identifying the area for potential future improvements of M-GAPSO.
Figure~\ref{fig:algorithm-results-all-D-restarts} presents the results
of four configurations. Two of them (\textit{PDL} and \textit{PDLP}) were copied from the previous experiment, while the two others
(\textit{PDLr} and \textit{PDLPr}) are their counterparts, with improved initialization scheme in place of a simple population reset.
The results show that the improved initialization procedure has indeed potential, for instance for 10D functions \textit{PDLPr}
is clearly better than \textit{PDLP} and \textit{PDLr}
outperforms \textit{PDL}.

\subsection{BBComp competition results}
\begin{figure}[H]
	\begin{tabular}{lr}
	2018 - 1 objective (standard) &
	2018 - 1 objective (expensive) \\
	\includegraphics[width=0.465\textwidth]{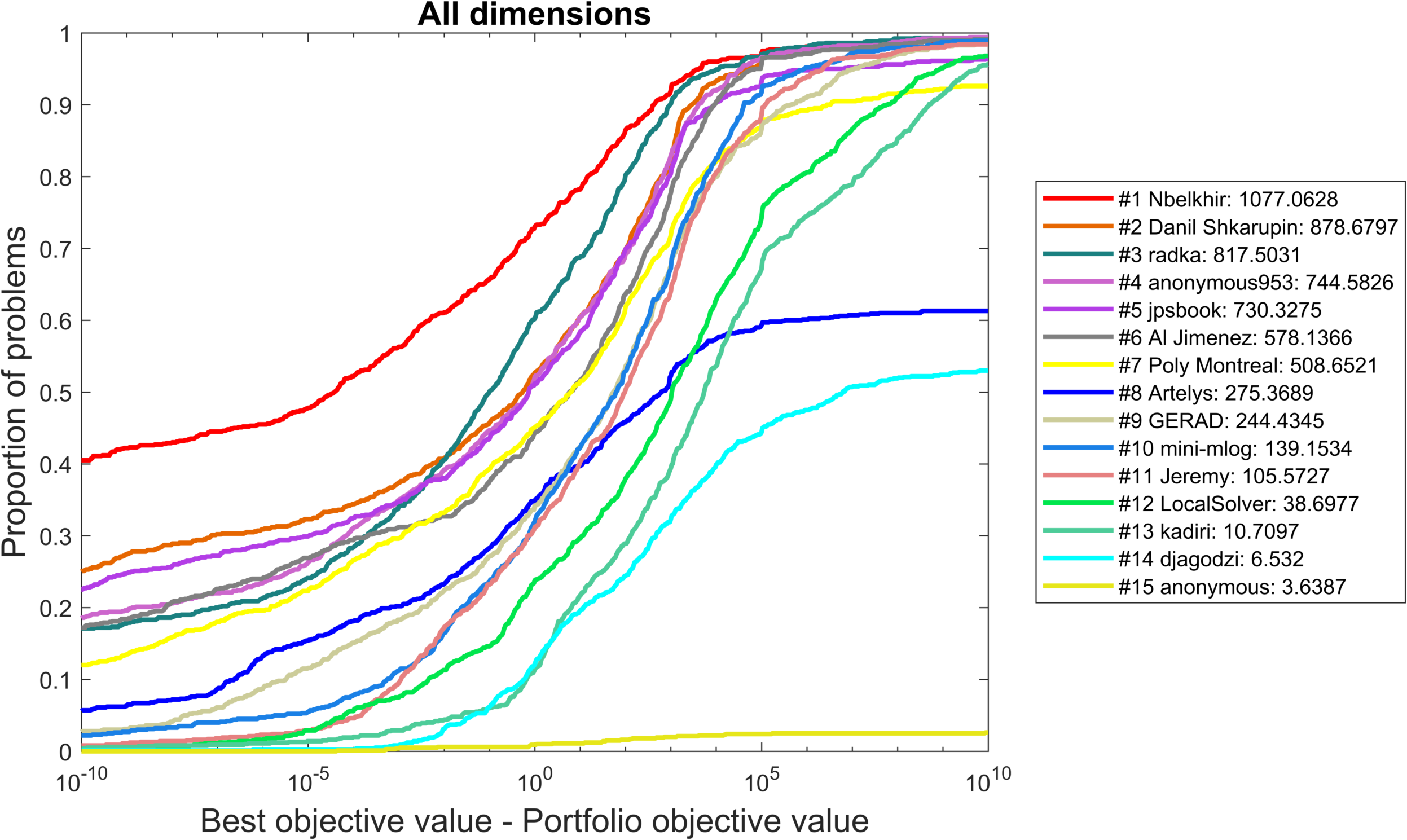} &
	\includegraphics[width=0.465\textwidth]{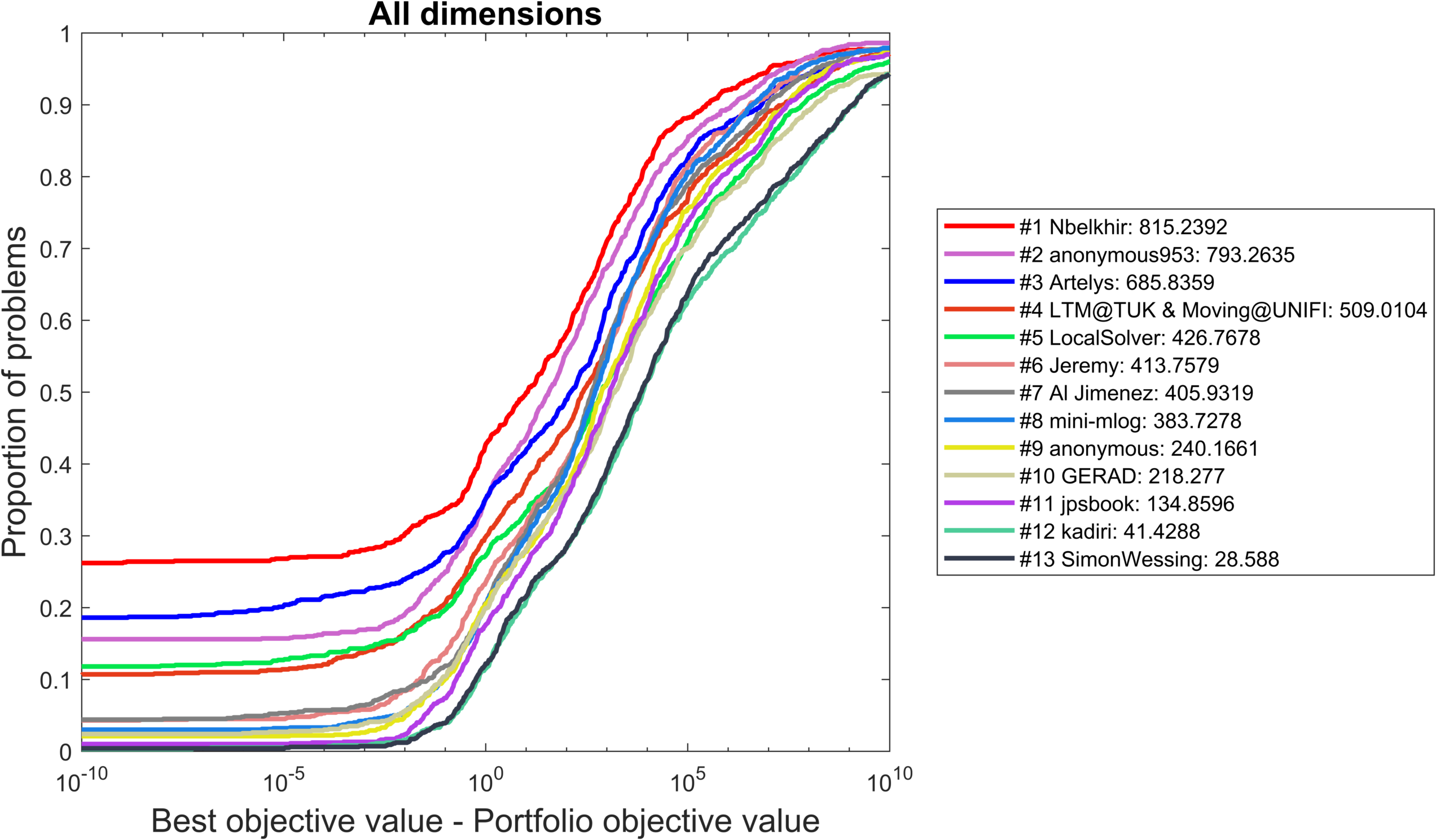} \\
	2019 - 1 objective (standard) &
	2019 - 1 objective (expensive) \\
	\includegraphics[width=0.465\textwidth]{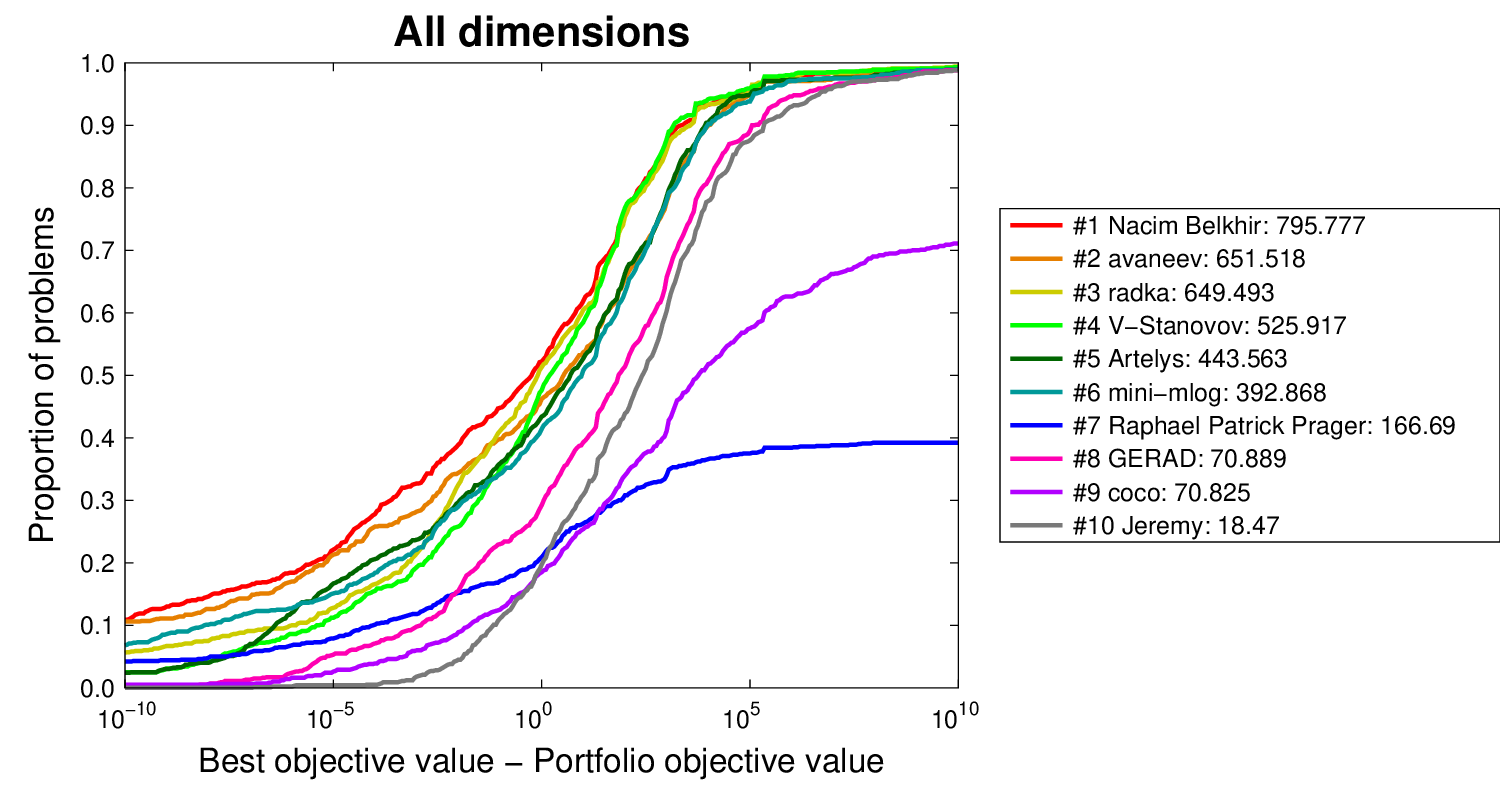} &
	\includegraphics[width=0.465\textwidth]{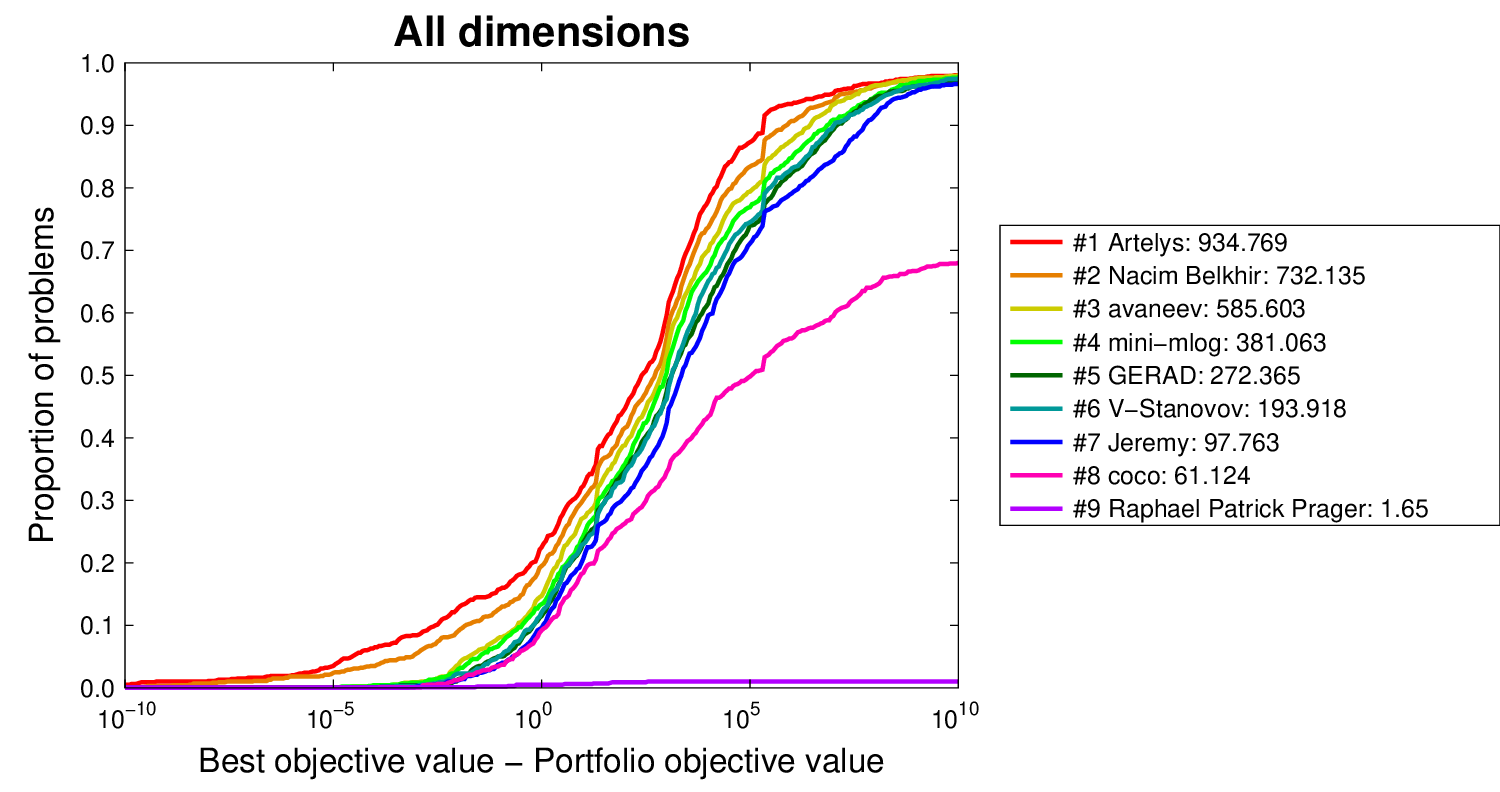} \\
	\end{tabular}
	\caption{Summarized results from the BBComp black-box optimization competition in
	single objective function setup in years 2018 and 2019. GAPSO (2018) and M-GAPSO (2019)
	algorithms were fielded by our \texttt{mini-mlog} team.
	\label{fig:bbcomp-official}}
\end{figure}

\begin{figure}[t]
	\includegraphics[width=\textwidth]{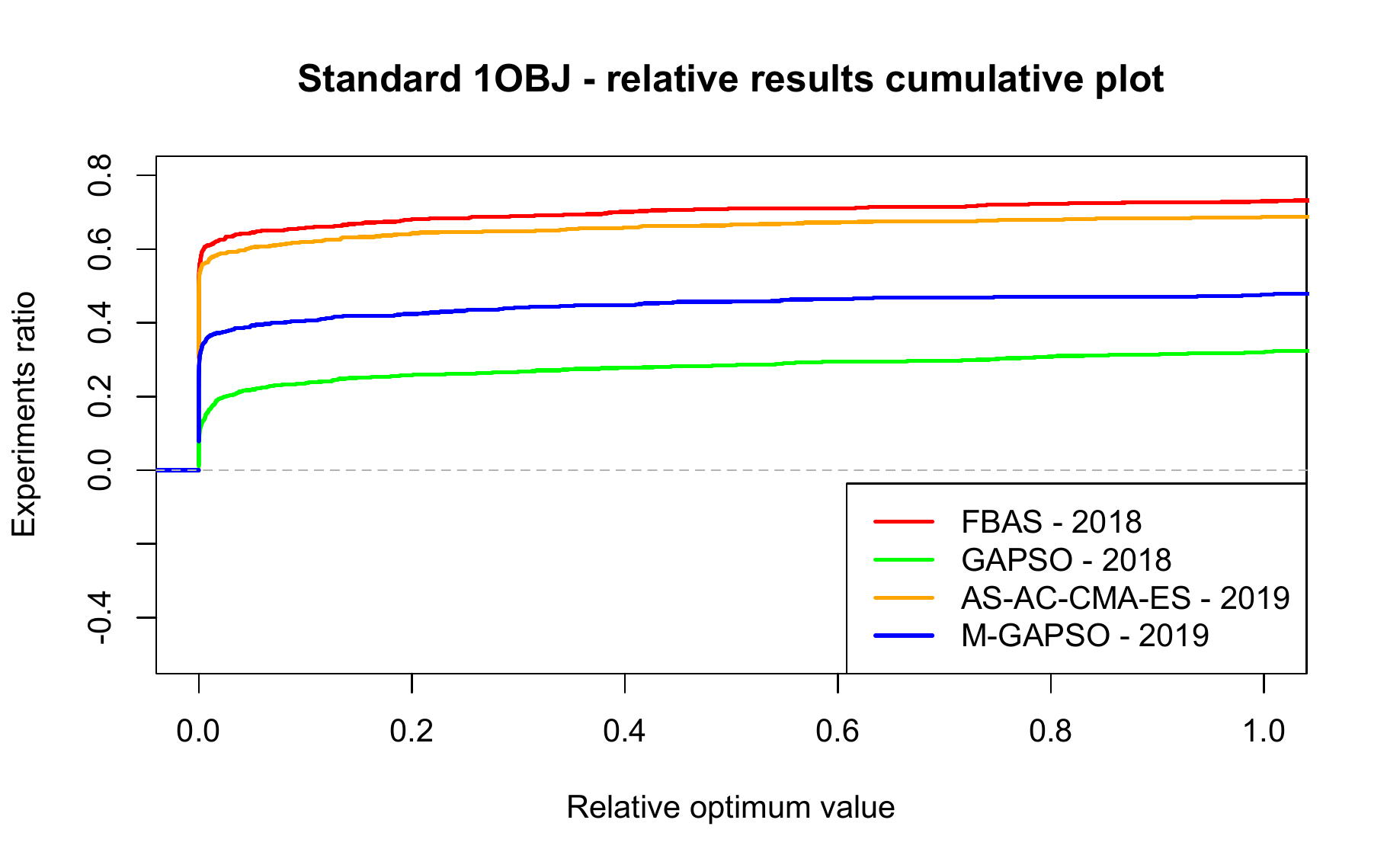}
	\vspace{-4em}
	\caption{Plot of the cumulative relative results of GAPSO (2018), M-GAPSO (2019) and the winning approaches from 2018 and 2019, respectively, in the standard single goal optimization setup.
	\label{fig:bbcomp-1std-relative}}
\end{figure}

\begin{figure}[t]
	\includegraphics[width=\textwidth]{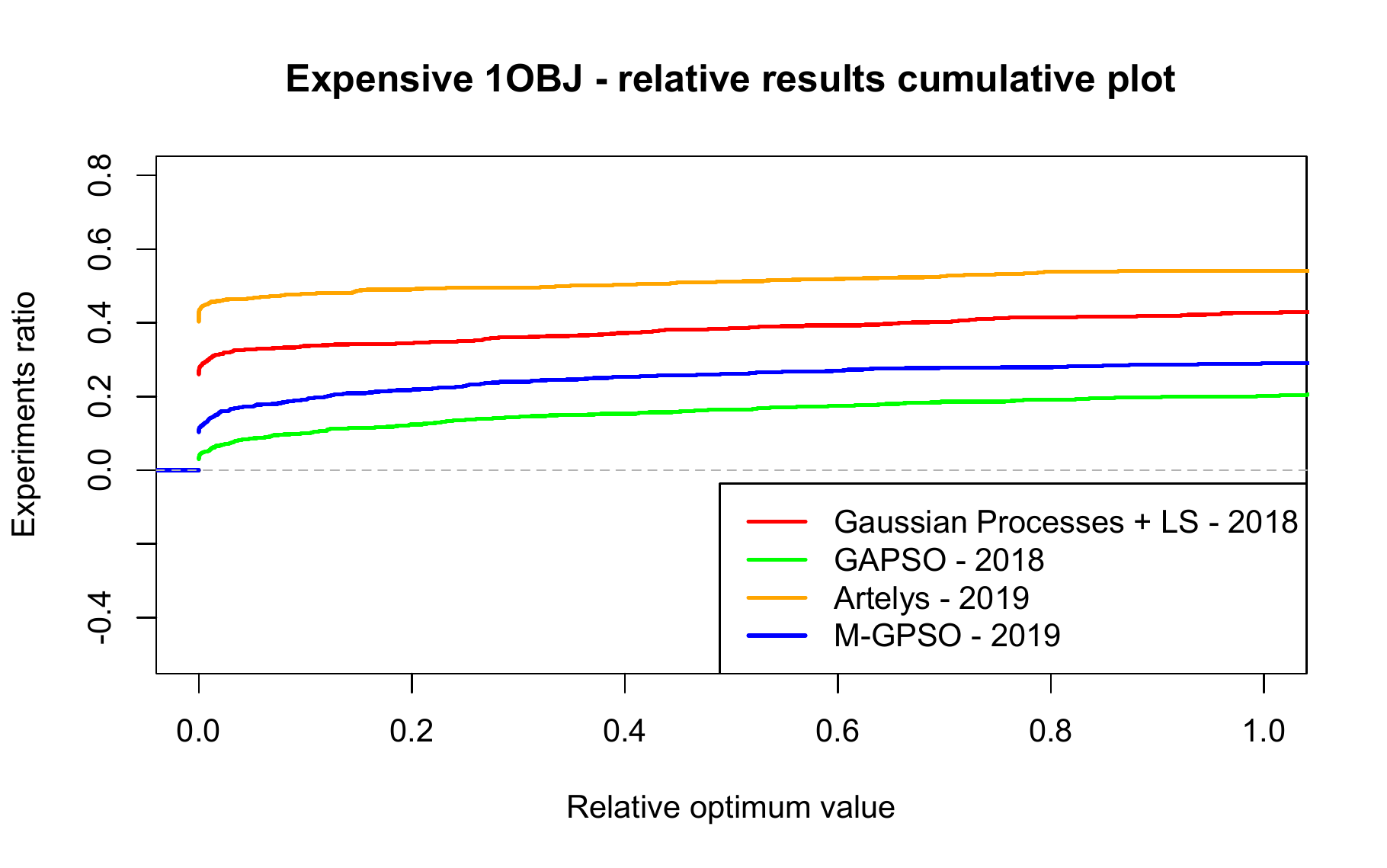}
	\vspace{-4em}
	\caption{Plot of the cumulative relative results of GAPSO (2018), M-GAPSO (2019) and the winning approaches from 2018 and 2019, respectively, in the expensive single goal optimization setup.
	\label{fig:bbcomp-1expensive-relative}}
\end{figure}

GAPSO and M-GAPSO were evaluated against other optimization approaches
during black-box optimization competition BBComp\footnote{\url{https://bbcomp.ini.rub.de/})}
at GECCO conferences.
Summarized results of 2018 and 2019 editions are depicted in Fig.~\ref{fig:bbcomp-official}.
GAPSO run by the authors' faculty team \textit{mini-mlog} scored a 10th place (out of 15 competitors)
in standard single objective optimization and an 8th place (out of 13)
in the expensive single objective optimization tracks
in 2018.
M-GAPSO improved those results in 2019, and was classified 6th (out of 10)
in standard setup and 4th (out of 9) in the expensive optimization track.
In both cases M-GAPSO performed better in 2019, than GAPSO did in 2018, however, the relative difference between results obtained by the authors' approach
and the best approach was reduced only in the standard setup (cf. Fig.~\ref{fig:bbcomp-1std-relative}),
while remained roughly the same in the expensive setup (see Fig.~\ref{fig:bbcomp-1expensive-relative}).

\vspace{-2em}


\section{Conclusions}
\label{sec:conclusions}

The results obtained by M-GAPSO on COCO BBOB benchmark set \cite{nikolaus_hansen_2019_2594848} and in the BBComp competition (\url{https://bbcomp.ini.rub.de/}) are very promising for lower-dimensional functions. At the same time further work is needed to make M-GAPSO competitive also for higher-dimensional functions.


The results support the claim that the behavior pool should consist of significantly variable sampling methods.
The outcomes presented in Fig.~\ref{fig:algorithm-results-all-D-models-vs-no-models} confirm that it is definitely beneficial to supplement PSO and DE with either of the function modeling behaviors, although concurrent use of both linear and polynomial models does not lead to visible performance improvement.
These model-based approaches improved a general performance of M-GAPSO mainly by efficiently
solving instances of \texttt{f1} (spherical) and \texttt{f5} (linear) BBOB functions, however some improvement during the initial function evaluations was also observable.
%
Furthermore, the success of the PSO+DE hybrid approach
should be mainly attributed to the mixing of behaviors mechanism, rather than to adaptation
of the behavior pool (see Fig.~\ref{fig:algorithm-results-all-D-mixing-effects}).

Generally speaking, the improvement introduced in M-GAPSO allowed achieving comparable
results with state-of-the-art CMA-ES \cite{TakahiroYamaguchi2017}
for lower dimensions (up to 5D) and greatly improved performance
over the original GAPSO formulation~\cite{Ulinski2018} (see Fig.~\ref{fig:algorithm-gapso-improvements}).
This improvement can be partially attributed to the already mentioned
samples caching, the addition of modeling approaches, and focusing on mixing behaviors.
\textbf{Significant part of the improvement came from changing the underlying philosophy of the method.
When deciding on the GAPSO's reset and adaptation mechanisms, we were focused on preventing the potential swarm collapse (during the algorithm's run), which was motivated by a theoretical perspective of global optimization algorithms. Within this view, the algorithm should be constructed in a way which ensures that the limit of probability $P(x^{*} \in X_n)$ of including a global optimum $x^{*}$ in the set of tested solutions $X_n$ approaches 1,
as the number of algorithm's iterations $n$ approaches infinity \cite{Eiben1991}.}

\textbf{With M-GAPSO we took a different approach, where individual particles focus
only on the task at hand (relatively quick convergence to high-quality local optimum),
and the burden of exploration lies on higher level mechanisms: the convergence detector ($RestartManager$)
and re-initialization module ($SearchSpaceManager$).}

Our future research is concentrated on utilization of the knowledge of multiple local optima estimations. As could have been observed in Figure~\ref{fig:algorithm-run} positions of 
local optima gathered during the optimization process may form predictable structures which we plan to exploit in our future research.
M-GAPSO should greatly benefit from a more structured approach
to selecting search space boundaries in subsequent algorithm runs.
Therefore, we believe that \textbf{multiple and relatively short algorithm runs,
combined with predictive setup of the algorithm search space boundaries are the key factors in
further improvement of M-GAPSO results.}

\small

\bibliographystyle{apalike}
\bibliography{gapso-with-memory}

\begin{thebibliography}{}

\bibitem[Brest et~al., 2016]{Brest2016}
Brest, J., Maucec, M.~S., and Boskovic, B. (2016).
\newblock {iL-SHADE: Improved L-SHADE algorithm for single objective
  real-parameter optimization}.
\newblock In {\em 2016 IEEE Congress on Evolutionary Computation (CEC)}, pages
  1188--1195. IEEE.

\bibitem[Castro et~al., 2018]{Castro2018}
Castro, O.~R., Fritsche, G.~M., and Pozo, A. (2018).
\newblock {Evaluating selection methods on hyper-heuristic multi-objective
  particle swarm optimization}.
\newblock {\em Journal of Heuristics}, 24(4):581--616.

\bibitem[Cheng et~al., 2019]{Cheng2019}
Cheng, S., Lu, H., Lei, X., and Shi, Y. (2019).
\newblock {Brain Storm Optimization Algorithms: More Questions than Answers}.
\newblock pages 3--32.

\bibitem[Clerc, 2012]{clerc2012standard}
Clerc, M. (2012).
\newblock {Standard particle swarm optimisation}.

\bibitem[{De Jong}, 1975]{DeJong1975}
{De Jong}, K.~A. (1975).
\newblock {\em {Analysis of the behavior of a class of genetic adaptive
  systems}}.
\newblock Phd thesis, University of Michigan.

\bibitem[Du and Li, 2008]{Du2008}
Du, W. and Li, B. (2008).
\newblock {Multi-strategy ensemble particle swarm optimization for dynamic
  optimization}.
\newblock {\em Information Sciences}, 178(15):3096--3109.

\bibitem[Eiben et~al., 1991]{Eiben1991}
Eiben, A.~E., Aarts, E. H.~L., and {Van Hee}, K.~M. (1991).
\newblock {Global convergence of genetic algorithms: A markov chain analysis}.
\newblock pages 3--12.

\bibitem[Hansen et~al., 2019]{nikolaus_hansen_2019_2594848}
Hansen, N., Brockhoff, D., Mersmann, O., Tusar, T., Tusar, D., ElHara, O.~A.,
  Sampaio, P.~R., Atamna, A., Varelas, K., Batu, U., Nguyen, D.~M., Matzner,
  F., and Auger, A. (2019).
\newblock {COmparing Continuous Optimizers: numbbo/COCO on Github}.

\bibitem[Hansen et~al., 2003]{Hansen2003}
Hansen, N., M{\"{u}}ller, S.~D., and Koumoutsakos, P. (2003).
\newblock {Reducing the Time Complexity of the Derandomized Evolution Strategy
  with Covariance Matrix Adaptation (CMA-ES)}.
\newblock {\em Evolutionary Computation}, 11(1):1--18.

\bibitem[Harrison et~al., 2017]{Harrison2017a}
Harrison, K.~R., Engelbrecht, A.~P., and Ombuki-Berman, B.~M. (2017).
\newblock {An adaptive particle swarm optimization algorithm based on optimal
  parameter regions}.
\newblock In {\em 2017 IEEE Symposium Series on Computational Intelligence
  (SSCI)}, pages 1--8. IEEE.

\bibitem[Holland, 1992]{Holland1992}
Holland, J.~H. (1992).
\newblock {\em {Adaptation in natural and artificial systems: an introductory
  analysis with applications to biology, control, and artificial
  intelligence}}.
\newblock MIT Press.

\bibitem[Kennedy and Eberhart, 1995]{PSO:Introduction}
Kennedy, J. and Eberhart, R.~C. (1995).
\newblock {Particle Swarm Optimization}.
\newblock {\em Proceedings of IEEE International Conference on Neural Networks.
  IV}, pages 1942--1948.

\bibitem[Lin et~al., 2017]{LIN2017124}
Lin, J., Wang, Z.-J., and Li, X. (2017).
\newblock {A backtracking search hyper-heuristic for the distributed assembly
  flow-shop scheduling problem}.
\newblock {\em Swarm and Evolutionary Computation}, 36:124--135.

\bibitem[Liu et~al., 2010]{Liu2010}
Liu, H., Cai, Z., and Wang, Y. (2010).
\newblock {Hybridizing particle swarm optimization with differential evolution
  for constrained numerical and engineering optimization}.
\newblock {\em Applied Soft Computing}, 10(2):629--640.

\bibitem[Loshchilov et~al., 2013]{Loshchilov2013}
Loshchilov, I., Schoenauer, M., and S{\`{e}}bag, M. (2013).
\newblock {BI-population CMA-ES Algorithms with Surrogate Models and Line
  Searches}.
\newblock In {\em GECCO '13 Companion Proceedings of the 15th annual conference
  companion on Genetic and evolutionary computation}, pages 1177--1184.

\bibitem[Lynn and Suganthan, 2015]{Lynn2015}
Lynn, N. and Suganthan, P.~N. (2015).
\newblock {Heterogeneous comprehensive learning particle swarm optimization
  with enhanced exploration and exploitation}.
\newblock {\em Swarm and Evolutionary Computation}, 24:11--24.

\bibitem[Ma{\'{n}}dziuk, 2019]{Mandziuk2019survey}
Ma{\'{n}}dziuk, J. (2019).
\newblock {New shades of the Vehicle Routing Problem: Emerging problem
  formulations and Computational Intelligence solution methods}.
\newblock {\em IEEE Transactions on Emerging Topics in Computational
  Intelligence}, 3(3):230--244.

\bibitem[Ma{\'{n}}dziuk and {\.{Z}}ychowski, 2016]{mandziuk2016}
Ma{\'{n}}dziuk, J. and {\.{Z}}ychowski, A. (2016).
\newblock {A memetic approach to vehicle routing problem with dynamic
  requests}.
\newblock {\em Applied Soft Computing}, 48:522--534.

\bibitem[Mladenovi{\'{c}} and Hansen, 1997]{mladenovic1997variable}
Mladenovi{\'{c}}, N. and Hansen, P. (1997).
\newblock {Variable neighborhood search}.
\newblock {\em Computers {\&} operations research}, 24(11):1097--1100.

\bibitem[{Montes de Oca} et~al., 2009]{deOCA4983013}
{Montes de Oca}, M.~A., Pena, J., Stutzle, T., Pinciroli, C., and Dorigo, M.
  (2009).
\newblock {Heterogeneous particle swarm optimizers}.
\newblock In {\em 2009 IEEE Congress on Evolutionary Computation}, pages
  698--705.

\bibitem[Nepomuceno and Engelbrecht, 2012]{nepomuceno2012self}
Nepomuceno, F.~V. and Engelbrecht, A.~P. (2012).
\newblock {A Self-adaptive Heterogeneous PSO Inspired by Ants}.
\newblock In {\em International Conference on Swarm Intelligence}, pages
  188--195.

\bibitem[Okulewicz, 2016]{Okulewicz2016}
Okulewicz, M. (2016).
\newblock {Finding an Optimal Team}.
\newblock In {\em Position Papers of the 2016 Federated Conference on Computer
  Science and Information Systems}, pages 205--210. Polish Information
  Processing Society.

\bibitem[Okulewicz and Ma{\'{n}}dziuk, 2013]{DVRP:2PSO}
Okulewicz, M. and Ma{\'{n}}dziuk, J. (2013).
\newblock {Application of Particle Swarm Optimization Algorithm to Dynamic
  Vehicle Routing Problem}.
\newblock {\em Lecture Notes in Computer Science (including subseries Lecture
  Notes in Artificial Intelligence and Lecture Notes in Bioinformatics)},
  7895:547--558.

\bibitem[Okulewicz and Mandziuk, 2014]{DVRP:2MPSO}
Okulewicz, M. and Mandziuk, J. (2014).
\newblock {Two-phase multi-swarm PSO and the dynamic vehicle routing problem}.
\newblock In {\em 2014 IEEE Symposium on Computational Intelligence for
  Human-like Intelligence (CIHLI)}, pages 1--8, Orlando, Fl, USA. IEEE.

\bibitem[Okulewicz and Ma{\'{n}}dziuk, 2017]{OkulewiczMandziuk2017}
Okulewicz, M. and Ma{\'{n}}dziuk, J. (2017).
\newblock {The impact of particular components of the PSO-based algorithm
  solving the Dynamic Vehicle Routing Problem}.
\newblock {\em Applied Soft Computing}, 58:586--604.

\bibitem[Okulewicz and Ma{\'{n}}dziuk, 2019]{OkulewiczMandziuk2019}
Okulewicz, M. and Ma{\'{n}}dziuk, J. (2019).
\newblock {A metaheuristic approach to solve Dynamic Vehicle Routing Problem in
  continuous search space}.
\newblock {\em Swarm and Evolutionary Computation}, 48:44--61.

\bibitem[Poa{\'{i}}k and Klema, 2012]{Poaik2012}
Poa{\'{i}}k, P. and Klema, V. (2012).
\newblock {JADE, an adaptive differential evolution algorithm, benchmarked on
  the BBOB noiseless testbed}.
\newblock In {\em Proceedings of the fourteenth international conference on
  Genetic and evolutionary computation conference companion - GECCO Companion
  '12}, page 197, New York, New York, USA. ACM Press.

\bibitem[Poli, 2009]{Poli2009}
Poli, R. (2009).
\newblock {Mean and Variance of the Sampling Distribution of Particle Swarm
  Optimizers During Stagnation}.
\newblock {\em IEEE Transactions on Evolutionary Computation}, 13(4):712--721.

\bibitem[Schl{\"{u}}nz et~al., 2018]{SCHLUNZ201858}
Schl{\"{u}}nz, E.~B., Bokov, P.~M., and van Vuuren, J.~H. (2018).
\newblock {Multiobjective in-core nuclear fuel management optimisation by means
  of a hyperheuristic}.
\newblock {\em Swarm and Evolutionary Computation}, 42:58--76.

\bibitem[Shi, 2011a]{Shi2011}
Shi, Y. (2011a).
\newblock {An Optimization Algorithm Based on Brainstorming Process}.
\newblock {\em International Journal of Swarm Intelligence Research}, 2(4):35
  -- 62.

\bibitem[Shi, 2011b]{SHI10.1007/978-3-642-21515-5_36}
Shi, Y. (2011b).
\newblock {Brain Storm Optimization Algorithm}.
\newblock In Tan, Y., Shi, Y., Chai, Y., and Wang, G., editors, {\em Advances
  in Swarm Intelligence}, pages 303--309, Berlin, Heidelberg. Springer Berlin
  Heidelberg.

\bibitem[Storn and Price, 1997]{DE}
Storn, R. and Price, K. (1997).
\newblock {Differential Evolution -- A Simple and Efficient Heuristic for
  global Optimization over Continuous Spaces}.
\newblock {\em Journal of Global Optimization}, 11(4):341--359.

\bibitem[Taillard et~al., 2001]{Taillard2001}
Taillard, {\'{E}}.~D., Gambardella, L.~M., Gendreau, M., and Potvin, J.-Y.
  (2001).
\newblock {Adaptive memory programming: A unified view of metaheuristics}.
\newblock {\em European Journal of Operational Research}, 135(1):1--16.

\bibitem[Uli{\'{n}}ski et~al., 2018]{Ulinski2018}
Uli{\'{n}}ski, M., {\.{Z}}ychowski, A., Okulewicz, M., Zaborski, M., and
  Kordulewski, H. (2018).
\newblock {Generalized Self-adapting Particle Swarm Optimization Algorithm}.
\newblock In {\em Lecture Notes in Computer Science (including subseries
  Lecture Notes in Artificial Intelligence and Lecture Notes in
  Bioinformatics)}, volume 3242, pages 29--40. Springer, Cham.

\bibitem[{Van Den Bergh} and Engelbrecht, 2010]{VanDenBergh2010}
{Van Den Bergh}, F. and Engelbrecht, A.~P. (2010).
\newblock {A convergence proof for the particle swarm optimiser}.
\newblock {\em Fundamenta Informaticae}, 105(4):341--374.

\bibitem[van~der Stockt and Engelbrecht, 2018]{VANDERSTOCKT2018127}
van~der Stockt, S. A.~G. and Engelbrecht, A.~P. (2018).
\newblock {Analysis of selection hyper-heuristics for population-based
  meta-heuristics in real-valued dynamic optimization}.
\newblock {\em Swarm and Evolutionary Computation}, 43:127--146.

\bibitem[Wolpert and Macready, 1997]{Wolpert1997}
Wolpert, D.~H. and Macready, W.~G. (1997).
\newblock {No free lunch theorems for optimization}.
\newblock {\em IEEE transactions on evolutionary computation}, 1(1):67--82.

\bibitem[Yamaguchi and Akimoto, 2017]{TakahiroYamaguchi2017}
Yamaguchi, T. and Akimoto, Y. (2017).
\newblock {Benchmarking the novel CMA-ES restart strategy using the search
  history on the BBOB noiseless testbed}.
\newblock In {\em GECCO '17 Proceedings of the Genetic and Evolutionary
  Computation Conference Companion}, pages 1780--1787.

\bibitem[Yu et~al., 2014]{Yu2014}
Yu, X., Cao, J., Shan, H., Zhu, L., and Guo, J. (2014).
\newblock {An adaptive hybrid algorithm based on particle swarm optimization
  and differential evolution for global optimization.}
\newblock {\em The Scientific World Journal}, 2014:215472.

\bibitem[Zaborski et~al., 2019]{ZaborskiOkulewiczMandziuk2019}
Zaborski, M., Okulewicz, M., and Ma{\'{n}}dziuk, J. (2019).
\newblock {Generalized Self-Adapting Particle Swarm Optimization algorithm with
  model-based optimization enhancements}.
\newblock In {\em Proceedings of 2nd PPRAI Conference}, pages 380--383.

\bibitem[Zhan et~al., 2009]{Zhi-HuiZhan2009a}
Zhan, Z.-H., Zhang, J., Li, Y., and Chung, H. S.-H. (2009).
\newblock {Adaptive particle swarm optimization.}
\newblock {\em IEEE transactions on systems, man, and cybernetics. Part B,
  Cybernetics : a publication of the IEEE Systems, Man, and Cybernetics
  Society}, 39(6):1362--1381.

\end{thebibliography}

\end{document}